\documentclass[twoside, twocolumn]{article}

\usepackage{ArxivVersion/arxiv}

\usepackage{stfloats} 

\usepackage[round,authoryear]{natbib}

\usepackage{amsmath, amsthm, amsfonts, amssymb}
\usepackage{graphicx}
\usepackage{xcolor}
\usepackage{caption}
\usepackage{subcaption}
\usepackage{natbib}

\usepackage{url}
\usepackage{enumitem}
\usepackage{hyperref}

\newtheorem{theorem}{Theorem}[section]
\newtheorem{proposition}[theorem]{Proposition}

\newtheorem{corollary}[theorem]{Corollary}
\newtheorem{assumption}[theorem]{Assumption}

\theoremstyle{definition}
\newtheorem{definition}[theorem]{Definition}
\newtheorem{remark}[theorem]{Remark}
\newtheorem{example}[theorem]{Example}

\newcommand{\gr}[1]{{\color{black}#1}}

\newcommand{\R}{\mathbb{R}}
\newcommand{\E}{\mathbb{E}}

\newcommand{\Law}{\mathcal{L}}

\newcommand{\KSD}{\operatorname{KSD}}

\newcommand{\diag}{\operatorname{diag}}

\newcommand{\norm}[1]{\left\lVert #1\right\rVert}
\newcommand{\abs}[1]{\left\lvert #1\right\rvert}
\newcommand{\dd}{\,\mathrm{d}}

\newcommand{\Lap}{\operatorname{Lap}}
\newcommand{\sgn}{\operatorname{sgn}}

\providecommand{\Var}{\operatorname{Var}}

\newcommand{\QPF}{\mathcal{Q}}
\newcommand{\SPF}{\mathcal{S}}
\newcommand{\TPF}{\Theta}
\newcommand{\EPF}{\mathcal{E}^{\mathrm{PF}}}

\newcommand{\Hk}{{\mathcal H}_K}

\newcommand{\Eta}{\mathrm{H}}

\usepackage{tikz}
\usepackage[normalem]{ulem}

\begin{document}

\twocolumn

\title{Task-aware Kernel Stein certificates for R\'{e}nyi differential privacy}

\author{Leoni Carla Wirth\\
Department of Mathematics \\
Imperial College London\\
London SW7 2AZ, UK\\
\texttt{ l.wirth@imperial.ac.uk}
\And Tara Trauthwein \\
Institute for Mathematical Stochastics\\
University of Münster\\
Münster, Germany  \\
\texttt{ tara.trauthwein@uni-muenster.de}
\And  Gesine Reinert \\
Department of Statistics\\
University of Oxford\\
Oxford OX1 3LB, UK\\
\texttt{ gesine.reinert@stats.ox.ac.uk}
}

\twocolumn[
  \begin{@twocolumnfalse}
    \maketitle
    \vspace{2em} 
  \end{@twocolumnfalse}
]

\begin{abstract}
R\'{e}nyi differential privacy, based on R\'{e}nyi divergence, has become a notion of choice for privacy guarantees. However, outside the Gaussian setting, R\'{e}nyi divergence can be difficult to compute. In particular, R\'{e}nyi differential privacy guarantees that exploit task-specific information, as in R\'{e}nyi Pufferfish privacy, can be challenging. Here we use kernel Stein discrepancies based on escort distributions to provide certificates for R\'{e}nyi differential privacy. By designing the kernel to focus on the relevant privacy directions,  we also derive task-aware privacy certificates. 
We provide {exact identities for Gaussian and Laplace mechanisms and bounds for exponential mechanisms, with separate task-aware guarantees. A real data set example illustrates how the protected comparison set and the released subspace jointly determine the resulting certificates.}
\end{abstract}

\section{Introduction}
Privacy is important because there are many types of information that
one would like to use without allowing them to circulate freely: bank
records, medical data, personal information, or training data for
machine learning models. The goal of privacy mechanisms is to make it
possible to extract useful information from a dataset while limiting
what can be inferred about any single individual. Differential privacy, introduced in \cite{dwork2006differential}, formalizes this by controlling how much the output distribution of a randomized mechanism can change when a single dataset entry is changed. R\'enyi differential privacy (RDP) \citep{mironov2017renyi} quantifies this change using the R\'{e}nyi divergence and is, for example, employed in the popular DP-SGD algorithm \citep{abadi2016deep}. 

{
Standard RDP gives a privacy guarantee uniformly over all neighbouring dataset pairs. This requirement can be stronger than necessary when only certain components of the dataset are sensitive or when only particular changes of the dataset need to be protected, see for example \cite{song2017pufferfish}. Several privacy notions introduce more flexible privacy specifications, for example through protected secrets or comparisons \citep{kifer2014pufferfish,he2014blowfish,pierquin2024renyi}, individual privacy preferences \citep{jorgensen2015conservative}, or sensitive dataset attributes \citep{zhang2022attribute}.

A complementary form of privacy specification concerns the mechanism output. Divergence-based privacy notions typically compare the full output distributions of a mechanism and do not distinguish between the different directions in which these distributions may change. In some applications, however, only selected output components or distributional directions are relevant to the privacy task. For example, \cite{gao2022subspace} introduce subspace differential privacy, which restricts the classical notion of $(\epsilon,\delta)$-differential privacy to a subspace of the output space, allowing for invariants in complementary directions to be preserved.  
 
Here, we combine these two forms of privacy specification using kernel Stein discrepancies. R\'{e}nyi Pufferfish privacy  \citep{pierquin2024renyi} specifies which dataset comparisons require protection, while, in the sense of subspace differential privacy, the kernel is designed to capture which output directions are privacy-relevant. This yields privacy certificates that are task-aware. }

Deriving a privacy certificate requires controlling the R\'{e}nyi divergence between the corresponding output distributions. However, outside the Gaussian setting, R\'{e}nyi divergence can be difficult to compute{, in particular when the distribution has an intractable normalising constant. {This motivates a certificate that is both computationally tractable and flexible enough to incorporate the privacy-relevant structure described above. We show that kernel Stein discrepancies (KSDs) provide such certificates under suitable kernel choices.}
We 
\begin{enumerate}[noitemsep,topsep=0pt]
\item connect KSDs to R\'{e}nyi divergence via escort distributions {and show how the kernel determines the directions captured by the discrepancy; 
    \item derive KSD-based privacy certificate for Gaussian, Laplace and exponential mechanism;
    \item  develop task-aware certificates combining protected Pufferfish comparisons with privacy-relevant output directions}
    \item illustrate the approach on a real data set from the Folktables benchmark \citep{ding2021retiring}. 
\end{enumerate}
{The key contribution is the construction of a discrepancy that matches the protected family of distributional changes, together with conditions under which that discrepancy controls privacy loss. It is not a universally better way to compute every Rényi divergence.}}
The code is available at {\url{https://anonymous.4open.science/r/KSD_Renyi_DP-495A/}}.

\section{Preliminaries}

{{\bf Notation and definitions}} {
Let $\mathcal{X}$ be the {sample} space 
and denote by $\mathfrak{D} = \mathfrak{D}_n := \mathcal{X}^n$ the set of all possible datasets of size $n$. We call $\mathbf{D},\mathbf{D'}\in\mathfrak{D}$ neighbouring and write $\mathbf{D}\sim \mathbf{D'}$ whenever they differ in only a single entry. Let $M:\mathfrak{D}\to \mathcal{Y}$ be a randomized mechanism and $   P_\mathbf{D}:=\mathcal L(M(\mathbf{D}))$ its output distribution.} Write $I_d$ for the $d$-dimensional identity and $\preceq $ for the Loewner order between 
positive semidefinite matrices. 
For a measurable map $T$ {and a probability distribution~$P$} let $T_\# P$ denote the pushforward of $P${, so that for $B$ measurable, $T_\# P(B) = P (T^{-1} (B))$}. For $T:\R^d\to\R^r$ let $J_T\in\R^{r\times d}$ denote its Jacobian. {For probability measures $P$ and $Q$, the notation $Q \ll P$ states that $Q$ is absolutely continuous with respect to $P$. The {\it score function} of $P$ is $s_P =\nabla \log p(x)$, when it exists.} 

For probability measures $Q\ll P$ and $\alpha>1$, the {\it order-$\alpha$ R\'enyi divergence between $Q$ and $P$} is
$D_\alpha(Q\|P)  =  \frac{1}{\alpha-1}\log Z_\alpha(P,Q),$  
where $Z_{\alpha}(P,Q):=\E_P [r^{\alpha}]$
and  $r = dQ/dP$;
   {for a survey see \cite{van2014renyi}.}
Furthermore, define the {\it order-$\alpha$ escort~$Q_{\alpha\mid P}$ of $Q$ relative to $P$} by $$\frac{\dd Q_{\alpha\mid P}}{\dd P}
= \frac{r^\alpha}{Z_\alpha(P,Q)}.$$
{Details on  escort distributions and standard R\'{e}nyi divergence can be found for example in \cite{bercher2012simple}.}

{W}e say that a mechanism $\mathrm{M}$ is $(\alpha,\varepsilon)$-R\'enyi differentially
private (RDP) if, for all neighbouring datasets $\mathbf{D},\mathbf{D'}\in\mathfrak{D}$, we have 
$  D_\alpha
  \left(
    \mathcal{L}(\mathrm{M}(\mathbf{D}))
    \,\middle\|\,
    \mathcal{L}(\mathrm{M}(\mathbf{D'}))
  \right)
  \leq
  \varepsilon .
  $
A comprehensive overview of differential privacy notions can be found in \cite{karmitsa2025comprehensive}.

{{\bf The Langevin KSD}}
 Throughout, we assume the relevant regularity and integrability conditions 
and refer to Appendix {\ref{sec:KSDSetup}} for precise conditions.
 
Let $P$  be a probability measure on $\mathbb R^d$ with
strictly positive smooth density $p$ and let $Q$ be a probability measure on $\R^d$.
Let $K: \R^d \times \R^d \rightarrow \R^{d \times d}$ be a   
{symmetric} positive {semi-}definite kernel, 
{with vector-valued RKHS} 
$\Hk$.
A special case is the 
{isotropic matrix-valued kernel} $K(x,y) = k(x,y) I_d,$ where $I_d$ denotes the  $d$-dimensional identity matrix {and $k(x,y) \in \R$ is a scalar function; 
we then also call $K$ a {\it scalar kernel}}. 
The vector Langevin-Stein operator {is}
\begin{equation}
    \mathcal T_P \phi(x)
    := 
       \nabla \cdot \phi(x)
       + 
       s_P(x)^\top \phi(x),
    \qquad \phi \in \mathcal H_K.
    \label{eq:langevin-stein-operator}
\end{equation} 
The corresponding KSD is
\begin{equation}
    \operatorname{KSD}_K(P,Q)
    :=
    \sup_{\|\phi\|_{\mathcal H_K}\leq 1}
    \left|
        \mathbb E_Q\left[\mathcal T_P\phi(X)\right]
    \right|,
    \label{eq:ksd-definition}
\end{equation}
{which, if we assume that $Q$ has strictly positive density $q$, and under additional assumptions detailed in Appendix~{\ref{sec:KSDSetup}} can also be expressed as}
\begin{align}\label{eq: KSD2 as score difference}
    {\KSD}_K(P,Q)^2 
   &=\E_{X,Y\sim Q}\big[(s_P(X)-s_{Q}(X))^\top \\
   &\qquad \qquad \, \, \, K(X,Y) (s_P(Y)-s_Q(Y))\big]. \notag
\end{align}
{Note that in 
\eqref{eq: KSD2 as score difference}} only the scores of the $P$ {and $Q$ are} 
required, not their normalising constants.

\section{Kernel Stein discrepancy and R\'{e}nyi divergence}\label{sec: renyi ksd comparison}

Here we establish a connection between R\'{e}nyi divergence and kernel Stein discrepancies {(KSDs)} that motivates our use of KSDs as {task-aware} privacy certificates. {Proposition \ref{prop:renyi-first-variation}} 
shows that the first variation of the R\'{e}nyi divergence can be {linked to a KSD involving} an escort distribution. {This identity in particular shows that KSDs characterize directions of change of R\'{e}nyi divergence and are thus suitable to capture the relative importance of different privacy directions.}

From now on, assume that $Q$ has a strictly positive smooth density $q$. We consider perturbations of $Q$ induced by a differentiable vector field $\phi: \R^d \rightarrow \R^d$. Define the pushforward $Q_t:=(I+t\phi)_\#Q$ and note that {with $T_t = I  + t \phi$} has density
\[
q_t(y) =  q( T_t^{-1} (y)) | {\rm{det}} (I_d + t J_\phi (T_t^{-1}(y)) )|^{-1},
\]
where $J_\phi$ denotes the Jacobian of $\phi$. Hence $Q_t$ describes a perturbation of $Q$ with $Q_0=Q$ and initial direction $\phi$. 

\begin{proposition}
\label{prop:renyi-stein-dual-norm}
Under suitable regularity assumptions {(see Appendix~\ref{sec:RenKSDProofs})},
\begin{multline*}
    \sup_{\|\phi\|_{\Hk}\leq 1}
    \left| \frac{\mathrm d}{\mathrm dt}       D_\alpha((I+t\phi)_\#Q\|P) \Big|_{t=0}\right|
    \\=\operatorname{KSD}_K(P,Q_{\alpha\mid P}).
    \label{eq:renyi-ksd-first-variation}
\end{multline*}
\end{proposition}

Proposition~\ref{prop:renyi-stein-dual-norm} provides a local characterization: the escort KSD is the maximal first variation of the R\'{e}nyi divergence over the unit ball of the RKHS. {In particular, the choice of the RKHS and thus of the kernel determines the perturbation directions~$\phi$ along which changes in the R\'{e}nyi divergence are captured. {In Section \ref{sec:pufferfish privacy}, this} 
observation will 
allow us to adapt the KSD to 
privacy-relevant directions.}

Moreover,  for $\alpha=1$, the escort distribution is $Q$ and Proposition \ref{prop:renyi-stein-dual-norm}  illustrates a connection between KSD and the first variation of Kullback--Leibler divergence under smooth transport; see \cite{liu2016svgd}.

\begin{remark}
Under suitable assumptions, the  local representation {in Proposition \ref{prop:renyi-stein-dual-norm}} can be extended to a global representation of the R\'{e}nyi divergence along a particular flow{, see Remark~\ref{rem: laplace-renyi-path-integral} for an example. However, an evaluation would require }
to access the KSD along the entire  
flow. Our privacy certificates {in Section \ref{sec:ksdprivate}} 
instead seek to control Rényi divergence from the KSD evaluated at a single distribution pair.
\end{remark}

\section{KSD as privacy certificate}
\label{sec:ksdprivate}
    We now view {the local connection between R\'{e}nyi divergence and the KSD derived in 
    {Proposition \ref{prop:renyi-first-variation}}
    }
    in the framework of differential privacy {by considering $   P = P_\mathbf{D}$ and $Q = P_\mathbf{D'}$ for neighbouring  $\mathbf{D}\sim \mathbf{D'}$. Then, any bound $D_\alpha(P_{\mathbf D'}\|P_{\mathbf D})\leq \epsilon$ obtained uniformly over $\mathbf D\sim\mathbf D'$ yields an $(\alpha,\epsilon)$-RDP guarantee.}  We identify 
kernel choices under which the KSD tightly controls the corresponding R\'{e}nyi divergence.  {Some}  numerical illustrations  supporting these theoretical results are given in Appendix \ref{app:track}.

\subsection{The Gaussian mechanism}\label{ssec: gaussian mechanism}
For the Gaussian location family, one can compare the  KSD and R\'enyi divergence {directly.} 
Since both quantities are explicitly computable, this provides a tractable setting in which to compare the influence of the operator-valued kernel choice in KSD. {All proofs of this subsection can be found in Appendix~\ref{app: proofs of gaussian mechanism}.}

Let $m_\mathbf{D}\in\R^d$ be the parameter associated to a dataset $\mathbf{D}\in\mathfrak{D}$. The output distribution of the {\it Gaussian mechanism}  is given by
$P_\mathbf{D}=\mathcal N(m_\mathbf{D},\Sigma),$
where $\Sigma\succ0$ {is assumed to be public}. 
For a fixed neighbouring pair $(\mathbf{D},\mathbf{D'})$ we write $P := P_\mathbf{D}$ and $Q := P_{\mathbf{D'}}$, and denote by $\delta := m_{\mathbf{D'}} -m_\mathbf{D}$ the displacement between the means.  
{The following result is standard; see for example Table 1 in \cite{thierrin2022renyi}.
}

\begin{proposition}[R\'{e}nyi divergence]\label{prop:gaussian-renyi}
    The order-$\alpha$ R\'{e}nyi divergence is given by
$
    D_\alpha(Q\|P)
    =
    \frac{\alpha}{2}
    \delta^{\mathsf T}\Sigma^{-1}\delta.
$ 
\end{proposition}

On the other hand, let $K_{B,A} = k_B(x,y) A$ with
\begin{equation}
    \label{eq:gaussiank} k_B(x,y):= \exp\left[    -\frac12(x-y)^{\mathsf T}B^{-1}(x-y)\right]\end{equation}
 be an operator-valued kernel, where $A\succeq 0$
 and $B \succ 0$. {As w}e assume that $\Sigma$ is public, 
 $A$ and $B$ are allowed to depend on $\Sigma$. 

 With the  
 KSD given in \eqref{eq:ksd-definition}, the escort KSD can be expressed as follows.
\begin{proposition}[Escort KSD]
\label{prop:gaussianExactEscordKSD}
    {Under appropriate conditions (see Appendix~\ref{sec:KSDSetup})}, the KSD based on {the} kernel~$K = K_{B,A}$ satisfies 
\begin{equation*}
\KSD_{K_{B,A}}(P,Q_{\alpha\vert P})^2
=
c_{B,\Sigma}\,\alpha^2\,(\Sigma^{-1}\delta)^\top A  \Sigma^{-1}\delta,
\end{equation*}
where $c_{B,\Sigma}:=\E[k_B(X,Y)]={ \{ } \det(I+2\Sigma B^{-1}){ \} } ^{-1/2}.$
\end{proposition}

Thus, 
{for the kernel $K = K_{B,A}$}  the comparison between KSD and R\'{e}nyi divergence {given in Proposition \ref{prop:gaussian-renyi}} reduces to 
comparing the variance~$\Sigma$ and $A$.

{\bf A specific kernel choice.}
We now consider 
{the} anisotropic kernel $K_{B,A}$ {obtained} by setting $A = \Sigma$ to match the covariance structure of the privacy mechanism, i.e. $K_{B,\Sigma}(x,y):=k_B(x,y)\Sigma.$
We 
refer to the Appendix~\ref{app: proofs of gaussian mechanism} for {the isotropic choice} $A = I_d$.

The following theorem shows that this covariance-matched kernel allows to derive an equality (up to constants) of both quantities 
\begin{theorem}
\label{thm:covariance-matched}
For the kernel~$K = K_{B,\Sigma}$ we obtain,
\begin{equation*}
D_\alpha(Q\|P)
=
\frac{1}{2\alpha c_{B,\Sigma}}
\KSD_{K_{B,\Sigma}}(P,Q_{\alpha\vert P})^2 .
\end{equation*}
 {Thus, {if} $\KSD_{K}(P,Q_{\alpha\vert P})^2\leq  \epsilon$ uniformly  over all neighbourhoods $\mathbf{D}\sim\mathbf{D'}$ {then the Gaussian mechanism satisfies} 
 {$(\alpha,\frac{\epsilon}{2\alpha c_{B,\Sigma}})$-RDP.} }
\end{theorem}

{
\begin{remark}[Kernel choice]
The covariance-matched kernel $K_{B,\Sigma}(x,y)=k_B(x,y)\Sigma$ aligns the KSD with R\'{e}nyi divergence, yielding the equality in Theorem~\ref{thm:covariance-matched}. In contrast, the isotropic kernel $K_{B,I_d}(x,y)=k_B(x,y)I_d$ weights directions~$\delta$ by $\Sigma^{-2}$ rather than $\Sigma^{-1}$, and can give a substantially looser certificate; see Appendix~\ref{app: proofs of gaussian mechanism}. Thus, even in this simple Gaussian setting, a suitable kernel choice is essential for obtaining a tight privacy certificate.
\end{remark}}

\subsection{The Laplace mechanism} \label{ssec: laplace mechanism}
The previous subsection showed that adapting the operator-valued kernel to the privacy mechanism can improve the tightness of the bounds. We now focus on the underlying scalar kernel and show that its choice can determine whether the KSD provides a useful global bound. To isolate this effect, we study the Laplace mechanism in one dimension
, so that the relevant choice is the underlying scalar kernel. The proofs of this subsection can be found in Appendix~\ref{app: proofs of laplace mechanism}.

Let $m_\mathbf{D}\in\R$ be the parameter associated to a dataset $\mathbf{D}\in\mathfrak{D}$. The output distribution of the {\it Laplace mechanism}  is given by $P_\mathbf{D} = \Lap(m_\mathbf{D},b)$ with
\begin{equation}
    p_\mathbf{D}(x)=\frac{1}{2b}\exp\left(-\frac{|x-m_\mathbf{D}|}{b}\right),
    \label{eq:laplace-pq}
\end{equation}
where $b > 0$. For a fixed neighbouring pair $(\mathbf{D},\mathbf{D'})$ we write $P:= P_\mathbf{D}$ and $Q:= P_{\mathbf{D'}}$ and denote the corresponding densities by $p = p_\mathbf{D}$ and $q = p_{\mathbf{D'}}$, respectively. Furthermore, we again set $\delta := {m_{\mathbf{D'}} - m_\mathbf{D}}$ to be the displacement between the means.  

{The R\'enyi divergence is available in closed form; see  \cite{gil2013renyi,mironov2017renyi}. }
For $\alpha>1$, define 
\begin{equation*}
    R_{\alpha}(u)
    :=
    \frac{1}{\alpha-1}
    \log\left[
        \frac{
            \alpha e^{(\alpha-1)u}
            +(\alpha-1)e^{-\alpha u}
        }{2\alpha-1}
    \right], \; u>0.
    \label{eq:laplace-R-alpha}
\end{equation*}

\begin{proposition}[R\'{e}nyi divergence]
\label{prop:laplace-renyi}
For $P$ and $Q$ the Laplace output distributions of a neighbouring dataset pair $(\mathbf{D},\mathbf{D'})$, we obtain 
 $  {
    D_\alpha(Q\|P)
    =D_\alpha(P\|Q)
    =R_{\alpha}\big(\frac{{\abs{\delta}}}{b}\big).
    }
    $

\end{proposition}

On the other hand, let $k:\R\times\R\to\R$ be a positive semidefinite scalar kernel. 
Using \eqref{eq: KSD2 as score difference}, the escort KSD can be expressed as follows.
\begin{proposition}[Escort KSD]\label{prop:laplace-general-kernel-ksd}
Under the {conditions detailed in Appendix~\ref{sec:KSDSetup}}, the KSD based on the scalar kernel~$k$ satisfies 
\begin{align*}
&{\KSD}_K(P,Q_{\alpha\vert P})^2\\
&= \frac{4\alpha^2 }{b^2}
    \mathbb E_{X,Y {\sim} Q_{\alpha\vert P}}\left[
        k(X,Y)
        \mathbf 1_{\{X \in I_\delta\}}
        \mathbf 1_{\{Y \in I_\delta\}}
     \right],
\end{align*}
where $X$ and $Y$ are independent {and $I_{\delta} := (\min\{m_\mathbf{D},m_\mathbf{D}+\delta\},\max\{m_\mathbf{D},m_\mathbf{D}+\delta\})$}.
\end{proposition}

Proposition~\ref{prop:laplace-general-kernel-ksd} shows that the discrepancy is supported only on the interval between the two modes. Thus, the global behavior of the KSD is determined by how the underlying kernel~$k$ accounts for this localized discrepancy. We study the constant kernel~$k(x,y):= k_0(x,y):=1$ and refer to Appendix~\ref{app: proofs of laplace mechanism} for a study of the Gaussian RBF kernel.

\begin{theorem}
\label{thm:laplace-escort-derivative}
For $k_0(x,y)\equiv 1$,
\begin{equation*}
    {\operatorname{KSD}_{k_0}(P,Q_{\alpha\vert P})
    =\frac{\alpha}{b}\frac{e^{(\alpha-1)\frac{\abs{\delta}}{b}}-e^{-\alpha\frac{\abs{\delta}}{b}}}{       \alpha e^{(\alpha-1)\frac{\abs{\delta}}{b}}+(\alpha-1)e^{-\alpha\frac{\abs{\delta}}{b}}}.}
\end{equation*}
Set $\kappa_\alpha:=b\operatorname{KSD}_{k_0}(P,Q_{\alpha\vert P})
    \in[0,1).$ Then 
\begin{equation}
  {
    \begin{aligned}
    &D_\alpha(Q\|P) = 
    \frac{\log\left(\frac{1}{1-\kappa_\alpha}\right)}{2\alpha-1}
    -
    \frac{\alpha \log\left(
        1+\frac{\alpha-1}{\alpha}\kappa_\alpha
    \right)}{(\alpha-1)(2\alpha-1)}.
    \end{aligned}
    }
    \label{eq:laplace-renyi-from-escort-kappa}
\end{equation}
\end{theorem}

\begin{remark}[Kernel choice]
The corresponding RKHS consists of constant vector fields and thus $k_0$ is not characteristic on a general class of probability measures. However, as Theorem~\ref{thm:laplace-escort-derivative} shows, it is sufficient here because the $P$ and $Q$ differ only in the location parameter; see \eqref{eq:laplace-pq}. 
{The constant kernel gives an explicit inverse, whereas the RBF kernel requires numerical inversion. For more details see}
Appendix~\ref{app: proofs of laplace mechanism}.
\end{remark}

\subsection{The exponential mechanism} \label{ssec: exponential mechanism}

Section~\ref{ssec: gaussian mechanism} and Section~\ref{ssec: laplace mechanism} showed that the escort KSD can certify R\'{e}nyi differential privacy of the Gaussian and Laplace mechanism, respectively. 
{We now study this correspondence for the exponential mechanism with a focus on regular natural exponential families. In contrast to the Gaussian and Laplace mechanisms, the R\'{e}nyi divergence may involve intractable normalizing constants, whereas the score differences entering the KSD do not. Thus, the derived certificate will not only allow us to incorporate restricted privacy assumptions efficiently, but also improve the computability.}

Let $(P_\eta)_{\eta\in\Eta}$ be the regular natural exponential family
\begin{align}
    P_\eta(\mathrm dy)
    =p_\eta(y)\,\nu(\mathrm dy) 
    =\exp(\eta^\top T(y)-A(\eta)) \nu(\mathrm dy)
    \label{eq:exp-mechanism-definition}
\end{align}
with differentiable sufficient statistic $T{: \R^d \rightarrow \R^r}$ and base measure $ \nu(\mathrm dy)=h_0(y)\,\mathrm dy$ for  $h_0(y)>0$ smooth. Moreover, $A$ is the log-partition function defined by
\begin{equation}
    A(\eta):=\log\int
        \exp\left\{\bigl(\eta^\top T(y)\bigr)\right\}\,\nu(\mathrm dy).
    \label{eq:exp-A-definition}
\end{equation}
We assume that every {$A(\eta)$} 
appearing below is finite. Furthermore, we assume that the exponential family is minimal, so that the Fisher information matrix  $I(\eta) = \nabla^2 A(\eta)$ is positive definite for every $\eta \in \Eta$.

Suppose that the data dependence enters through the natural parameter $\eta = \eta_\mathbf{D}$ and assume that the base density $h_0$ is independent of the dataset.
For neighbouring dataset pairs $(\mathbf{D}, \mathbf{D'})$ with corresponding natural parameters $\eta = \eta_\mathbf{D}$ and $\eta_{\mathbf{D'}}$, we write $P=P_\eta$, $Q=P_{\eta+\delta}$, $\delta := \eta_{\mathbf{D'}}- \eta_\mathbf{D} $, and $\eta_t=\eta+t\delta$, $t\in [0,\alpha]$. Furthermore, we assume that $\eta_t\in\Eta$ for every $t\in[0,\alpha]$. {Given a dataset $\mathbb{D}$, }{ the {\it exponential mechanism} }{returns a random output $Y\sim P_{\eta_{\mathbb{D}}}$.} A detailed introduction to the exponential mechanism in this setting as well as all proofs are given in Appendix~\ref{app: proofs of exponential mechanism}.

{
For every $\alpha>1$ for which $A(\eta_\alpha)<\infty$, 
\begin{equation}
    \label{eq:exp-renyi-log-partition}
 D_\alpha(Q\|P)
    =
    \frac{
       A(\eta_\alpha)-\alpha A(\eta_1)+(\alpha-1)A(\eta_0)
    }{\alpha-1};
    \end{equation} 
    see for example (15) in \cite{thierrin2022renyi}.}
{Here we give} 
an {alternative}  expression for the R\'{e}nyi divergence.

\begin{proposition}[R\'enyi divergence]
\label{prop:exp-exact-renyi}
Let $P = P_{\eta}$ be a member of the natural exponential family defined above.
Then for every $\alpha>1$ for which $A(\eta_\alpha)<\infty$,
\begin{equation}
\begin{split}
    D_\alpha(Q\|P)
    &= \frac{\alpha}{2} \delta^\top \bar{I}((\eta_t)_{t=0}^\alpha) \delta
\end{split}
    \label{eq:exp-renyi-log-partition2}
\end{equation}
with $ { \bar{I}(\!(\eta_t)_{t=0}^\alpha)\!:= \!\! \frac{2}{\alpha} \!\int_0^1 \!t  I(\eta_t)  dt\! + \!\frac{2}{\alpha(\alpha-1)} \!\int_1^\alpha\! (\alpha-t)  \!I(  \eta_t )   dt.}$
\end{proposition}

Although \eqref{eq:exp-renyi-log-partition} looks explicit, its ingredients may not be easy to compute as it requires evaluating the (possibly intractable) log-partition function~$A$.  
In contrast, the KSD is built up on score functions which can be evaluated without knowledge of the normalization constants; {the computational burden depends on escort sampling and on establishing the comparison constants}. To make this precise, we now derive an explicit expression of the KSD certificate.

Let $K:\mathbb R^d\times\mathbb R^d\to\mathbb R^{d\times d}$ be an operator-valued kernel {such that the} {matrix}
\begin{equation*}
    G_K(\eta_t)
    :=
    \mathbb E_{X,Y\sim P_{\eta_t}}
    \left[
        J_T(X) K(X,Y)J_T(Y)^{\mathsf T}
    \right]\in \R^{{r \times r} }
\end{equation*}
{is well-defined.} 
Then the escort KSD obtains the following explicit expression.
\begin{proposition}(Escort KSD)\label{prop:exp-exact-escort-ksd}
Let $P = P_{\eta}$ be a member of the natural exponential family {\eqref{eq:exp-mechanism-definition}}. 
Under the {conditions laid out in Appendix~\ref{sec:KSDSetup}}, 
\begin{equation*}
   {
    \operatorname{KSD}_K(
        P,Q_{\alpha\vert P}
    )^2
    =
    \alpha^2\,\delta^{\mathsf T}
    G_K(\eta_\alpha)
    \delta.
    }
\end{equation*}
\end{proposition}

{
A comparison of Proposition~\ref{prop:exp-exact-renyi} and \ref{prop:exp-exact-escort-ksd} reveals the same role of the kernel as in the previous two subsections. More precisely, controlling  R\'{e}nyi divergence by the escort KSD reduces to comparing the Fisher information~$I(\eta)$ and the kernelized metric~$G_K(\eta_\alpha)$. }

For every $\eta\in\Eta$ define $\Delta_\eta = \{ \eta_{\mathbf{D'}} - \eta_{\mathbf{D}}: \mathbf{D} \sim \mathbf{D'} \text{ and }\eta_\mathbf{D} = \eta\}\subset \R^r$ to be the set of attainable neighbouring directions at $\eta$. 
To relate the Fisher information and the kernelized metric uniformly over neighbouring datasets, we make the following two assumptions.

\begin{assumption}[Poincare inequality on $\Eta$] \label{ass: restrictedpoincare}
Assume that there exist a constant $0<C_\Eta<\infty$ such that, for every $\eta\in\Eta$, 
$\delta\in\Delta_\eta$ and 
$t\in [0,\alpha]$,
\begin{align*}
\Var_{P_{\eta_t}}\left(\delta^\top T(X)\right)
&\leq C_\Eta \,\E_{P_{\eta_t}}\left\|J_T(X)^\top \delta\right\|^2.
\end{align*} 
\end{assumption}
For the regular natural exponential family, $\delta^\top I(\eta) \delta = \Var_{P_{\eta}}\left(\delta^\top T(X)\right)$, see \cite[Corollary 3.1]{wainwright2008graphical}. Thus, {Assumption \ref{ass: restrictedpoincare}} 
relates the Fisher information to the squared gradient of the sufficient statistic~$T${, while the {next} assumption 
relates this gradient to the kernelized matrix~$G_K$.}
\begin{assumption}\label{ass: coercivity}
Assume  that there exists a constant $\kappa_\Eta >0$ such that, for every $\eta\in\Eta$ and 
$\delta \in \Delta_{\eta}$,
    \begin{align*}  
    \bar{J}((\eta_t)_{t=0}^\alpha) \le \frac{1}{\kappa_\Eta} \delta^\top G_K(\eta_\alpha) \delta, 
\end{align*}
where $\bar{J}((\eta_t)_{t=0}^\alpha):= \Big\{  \frac{2}{\alpha} \int_0^1 t  \E_{P_{\eta_t}}\left\| J_T(X)^\top \delta\right\|^2  \, dt + \frac{2}{\alpha(\alpha-1)} \int_1^\alpha (\alpha -t) \E_{P_{\eta_t}} \left\| J_T(X)^\top \delta\right\|^2 \, dt \Big\}. $
\end{assumption}
{Assumption \ref{ass: coercivity} excludes kernels that are insensitive to a protected change and quantifies how strongly the endpoint discrepancy detects that change relative to its variation along the interpolation path.}
Under these assumptions, we finally derive an escort KSD certificate for the R\'{e}nyi divergence.
\begin{theorem}\label{thm:exp-escort-ksd-certificate}
Under Assumptions \ref{ass: restrictedpoincare} and \ref{ass: coercivity} we have
    \begin{align}\label{eq:diffbound}
     D_\alpha (Q  || P) \le  \frac{ C_\Eta}{2\alpha \, \kappa_\Eta} \KSD_{K} (P , Q_{\alpha | P})^2.
\end{align}
{Thus, {if} $\KSD_{K}(P,Q_{\alpha\vert P})^2\leq  \epsilon$ uniformly  over all neighbourhoods $\mathbf{D}\sim\mathbf{D'}$ 
{then the exponential mechanism satisfies} $(\alpha,\frac{ C_\Eta}{2\alpha \, \kappa_\Eta}\epsilon)$-RDP.} 
\end{theorem}

\begin{remark}
    {W}ithin the class of natural regular exponential families we cannot hope for a general escort KSD bound of the R\'{e}nyi divergence. For example for the exponential family $p_\eta(x) = -\eta e^{\eta x}$, for $x\ge0$ and $\eta <0$, we have for all $\eta_\mathbf{D'} = \eta + \delta <0$, so that 
    by {\eqref{eq:exp-renyi-log-partition}}, 
$$ D_\alpha (Q || P) = \frac{1}{\alpha-1} \log \Big[ \frac{ (-\eta - \delta)^\alpha (-\eta)^{1-\alpha}}{-\alpha \delta - \eta}\Big], $$
   which is finite only when $\eta+\alpha\delta<0$. In particular, the escort density cannot be normalized. Our assumptions provide sufficient conditions for a certificate, but are not claimed to be necessary.
\end{remark}

We conclude this section with three applications for which Assumptions~\ref{ass: restrictedpoincare} and \ref{ass: coercivity} are satisfied.

\begin{corollary}[Laplace family]\label{cor:exp-laplace}
Let $P_\eta =\operatorname{Lap}(0,-\eta^{-1})$, $\eta \in \Eta\subset \R_{<0}$, be the centered Laplace family, written in natural form as
$p_{\eta}(x) = \exp(\eta \abs{x} -A(\eta))$ 
with sufficient statistic~$T(x) = \abs{x}$ and Lebesgue base measure.
Assume that $\Eta$ is chosen such that $\eta_{\min}\leq \eta_t\leq \eta_{\max}<0$ for all $t\in[0,\alpha]$ and let $K(x,y) = k(x,y) = xy$.
    Then $\KSD_{K} (P , Q_{\alpha | P})^2 \leq \epsilon$ uniformly over all $\mathbf{D}\sim\mathbf{D'}$ implies $(\alpha,\frac{\eta_{\min}^2}{2\alpha \eta_{\max}^2}\epsilon )$-RDP.
\end{corollary}

\begin{corollary}[Gaussian family]\label{cor:exp-gaussian}
Let $P = P_\eta$ belong to the Gaussian
natural exponential family with sufficient statistic $T(x)=x$, base measure $\nu = \mathcal N(a,\Sigma)$,  $a\in\R^d$ and ${\Sigma\succ 0}$. Set
$K(x,y)=\psi(\norm{x-y})I_d$ for some ${\psi:\R\to\R_+}$. Then, for ${\kappa_\Eta = \E_{{Z}\sim\mathcal{N}(0,2\Sigma)}[\psi(\norm{{Z}
})] }$ and $ \lambda_{\max}(\Sigma) $ the largest eigenvalue of $\Sigma$, a bound $\KSD_{K} (P , Q_{\alpha | P})^2 \leq \epsilon$ uniformly over all $\mathbf{D}\sim\mathbf{D'}$ implies $(\alpha,\frac{ \lambda_{\max}(\Sigma)}{2\alpha \, \kappa_\Eta}\epsilon )$-RDP.
\end{corollary}

\begin{corollary}[Quartic family]\label{cor:exp-quartic}
Let $P = P_\eta$ belong to the natural exponential family with sufficient statistic $T(x) = x$ and base density $h_0(x)\propto \exp\left(-\frac12x^\top\Sigma x
-\frac{\gamma}{4}\sum_{i=1}^d x_i^4\right),$  for some $ \Sigma\succ 0$ and $\gamma>0$. Consider the constant kernel $K(x,y) = I_d$ and let $\lambda_{\min}(\Sigma)$ be the smallest eigenvalue of $\Sigma$. Then $\KSD_{K} (P , Q_{\alpha | P})^2 \leq \epsilon$ uniformly over all $\mathbf{D}\sim\mathbf{D'}$ implies $(\alpha,\frac{ \epsilon}{2\alpha\lambda_{\min}(\Sigma) } )$-RDP.
\end{corollary}

\section{Task-aware KSD 
certificates for R\'{e}nyi Pufferfish privacy}\label{sec:pufferfish privacy}
In {Section \ref{sec:ksdprivate}} 
we showed how escort KSDs can be used to derive certificates for {RDP},  
{and} that the kernel choice determines which changes in the output distribution are captured by the certificate. We now build on this observation by constructing kernels that target specified privacy tasks, leading to task-aware R\'{e}nyi privacy certificates.

\subsection{R\'{e}nyi Pufferfish privacy}\label{ssec: renyi pufferfish privacy}

A Pufferfish framework \citep{kifer2014pufferfish} is a triple $(\SPF,\QPF,\TPF),$ where $\SPF$ is a collection of secrets, $\QPF\subseteq\SPF\times\SPF$ is the collection of protected secret pairs, 
{while} $\TPF$ {indexes} a class of data-generating distributions~$\mathbf{P}_\theta$ and represents admissible adversarial prior knowledge. {For any $\theta\in\TPF$, let $\mathbf{P}_{\theta}$ denote a dataset distribution on $\mathfrak{D}$ and let $\mathbf{D}\sim\mathbf{P}_{\theta}$. In particular, the dataset~$\mathbf{D}$ is now random. We 
remark that the protected secret pairs in ~$\QPF$ are ordered because R\'{e}nyi divergence is {asymmetric. }}

For a secret $s\in\SPF$ with $\mathbf{P}_\theta(s)>0$, define the 
output law $P_{\theta,s}
:=
\Law\bigl(M(\mathbf{D})\mid s,\theta\bigr),$ where $\mathbf{D}\sim\textbf{P}_\theta$.

\begin{definition}[R\'{e}nyi Pufferfish privacy, \cite{pierquin2024renyi}]
\label{def:rpp}
Let $\alpha>1$ and $\varepsilon\geq0$. A mechanism $M$ is $(\alpha,\varepsilon)$-R\'{e}nyi Pufferfish private (RPP) in $(\SPF,\QPF,\TPF)$ if
$
D_\alpha\left(P_{\theta,s_j}\middle\|P_{\theta,s_i}\right)
\leq\varepsilon
$ 
for every $\theta\in\TPF$ and every $(s_i,s_j)\in\QPF$ such that $P_\theta(s_i) > 0 $ and $P_\theta(s_j) > 0 $.
\end{definition}
{We refer to Appendix~\ref{app: pufferfish} for a more detailed introduction to R\'{e}nyi Pufferfish privacy and a collection of its properties.}


\subsection{Task-aware R\'{e}nyi Pufferfish privacy}\label{ssec: task-aware renyi pufferfish privacy}
{Subsection~\ref{ssec: renyi pufferfish privacy} introduces RPP, which we will also refer to as \textit{full} RPP, as it controls the R\'{e}nyi divergence between the full outputs $P_{\theta,s_j}$ and $P_{\theta,s_i}$.  
We now introduce a weaker, task-specific guarantee by additionally restricting the output directions that are protected.}
Following the idea of subspace differential privacy \citep{gao2022subspace}, we restrict the privacy guarantee to a specified output subspace, but {in contrast to \cite{gao2022subspace} we} combine this restriction with Rényi Pufferfish privacy instead of $(\epsilon,\delta)$-differential privacy. This yields a task-aware privacy guarantee that restricts both the protected secret comparisons and the protected output directions.


Let $\mathbb{S}\subseteq \mathcal{Y} = \R^d$  be a public task subspace that is fixed independently of the datasets and set ${r = \text{dim}(\mathbb{S})}$. The subspace specifies the components of the mechanism output that are relevant for the required task. Let
${U_\mathbb{S}\in\R^{d\times r}}$ be the orthonormal basis of $\mathbb{S}$. 
We define the task-aware mechanism by mapping the output~$M(\mathbf{D})$ of the original mechanism to its coordinates in the task subspace $\mathbb{S}$, i.e. by~$ U_\mathbb{S}^\top M(\mathbf{D})$. 

\begin{definition}[Task-projected R\'{e}nyi Pufferfish private]\label{def: restricted output privacy}
Let $\alpha >1$ and $\varepsilon\geq 0$. We call a mechanism $M$ \textit{$(\alpha,\epsilon)$-task-aware R\'{e}nyi Pufferfish private} with task subspace~$\mathbb{S}$ and Pufferfish framework $(\SPF,\QPF,\TPF)$ if
$
D_\alpha\left((U_{\mathbb{S}}^\top)_\#P_{\theta,s_j}\middle\|(U_{\mathbb{S}}^\top)_\#P_{\theta,s_i}\right)
\leq\varepsilon
$ 
for every $\theta\in\TPF$ and every $(s_i,s_j)\in\QPF$ such that $\textbf{P}_\theta(s_i)>0$ and $\textbf{P}_\theta(s_j)>0$.
\end{definition}
{Task-aware RPP is in general strictly weaker than full RPP: it protects only the task-aware output~$ U_\mathbb{S}^\top M(\mathbf{D})$ and places no privacy constraint on changes in $\mathbb{S}^\perp$. In particular, full RPP implies task-aware RPP.}

\subsection{Task-aware kernel design}\label{ssec: task aware kernel design}
We now illustrate how the underlying kernel can be used to obtain task-aware privacy guarantees. We first return to the Gaussian setting of Subsection \ref{ssec: gaussian mechanism}, now for the secret-conditioned output laws. Assume that 
$P_{\theta,s}
=\mathcal{N}(m_{\theta,s},\Sigma),
$ where $m_{\theta,s}\in\R^d$ depends on the secret $s\in\SPF$ and 
$\theta\in\TPF$, while the covariance  $
{\Sigma\succ0}$ does not change across secrets. 
For a fixed protected pair $(s_i,s_j)\in\QPF$ and a 
$\theta\in\TPF$, we write $P:= P_{\theta,s_i} $ and $Q:= P_{\theta,s_j} $, and denote by $\delta=m_{\theta,s_j}-m_{\theta,s_i}$ the displacements between the means. 
{T}he order-$\alpha$ escort of $Q$ relative to $P$ is 
$Q_{\alpha\mid P} = \mathcal{N}(m_{\theta,s_i}+\alpha\delta,\Sigma)$, see \ref{app: proofs of gaussian mechanism}.

As in Subsection ~\ref{ssec: gaussian mechanism} we study the anisotropic kernel
$K_{B,A}(x,y) := k_B(x,y)A$ for some $B\succ 0$. We focus on 
$A = A_{\mathbb{S}}^d = (\Sigma U_\mathbb{S})(U_\mathbb{S}^{\top}\Sigma U)^{-1}(\Sigma U_\mathbb{S})^\top\in \R^{d\times d}$.

\begin{theorem}
\label{thm:task-rbf}
Let $\mathbb{S}$ be a public task subspace with orthonormal basis~$U_\mathbb{S}\in\R^{d\times r}$ and kernel $K_{B,A_\mathbb{S}^d}$. Then
\begin{equation*}
D_\alpha((U_\mathbb{S}^\top)_{\#}Q\|(U_\mathbb{S}^\top)_{\#}P)
=
\frac{1}{2\alpha c_{B,\Sigma}}
\KSD_{K_{B,A_\mathbb{S}^d}}(P,Q_{\alpha\mid P})^2.
\end{equation*}
If furthermore $ \Sigma^{-1}\delta\in \mathbb{S}$, then
 \begin{equation*}
 D_\alpha(Q\|P)=
 \frac{1}{2\alpha c_{B,\Sigma}}
 \KSD_{K_{B,A_\mathbb{S}^d}}(P,Q_{\alpha\mid P})^2.
 \end{equation*}
\end{theorem}

\begin{remark}[Task awareness] 
{
By projecting $P$ and $Q$ onto $\mathbb{S}$, we protect only the output directions and thus obtain only task-aware RPP. 
However, if all privacy-relevant changes lie in $\mathbb{S}$, i.e.\,
if $ \Sigma^{-1}\delta\in\mathbb{S}$, {then} task-aware and full RPP align and KSD certifies full RPP.}
\end{remark}
{
\begin{remark}[Kernel choice]
The kernel choice is again essential. Here, we require a kernel that retains the output directions specified by the task subspace while disregarding directions that are intentionally excluded by the task. This illustrates the power of KSD: through the choice of kernel, one can specify which distributional directions the certificate measures and, thus, 
which privacy-relevant changes it is designed to detect. We note that such a kernel should be built on a public, justified task specification, rather than chosen post hoc to suppress privacy-relevant directions. 
\end{remark}}

\subsection{Task-aware certificates for the exponential family}\label{ssec: task aware exponential mechanism}
Finally, we derive task-aware certificates for the exponential mechanism. As in Subsection ~\ref{ssec: exponential mechanism}, we focus on regular natural exponential families and thus assume that the secret-conditioned output laws~$P_{\theta,s}$ are part of one of these families, where the dependence on the secret~$s\in\SPF$ and on the prior~$\theta\in\TPF$ enters through the natural parameter $\eta = \eta_\mathbf{\theta,s}$. 
Then, for a pair of secrets $(s_i,s_j)\in\QPF$ we set $\eta = \eta_{\theta,s_i}$ and write $P=P_\eta$, $Q=P_{\eta+\delta}$, $\delta := \eta_{\theta,s_j}- \eta_{\theta,s_i} $, as well as $\eta_t=\eta+t\delta$. We assume that $\eta_t\in\Eta$ for every $t\in[0,\alpha]$. Finally, for every $\eta\in\Eta$ we set $\Delta_\eta^{\QPF} = \{ \eta_{\theta,s_j} - \eta_{\theta,s_i}: \theta\in\Theta, (s_i,s_j)\in \QPF \text{ and }\eta_{\theta,s_i} = \eta\}\subset \R^r$ to be the set of protected secret-comparison directions at~$\eta$. {Note that the pairs of secrets are ordered as  R\'{e}nyi divergence is {asymmetric}. If  both orientations are intended, both $(s_i,s_j)$ and $(s_j,s_i)$ should be included in $\QPF$ or verified separately.}

The specified {Pufferfish} restrictions allow to relax Assumptions~\ref{ass: restrictedpoincare} and \ref{ass: coercivity} by specifying the set of attainable neighbouring directions from $\Delta_\eta$ to $\Delta_\eta^{\QPF}$, see Assumptions~\ref{ass: restrictedpoincarePufferfish} and \ref{ass: coercivityPufferfish} in Appendix~\ref{app: task aware}.

{
We now consider a special kernel $K_{\mathbb S}$ that restricts the KSD to the task subspace $\mathbb S$. Under the assumptions above, this task-aware kernel can nevertheless provide a certificate for full RPP.}

\begin{proposition}\label{prop: task aware exponential mechanism certificate}
Let $\mathbb{S}$ be a public task subspace with projector $U_{\mathbb{S}}$ and define $K_{\mathbb{S}}(x,y):= k(x,y)U_{\mathbb{S}}U_{\mathbb{S}}^\top$ to be a task-aware kernel with scalar kernel~$k$.
Assume that Assumptions~\ref{ass: restrictedpoincarePufferfish} and \ref{ass: coercivityPufferfish} are satisfied with $K=K_{\mathbb{S}}$.
Then 
    \begin{align*}
     D_\alpha(Q\|P) \le  \frac{ C_\Eta}{2\alpha \, \kappa_\Eta} \KSD_{K_{\mathbb{S}}} (P , Q_{\alpha | P})^2.
\end{align*}
{
{A} uniform bound on $\KSD_{K_{\mathbb{S}}}$ for all $\theta\in\TPF$ and $(s_i,s_j)\in\QPF$ implies the corresponding RPP guarantee.}
\end{proposition}

\begin{remark}
We remark that the  {Pufferfish ($\Delta^\QPF_\eta$) restriction and the subspace restriction ($K =  K_{\mathbb{S}}$)} are complementary: 
$\Delta_\eta^\QPF$ specifies which secret comparisons require protection,   which makes Assumptions \ref{ass: restrictedpoincarePufferfish} and \ref{ass: coercivityPufferfish} easier to verify {than Assumptions \ref{ass: restrictedpoincare} and \ref{ass: coercivity}}, while $\mathbb{S}$ specifies the privacy-relevant output directions. If $J_T(x)^\top \delta\in \mathbb{S}$ for all $ \delta\in\Delta_{\eta}^\QPF$, then  Assumption~\ref{ass: coercivity} with kernel $K(x,y) = k_B(x,y)I_d$ implies Assumption~\ref{ass: coercivityPufferfish} with kernel $K = K_{\mathbb{S}}$. 
\end{remark}

\begin{corollary}\label{cor: task aware privacy example}
Let $P = P_{\theta,s}$ belong to the natural exponential family with sufficient statistic $T(x) = \frac{1}{2} (a^\top x)^2 \in \R$, $x\in \R^d$ and $a\in\R^d$ such that $\norm{a}=1$, and base density $h_0(x)$ 
as in Corollary~\ref{cor:exp-quartic}. Let $\mathbb{S}= \text{span}(a)$  be the public task subspace. 
Assume that $\Delta_\eta^\QPF \subset \R_+$ for all ${\eta\in \Eta}$,  and that ${\eta_{\max} := \sup_{\substack{\eta\in \Eta,\, \delta \in \Delta_\eta^\QPF}} (\eta + \alpha {\delta} ) < (a^\top \Sigma^{-1} a)^{-1}}$. {Also assume that 
$$\kappa_{\Eta} = \inf_{{\eta\in\Eta;  \delta\in\Delta_\eta^\QPF}} \E_{X\sim P_{\eta + \alpha\delta}}[(a^\top X)^2] >0.$$} 
{Take $k(x,y) = (a^\top x ) (a^\top y)$, so that  
$K_{\mathbb{S}} (x,y) =  (a^\top x ) (a^\top y) a a^\top.$} 
Then, if $\KSD_{K_{\mathbb{S}}} (P , Q_{\alpha | P})^2 \leq \epsilon$ uniformly over all ${\theta\in\TPF}$ and $(s_i,s_j)\in\QPF$,  {we obtain} 
$(\alpha,\frac{ {\epsilon} C_\Eta}{2\alpha \, \kappa_\Eta})$-RPP, 
where $C_\Eta = a^\top (\Sigma - a \eta_{\max} a^\top)^{-1} a $.
\end{corollary}
\begin{remark}
  {With the kernel $K_S$ of Corollary \ref{cor: task aware privacy example}, and  $m_\alpha = \E_{P_{\eta + \alpha \delta}} [( a^\top X)^2]$, Proposition \ref{prop:exp-exact-escort-ksd} gives that $\KSD_{K_S} (P, Q_{\alpha | P})^2 = \alpha^2 \delta^2 m_\alpha^2.$ If instead we chose the constant scalar kernel giving $K_\mathbb{S} = a a ^\top$ then by symmetry, $\KSD_{K_\mathbb{S}} (P, Q_{\alpha | P})^2 = \alpha^2 \delta^2 \E_{Q_{\alpha | P}} [( a^\top X)]^2 = 0$, see Appendix \ref{app:densequartic}. Hence the constant kernel misses the protected distributional change, while the feature kernel in Corollary \ref{cor: task aware privacy example} detects it.} 
\end{remark}

\begin{example}\label{ex:densequartic}
 {Take}
 $d=40, \alpha \in \{ 1, 2, 4, 8\}, \gamma=0.1$, $\delta = 0.1094 $, and $\Sigma$ and $a$ are generated such that ${\delta} a^\top \Sigma^{-1} a =0.05$. {Details can be found in Appendix \ref{app:densequartic}.} 
While the KSD can be calculated in closed form, the escort moment $m_\alpha$ still has to be estimated numerically; we use MCMC for this task. The results are shown in Table \ref{tab:dense-quartic-alpha-study}. The {estimated} KSD certificates are of the same order of magnitude as the approximated R\'{e}nyi divergence, but are faster to compute. 
 Thus, if runtime is an issue, KSD may be preferred.

 \begin{table*}[t]

\centering
\small
\begin{tabular}{c@{\hspace{1.1em}}rrrrrrr}
\hline 
$\alpha$ & $\widehat D_\alpha$ & {est.} KSD bound \eqref{bound:ksd} & bound / $\widehat D_\alpha$ & fallback \eqref{bound:fallback} & $t_D$ (s) & $t_{\rm KSD}$ (s) & $C_{{\rm H}}$ \\
\hline 
$1$ (KL) & $5.588e-04$ & $1.300e-03$ & $2.327$ & $1.385e-03$ & 1.75 & $0.57$ & $0.4810$\\
  & sd $2.98e{-05}$ & sd $4.69e{-05}$ &  & & & & \\ 
$2$ & $1.137e-03$ & $2.868e-03$ & $2.521$ & $3.086e-03$ & $2.83$ & $0.56$ & $0.5078$\\
& sd $5.12e{-05}$ & sd $1.03e{-05}$ &  & & & & \\ 
$4$ & $2.463e-03$ & $7.171e-03$ & $2.911$ & $7.812e-03$ & $5.21$ & $0.58$ & $0.5713$\\
& sd $6.60e{-05}$ & sd $2.65e-04$ &  & & & & \\ 
$8$ & $5.621e-03$ & $2.389e-02$ & $4.25$ & $2.778e-02$ & $9.79$ & $0.56$ & $0.7614$\\
& sd $1.23e-04$ & sd $9.35e-04$ &  & & & & \\ 
\hline
\end{tabular}
\caption{Dense-quartic order study; 
50 independent MCMC repetitions. The order $\alpha\!=\!1$ row is the KL limit. The  KSD bound  is the {estimated} normalized task-aware privacy certificate \eqref{bound:ksd} {(with the MCMC estimate for the escort moment)}. 
The fallback option is the cruder bound \eqref{bound:fallback}.  The $t_D$ and $t_{\KSD}$ columns show the runtimes.
}   
\label{tab:dense-quartic-alpha-study}
\end{table*}

\end{example}

{
\section{ACSIncome illustration}
\label{sec:acs-real-data}

We use the ACSIncome data from the Folktables benchmark \citep{ding2021retiring} to give a real-data
illustration of the {output-side task restriction} 
developed above.  
The ACSIncome data 
{is a popular benchmark; see for example \cite{bullwinkel2022evaluating}}.   
The classical task is to predict $y$, indicating whether an individual's annual income exceeds \$50,000, using demographic and employment variables such as age, education, occupation, marital status, hours worked, sex, and race. 
We are given a sample of 20,000 records from California, and 20,000 records from Texas. In this experiment the 2018 California records form public auxiliary data; the 2018 Texas records are the pseudo-private data. 
}  
We fit a public logistic regression model with ridge penalty to the public data release. For the full data release we use a Gaussian mechanism with a vector release, and for the task-aware mechanism we use the same Gaussian variables but  release only its scalar projection.  {Details are found in Appendix \ref{app:acs-real-data}.}

Figure \ref{fig:taskawareresults} {gives the standard RDP and the task-aware RDP, plotted against privacy budget; the dashed lines indicate empirical 95\% quantiles serving as diagnostics. The  
 task-aware RDP certificate uses a smaller privacy budget than the standard release, for a different release mechanism.
 
 In Appendix \ref{app:acs-real-data} we also consider a Pufferfish example, keeping sex and race indicators secret for a particular individual. Figure \ref{fig:puffer} shows that restricting the released output can change the relative importance of different protected comparisons.}

\begin{figure}
    \centering
    \includegraphics[width=\linewidth]{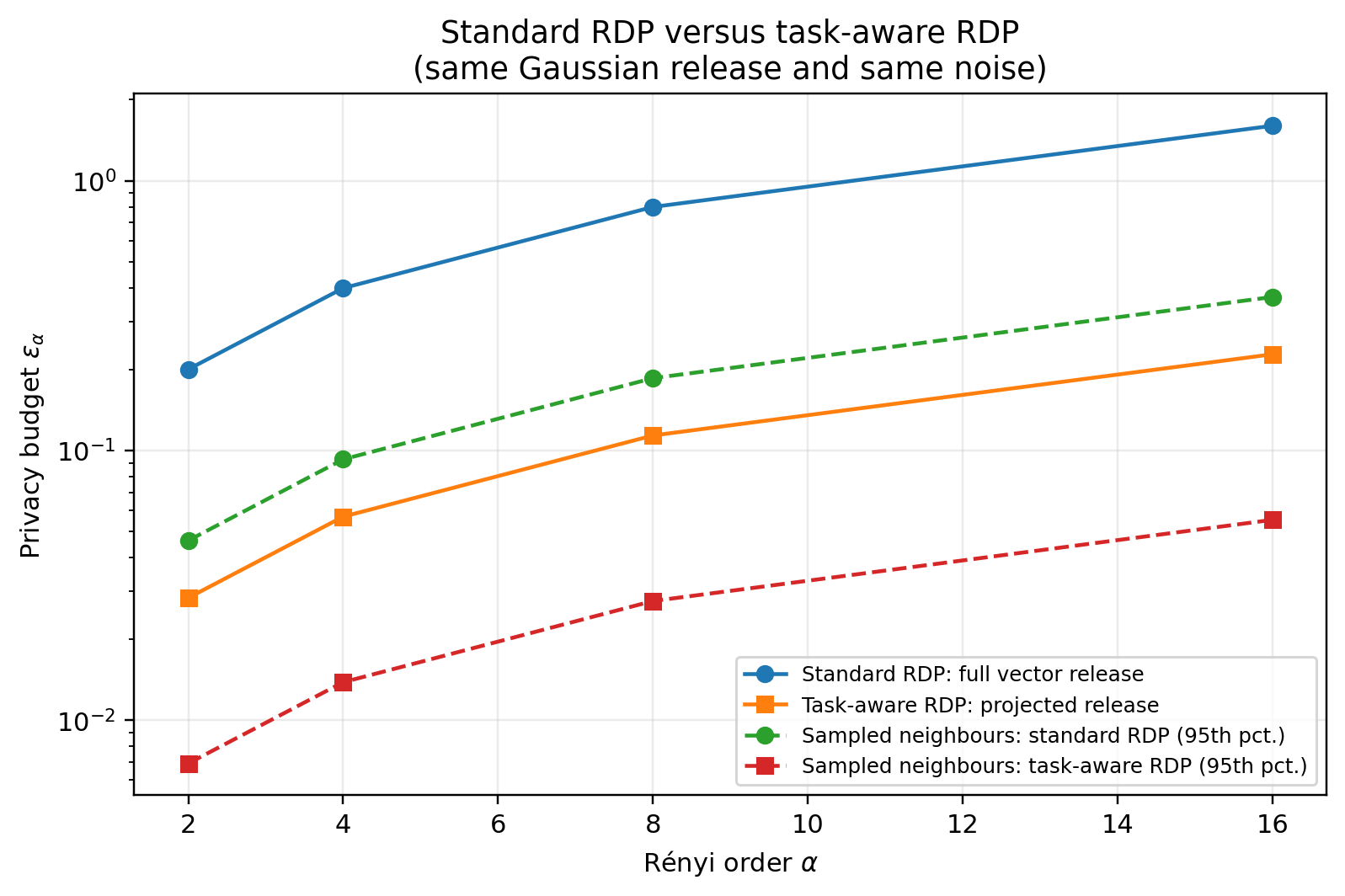}
    \caption{{The full Gaussian release and its one-dimensional task-aware release using the same Gaussian noise. Solid curves show the analytic certificates,  and dashed curves show the empirical 95th percentiles over sampled replace-one Texas training pairs.}}
    \label{fig:taskawareresults}
\end{figure}

\section{Conclusion}
{We used kernel Stein discrepancies to derive tractable certificates for R\'{e}nyi differential privacy that can incorporate privacy specifications. The kernel plays a central role in these certificates by determining which directions of change are captured.
This flexibility leads to task-aware certificates that combine restrictions on the protected comparisons with restrictions on the relevant output directions. Our theoretical examples and the ACSIncome illustration demonstrate the resulting framework in concrete settings.
} 
{While the KSD computations are not necessarily fast, in future work we shall explore how} 
Nystr\"om approximations can reduce the large-sample computational cost {for KSD computation}; see 
\cite{kalinke2024nystr}.

  {\gr{P}rivacy mechanisms should be evaluated also regarding their utility. We leave a study of connections with utility measures as future work; the choice of privacy kernel for certification could depend on the downstream task. Kernel choice in itself is a direction for future research, as is the treatment of other privacy notions in the KSD framework.}

Lastly we note that privacy can be in tension with other desirable criteria such as fairness; see for example \cite{fioretto2022differential}. Depending on the criteria, output of private mechanisms may create undesirable side-effects which have to be acknowledged separately.

\subsection*{AI use statement}

In this work, we used generative AI tools to help develop \gr{the conceptual framework of escort distributions,}
formulate mathematical claims \gr{relating to connections between RDP and KSD}, provide critical ingredients for proving mathematical claims \gr{in Proposition A.1} , assist in the writing of proofs, 
design  \gr{experiments} and 
provide feedback on research  methodology and experiments, implement methods, \gr{and} interpret results.
We have not used generative AI tools to generate synthetic data sets, assist with translation, clean and reformat dataset, support qualitative and thematic data analysis.
Additionally, we used generative AI tools to create 
software code, draft parts of a research paper, identify relevant literature, \gr{suggest edits to the paper}
to improve readability, 
to summarize 
existing literature \gr{and identify gaps}, for brainstorming, sourcing/searching for information, 
\gr{and to} suggest experimental parameters. 
We have reviewed all AI-assisted work. More precisely, we checked all LLM-generated research ideas for potential plagiarism {and for overlap with existing work} through a manual literature survey. Furthermore, we checked all LLM-assisted claims and proofs line by line {by independently working out the resulting arguments, checking every step for correctness, and supplying all necessary details, justifications, and assumptions.} The LLM-generated code was verified and tested for correctness by {one} author. We independently evaluated all LLM-suggested methodological approaches and experimental designs for their validity and suitability.  Additionally, all LLM-assisted interpretations and conclusions were independently assessed against the underlying results. All sources and citations were independently verified. The final research decisions remain with the authors.

We take responsibility for the final content of this work, including text, claims, or artifacts produced with the aid of generative AI.

\bibliography{ArxivVersion/references}

@article{thierrin2022renyi,
  title={R{\'e}nyi cross-entropy measures for common distributions and processes with memory},
  author={Thierrin, Ferenc Cole and Alajaji, Fady and Linder, Tam{\'a}s},
  journal={Entropy},
  volume={24},
  number={10},
  pages={1417},
  year={2022},
  publisher={MDPI}
}

@article{holly2011fourth,
  title={Fourth order pseudo maximum likelihood methods},
  author={Holly, Alberto and Monfort, Alain and Rockinger, Michael},
  journal={Journal of Econometrics},
  volume={162},
  number={2},
  pages={278--293},
  year={2011},
  publisher={Elsevier}
}

@article{bullwinkel2022evaluating,
  title={Evaluating the fairness impact of differentially private synthetic data},
  author={Bullwinkel, Blake and Grabarz, Kristen and Ke, Lily and Gong, Scarlett and Tanner, Chris and Allen, Joshua},
  journal={arXiv preprint arXiv:2205.04321},
  year={2022}
}

@inproceedings{abadi2016deep,
  title={Deep learning with differential privacy},
  author={Abadi, Martin and Chu, Andy and Goodfellow, Ian and McMahan, H Brendan and Mironov, Ilya and Talwar, Kunal and Zhang, Li},
  booktitle={Proceedings of the 2016 ACM SIGSAC Conference on Computer and Communications Security},
  pages={308--318},
  year={2016}
}

@article{karmitsa2025comprehensive,
  title={A comprehensive guide to differential privacy: From theory to user expectations},
  author={Karmitsa, Napsu and Airola, Antti and Pahikkala, Tapio and Pitk{\"a}m{\"a}ki, Tinja},
  journal={arXiv preprint arXiv:2509.03294},
  year={2025}
}

@article{ding2021retiring,
  title={Retiring adult: New datasets for fair machine learning},
  author={Ding, Frances and Hardt, Moritz and Miller, John and Schmidt, Ludwig},
  journal={Advances in neural information processing systems},
  volume={34},
  pages={6478--6490},
  year={2021}
}

@phdthesis{wynne2023thesis,
  author  = {Wynne, George},
  title   = {Contributions in Functional Data Analysis and Functional-Analytic Statistics},
  school  = {Imperial College London},
  year    = {2023},
  month   = mar
}

@inproceedings{awan2019exponential,
  author    = {Awan, Jordan and Kenney, Ana and Reimherr, Matthew and Slavkovi{\'c}, Aleksandra},
  title     = {Benefits and Pitfalls of the Exponential Mechanism with Applications to {H}ilbert Spaces and Functional {PCA}},
  booktitle = {Proceedings of the 36th International Conference on Machine Learning},
  series    = {Proceedings of Machine Learning Research},
  volume    = {97},
  pages     = {374--384},
  year      = {2019},
  publisher = {PMLR}
}

@article{van2014renyi,
  title={R{\'e}nyi divergence and {K}ullback-{L}eibler divergence},
  author={Van Erven, Tim and Harremoes, Peter},
  journal={IEEE transactions on information theory},
  volume={60},
  number={7},
  pages={3797--3820},
  year={2014},
  publisher={IEEE}
}

@inproceedings{liu2016svgd,
  author    = {Liu, Qiang and Wang, Dilin},
  title     = {{Stein} Variational Gradient Descent: A General Purpose {B}ayesian Inference Algorithm},
  booktitle = {Advances in Neural Information Processing Systems 29},
  year      = {2016}
}

@InProceedings{chwialkowski2016kernel,
  title = 	 {A Kernel Test of Goodness of Fit},
  author = 	 {Chwialkowski, Kacper and Strathmann, Heiko and Gretton, Arthur},
  booktitle = 	 {Proceedings of The 33rd International Conference on Machine Learning},
  pages = 	 {2606--2615},
  year = 	 {2016},
  editor = 	 {Balcan, Maria Florina and Weinberger, Kilian Q.},
  volume = 	 {48},
  series = 	 {Proceedings of Machine Learning Research},
  address = 	 {New York, New York, USA},
  month = 	 {20--22 Jun},
  publisher =    {PMLR},
  _pdf = 	 {http://proceedings.mlr.press/v48/chwialkowski16.pdf},
  _url = 	 {https://proceedings.mlr.press/v48/chwialkowski16.html},
}

@InProceedings{liu2016kernelized,
  title = 	 {A Kernelized {S}tein Discrepancy for Goodness-of-fit Tests},
  author = 	 {Liu, Qiang and Lee, Jason and Jordan, Michael},
  booktitle = 	 {Proceedings of The 33rd International Conference on Machine Learning},
  pages = 	 {276--284},
  year = 	 {2016},
  editor = 	 {Balcan, Maria Florina and Weinberger, Kilian Q.},
  volume = 	 {48},
  series = 	 {Proceedings of Machine Learning Research},
  address = 	 {New York, New York, USA},
  month = 	 {20--22 Jun},
  publisher =    {PMLR},
  _pdf = 	 {http://proceedings.mlr.press/v48/liub16.pdf},
  _url = 	 {https://proceedings.mlr.press/v48/liub16.html}
}

@inproceedings{mironov2017renyi,
  title={R{\'e}nyi differential privacy},
  author={Mironov, Ilya},
  booktitle={2017 IEEE 30th Computer Security Foundations Symposium (CSF)},
  pages={263--275},
  year={2017},
  organization={IEEE}
}

@inproceedings{NEURIPS2019_ba7609ee,
 author = {Barp, Alessandro and Briol, Francois-Xavier and Duncan, Andrew and Girolami, Mark and Mackey, Lester},
 booktitle = {Advances in Neural Information Processing Systems},
 editor = {H. Wallach and H. Larochelle and A. Beygelzimer and F. d\textquotesingle Alch\'{e}-Buc and E. Fox and R. Garnett},
 pages = {},
 publisher = {Curran Associates, Inc.},
 title = {Minimum Stein Discrepancy Estimators},
 volume = {32},
 year = {2019}
}

@inproceedings{dwork2006differential,
  title={Differential privacy},
  author={Dwork, Cynthia},
  booktitle={International colloquium on automata, languages, and programming},
  pages={1--12},
  year={2006},
  organization={Springer}
}

@article{kifer2014pufferfish,
  title={Pufferfish: A framework for mathematical privacy definitions},
  author={Kifer, Daniel and Machanavajjhala, Ashwin},
  journal={ACM Transactions on Database Systems (TODS)},
  volume={39},
  number={1},
  pages={1--36},
  year={2014},
  publisher={ACM New York, NY, USA}
}

@inproceedings{pierquin2024renyi,
  title={R{\'e}nyi pufferfish privacy: General additive noise mechanisms and privacy amplification by iteration via shift reduction lemmas},
  author={Pierquin, Cl{\'e}ment and Bellet, Aur{\'e}lien and Tommasi, Marc and Boussard, Matthieu},
  booktitle={Forty-first International Conference on Machine Learning},
  year={2024}
}

@article{kalinke2024nystr,
  title={Nystr\"{o}m kernel {S}tein discrepancy},
  author={Kalinke, Florian and Szab{\'o}, Zolt{\'a}n and Sriperumbudur, Bharath K},
  journal={arXiv preprint arXiv:2406.08401},
  year={2024}
}

@article{fioretto2022differential,
  title={Differential privacy and fairness in decisions and learning tasks: A survey},
  author={Fioretto, Ferdinando and Tran, Cuong and Van Hentenryck, Pascal and Zhu, Keyu},
  journal={arXiv preprint arXiv:2202.08187},
  year={2022}
}

@article{bercher2012simple,
  title={A simple probabilistic construction yielding generalized entropies and divergences, escort distributions and q-{G}aussians},
  author={Bercher, J-F},
  journal={Physica A: Statistical Mechanics and its Applications},
  volume={391},
  number={19},
  pages={4460--4469},
  year={2012},
  publisher={Elsevier}
}

@article{wainwright2008graphical,
author = {Wainwright, Martin J. and Jordan, Michael I.},
title = {Graphical Models, Exponential Families, and Variational Inference},
year = {2008},
issue_date = {January 2008},
volume = {1},
number = {1–2},

journal = {Found. Trends Mach. Learn.},
month = jan,
pages = {1–305},
numpages = {305}
}

@book{bakry2014analysis,
  title     = {Analysis and Geometry of Markov Diffusion Operators},
  author    = {Bakry, Dominique and Gentil, Ivan and Ledoux, Michel},
  series    = {Grundlehren der mathematischen Wissenschaften},
  volume    = {348},
  publisher = {Springer},
  year      = {2014}
}

@book{steinwart2008support,
  title     = {Support Vector Machines},
  author    = {Steinwart, Ingo and Christmann, Andreas},
  year      = {2008},
  publisher = {Springer},
  address   = {New York},
  doi       = {10.1007/978-0-387-77242-4}
}

@article{gil2013renyi,
  title={R{\'e}nyi divergence measures for commonly used univariate continuous distributions},
  author={Gil, Manuel and Alajaji, Fady and Linder, Tamas},
  journal={Information Sciences},
  volume={249},
  pages={124--131},
  year={2013},
  publisher={Elsevier}
}

@inproceedings{he2014blowfish,
  title={Blowfish privacy: Tuning privacy-utility trade-offs using policies},
  author={He, Xi and Machanavajjhala, Ashwin and Ding, Bolin},
  booktitle={Proceedings of the 2014 ACM SIGMOD International Conference on Management of data},
  pages={1447--1458},
  year={2014}
}

@inproceedings{song2017pufferfish,
  title={Pufferfish privacy mechanisms for correlated data},
  author={Song, Shuang and Wang, Yizhen and Chaudhuri, Kamalika},
  booktitle={Proceedings of the 2017 ACM International Conference on Management of Data},
  pages={1291--1306},
  year={2017}
}

@inproceedings{zhang2022attribute,
  title={Attribute privacy: Framework and mechanisms},
  author={Zhang, Wanrong and Ohrimenko, Olga and Cummings, Rachel},
  booktitle={Proceedings of the 2022 ACM Conference on Fairness, Accountability, and Transparency},
  pages={757--766},
  year={2022}
}

@inproceedings{jorgensen2015conservative,
  title={Conservative or liberal? {P}ersonalized differential privacy},
  author={Jorgensen, Zach and Yu, Ting and Cormode, Graham},
  booktitle={2015 IEEE 31St International Conference on Data Engineering},
  pages={1023--1034},
  year={2015},
  organization={IEEE}
}

@article{pike2012stein,
  title={Stein's method and the {L}aplace distribution},
  author={Pike, John and Ren, Haining},
  journal={ALEA},
  year={2014}, 
  volume={11}, 
  pages={571-587}
}

@inproceedings{gao2022subspace,
  title={Subspace differential privacy},
  author={Gao, Jie and Gong, Ruobin and Yu, Fang-Yi},
  booktitle={Proceedings of the AAAI Conference on Artificial Intelligence},
  volume={36},
  pages={3986--3995},
  year={2022}
}

@inproceedings{wang2019subsampled,
  title={Subsampled {R}{\'e}nyi differential privacy and analytical moments accountant},
  author={Wang, Yu-Xiang and Balle, Borja and Kasiviswanathan, Shiva Prasad},
  booktitle={The 22nd international conference on artificial intelligence and statistics},
  pages={1226--1235},
  year={2019},
  organization={PMLR}
}
\bibliographystyle{plainnat}

\section*{Checklist}

\begin{enumerate}

  \item For all models and algorithms presented, check if you include:
  \begin{enumerate}
    \item A clear description of the mathematical setting, assumptions, algorithm, and/or model. [Yes]
    \item An analysis of the properties and complexity (time, space, sample size) of any algorithm. [Not Applicable]
    \item (Optional) Anonymized source code, with specification of all dependencies, including external libraries. [Yes]
  \end{enumerate}

  \item For any theoretical claim, check if you include:
  \begin{enumerate}
    \item Statements of the full set of assumptions of all theoretical results. [Yes]
    \item Complete proofs of all theoretical results. [Yes]
    \item Clear explanations of any assumptions. [Yes]
  \end{enumerate}

  \item For all figures and tables that present empirical results, check if you include:
  \begin{enumerate}
    \item The code, data, and instructions needed to reproduce the main experimental results (either in the supplemental material or as a URL). [Yes]
    \item All the training details (e.g., data splits, hyperparameters, how they were chosen). [Yes]
    \item A clear definition of the specific measure or statistics and error bars (e.g., with respect to the random seed after running experiments multiple times). [Yes]
    \item A description of the computing infrastructure used. (e.g., type of GPUs, internal cluster, or cloud provider). [Yes]
  \end{enumerate}

  \item If you are using existing assets (e.g., code, data, models) or curating/releasing new assets, check if you include:
  \begin{enumerate}
    \item Citations of the creator If your work uses existing assets. [Yes]
    \item The license information of the assets, if applicable. [Yes]
    \item New assets either in the supplemental material or as a URL, if applicable. [Yes]
    \item Information about consent from data providers/curators. [Not Applicable]
    \item Discussion of sensible content if applicable, e.g., personally identifiable information or offensive content. [Not Applicable]
  \end{enumerate}

  \item If you used crowdsourcing or conducted research with human subjects, check if you include:
  \begin{enumerate}
    \item The full text of instructions given to participants and screenshots. Not Applicable]
    \item Descriptions of potential participant risks, with links to Institutional Review Board (IRB) approvals if applicable. [Not Applicable]
    \item The estimated hourly wage paid to participants and the total amount spent on participant compensation. [Not Applicable]
  \end{enumerate}

\end{enumerate}

\clearpage
\appendix
\thispagestyle{empty}

\onecolumn

{\section{KSD Setup}\label{sec:KSDSetup}
In this section we provide the technical details for the setup of kernel Stein discrepancies. {In particular we provide assumptions for \eqref{eq: KSD2 as score difference} in the main text to hold.}

We recall that we assume $P$ to be a probability measure on $\R^d$ with smooth positive density $p$ and $Q$ to be a probability measure on $\R^d$.

We start by defining the Stein class of $P$. A function $\phi:\R^d \rightarrow \R^d$ is in the Stein class of $P$ if 
\begin{itemize}
    \item $\phi \in \mathcal{C}^1(\R^d,\R^d)$;
    \item the vector field $p\phi$ is integrable and its integral over spheres vanishes:
    \[
    \lim_{R \rightarrow \infty} \int_{\partial B_R(0)} p(x) \phi(x) \cdot n(x) dS(x)=0,
    \]
    where $B_R(0)$ denotes the ball of radius $R$ centred at $0$ and $n(x)$ is the unit normal vector at $x$, with $S$ the surface measure on $\partial B_R(0)$;
    \item $\E_P |\mathcal{T}_P\phi(x)|<\infty$.
\end{itemize}

Let $K: \R^d \times \R^d \rightarrow \R^{d \times d}$ {be a} vector-valued, symmetric positive {semi}-definite kernel and let $\mathcal H_K$ denote its vector-valued RKHS.
{We further assume that} 
$K$ 
{is} twice differentiable (i.e. $K_{ij} \in \mathcal{C}^2(\R^d \times \R^d)$ for all $1\leq i,j\leq d$). {It then follows that} 
$\mathcal{H}_K \subset \mathcal{C}^1(\R^d)$, see \cite[Corollary 4.36]{steinwart2008support}. 
{Moreover} 
we assume that $\E_Q|\mathcal{T}_P\phi(X)|<\infty$ for every $\phi \in \Hk$. {Then the expression} 
\[
KSD_K(P,Q):= \sup_{\|\phi\|_{\Hk}\leq 1} |\E_Q [\mathcal{T}_P\phi(X)]|
\]
{is well-defined.}

Assume that $Q$ has strictly positive smooth density $q$ and assume in addition to the above that
\begin{itemize}
    \item for every $\phi \in \Hk$, one has that $\phi$ is in the Stein class of $Q$;
    \item it holds that
    \begin{equation}\label{eq:cond4}
        \E_Q \sqrt{(s_P(X)-s_Q(X))^\top K(X,X)(s_P(X)-s_Q(X))}<\infty.
    \end{equation}
\end{itemize}
Then \eqref{eq: KSD2 as score difference} holds.

This is shown in {Supplementary Material B.5.1 of \cite{NEURIPS2019_ba7609ee}; for} the one-dimensional setting see  also \cite[Lemma 2.3 \& Theorem~3.8]{liu2016kernelized}. 
For further RKHS formulations, including double-expectation or spectral representations of KSD, see for example Chapter 5 in \cite{wynne2023thesis} and \cite{liu2016kernelized,chwialkowski2016kernel}.}

{For completeness here we provide a proof.} 

\begin{proposition}\label{prop:ksdform}
Assume that $P$, $Q$, and $K$ satisfy the conditions stated above, including
that every $\phi\in\mathcal H_K$ belongs to the Stein class of $Q$ and that
Condition~\ref{eq:cond4} holds. Then, for independent $X,Y\sim Q$,
\[
\operatorname{KSD}_K(P,Q)^2
=
\mathbb E_{X,Y\sim Q}
\left[
(s_P(X)-s_Q(X))^\top K(X,Y)(s_P(Y)-s_Q(Y))
\right].
\]
\end{proposition}
\begin{proof}
Set
$
b(x):=s_P(x)-s_Q(x)$ and $
w(x):=\bigl(b(x)^\top K(x,x)b(x)\bigr)^{1/2},
$ and note that Assumption \eqref{eq:cond4} states that 
$$C:=\mathbb E_Q[w(X)]<\infty.$$ 
For $x\in\mathbb R^d$, the vector-valued reproducing property gives
that $ 
K(\cdot,x)b(x)\in\mathcal H_K,
$ that $
\|K(\cdot,x)b(x)\|_{\mathcal H_K}=w(x),
$ and that $
b(x)^\top\phi(x)
=
\langle K(\cdot,x)b(x),\phi\rangle_{\mathcal H_K}.
$  
Hence, for every $\phi\in\mathcal H_K$,
\[
\mathbb E_Q\bigl[|b(X)^\top\phi(X)|\bigr]
\leq C\|\phi\|_{\mathcal H_K}.
\]
Thus
$
\Lambda(\phi):=\mathbb E_Q[b(X)^\top\phi(X)]
$ 
is a bounded linear functional on the real Hilbert space $\mathcal H_K$.
Moreover, since $\phi$ belongs to the Stein class of $Q$,
\[
\mathbb E_Q[(T_Q\phi)(X)]
=
\lim_{R\to\infty}
\int_{\partial B_R(0)}q(x)\phi(x)\cdot n(x)\,dS(x)
=0.
\]
Using $T_P\phi=T_Q\phi+b^\top\phi$, we obtain
\[
\mathbb E_Q[(T_P\phi)(X)]=\Lambda(\phi).
\]
By the Riesz representation theorem, there is a unique $h\in\mathcal H_K$
such that
\[
\Lambda(\phi)=\langle h,\phi\rangle_{\mathcal H_K}
\quad\text{for all }\phi\in\mathcal H_K,
\qquad
\|h\|_{\mathcal H_K}\leq C.
\]
We next identify this representer pointwise. For every $y\in\mathbb R^d$
and $a\in\mathbb R^d$, applying the reproducing property and the definition
of $\Lambda$ yields
\[
\begin{aligned}
a^\top h(y)
&=\langle h,K(\cdot,y)a\rangle_{\mathcal H_K} =\Lambda(K(\cdot,y)a) =\mathbb E_Q[b(X)^\top K(X,y)a] =a^\top\mathbb E_Q[K(y,X)b(X)].
\end{aligned}
\]
The expectation is finite: by the reproducing property and Cauchy--Schwarz,
\[
|a^\top K(y,x)b(x)|
\leq
\bigl(a^\top K(y,y)a\bigr)^{1/2}w(x).
\]
Taking $a$ to be each coordinate vector establishes componentwise absolute
integrability with respect to $\E_Q$. Hence
\[
h(y)=\mathbb E_Q[K(y,X)b(X)],
\qquad y\in\mathbb R^d.
\]
In particular, the pointwise expectation above is an element of
$\mathcal H_K$, because it equals the Riesz representer already constructed.

The variational definition of KSD now gives
\[
\operatorname{KSD}_K(P,Q)
=
\sup_{\|\phi\|_{\mathcal H_K}\leq1}
|\langle h,\phi\rangle_{\mathcal H_K}|
=
\|h\|_{\mathcal H_K}.
\]
The last equality follows from Cauchy--Schwarz, with equality at
$\phi=h/\|h\|_{\mathcal H_K}$ when $h\neq0$; when $h=0$, both sides vanish.

Finally, for all $x,y\in\mathbb R^d$,
\[
\begin{aligned}
|b(x)^\top K(x,y)b(y)|
&=
\left|\left\langle
K(\cdot,x)b(x),K(\cdot,y)b(y)
\right\rangle_{\mathcal H_K}\right|\leq w(x)w(y).
\end{aligned}
\]
For independent $X,Y\sim Q$, the expectation of the right-hand side equals
$C^2<\infty$. Thus Fubini's theorem applies, and
\[
\begin{aligned}
\|h\|_{\mathcal H_K}^2
&=\Lambda(h) =\mathbb E_X[b(X)^\top h(X)] =\mathbb E_{X,Y\sim Q}[b(X)^\top K(X,Y)b(Y)].
\end{aligned}
\]
Combining the last two displays proves the assertion.
\end{proof}

{An example in which $P$ does not have a smooth density $p$ is that of the Laplace distribution, for which the density is continuous, and differentiable everywhere except at 0. \cite{pike2012stein} show that for $W \sim {\rm Laplace}(0,b),$ 
$$ \E g'(W) = \frac{1}{b} \E \sgn (W) g(W) $$ for all absolutely continuous functions $g$ for which these expectations exist, where $\sgn $ is the sign function; this is the weak integration by parts formula. Thus, the score function in the weak sense is $s_b(x)  = - \sgn(x)/b$. As in the proof of Proposition \ref{prop:ksdform}, all relevant arguments are in expectation, we deduce that the conclusion of Proposition \ref{prop:ksdform} also holds for the Laplace distribution.}

\section{Proofs omitted from previous sections}
\subsection{Proofs left out in Section~\ref{sec: renyi ksd comparison}}\label{sec:RenKSDProofs}

{In order to prove Proposition~\ref{prop:renyi-stein-dual-norm}, we start by providing a stronger result that holds for more general flows $T_t$.
In the following, we denote by $J_{f_t}(x)$ the spatial Jacobian with entries $[J_{f_t}(x)]_{i,j}=\partial_{x_i}\partial_{x_j}f_t(x)$, for any differentiable function $f:\R\times \R^d \rightarrow \R^d:(t,x)\mapsto f_t(x)$.}
\begin{proposition}[First variation of R\'enyi divergence]\label{prop:renyi-first-variation}
Let $\phi:\mathbb R^d\to\mathbb R^d$ be {a $\mathcal{C}^1$ function, {let $\epsilon >0$,} and let 
the function $T:(-\epsilon,\epsilon)\times\R^d \rightarrow \R^d$ be such that
\begin{itemize}
    \item $\partial_tT_t(x)$ and $J_{\partial_tT_t}(x)$ exist and are continuous in $(t,x)$;
    \item for every $t\in (-\epsilon,\epsilon)$, the function $T_t:\R^d \rightarrow \R^d$ is a diffeomorphism;
    \item $T_0(x)=x$ for all $x\in \R^d$;
    \item $\left.{\partial_tT_t}\right|_{t=0}=\phi$.
\end{itemize}
} Define  $Q_t:=(T_t)_\#Q$ and denote its density by $q_t$. We assume in addition that
\begin{itemize}
    \item there exists a function $G\in L^1(\R^d)$ such that
    \begin{equation}\label{eq:cond1}
    p(x)^{1-\alpha} \left|\frac{q_t(x)^\alpha-q(x)^{\alpha}}{t}\right| \leq G(x)
    \end{equation}
    for all $x \in \R^d$ and $t \in (-\epsilon,\epsilon)$;
    \item we have
    \begin{equation}\label{eq:cond2}
        \int_{\R^d} q(x)^\alpha p(x)^{1-\alpha} |\phi(x)| dx < \infty;
    \end{equation}
    \item and
    \begin{equation}\label{eq:cond3}
        \int_{\R^d} q(x)^\alpha p(x)^{1-\alpha} |\phi^\top \nabla \log\left(\tfrac{p(x)}{q(x)}\right)| dx < \infty.
    \end{equation}
\end{itemize}  Then
\begin{equation}
    \frac{\mathrm d}{\mathrm dt}
    D_\alpha(Q_t\|P)
    \Big|_{t=0}=
    -\mathbb E_{Q_{\alpha\mid P}}\big[
        \mathcal T_P\phi
    \big].   
    \label{eq:renyi-first-variation}
\end{equation}
\end{proposition}
\begin{proof}
{We start by claiming that the density $q_t$ of $Q_t$ satisfies 
$
    \left.\partial_t q_t\right|_{t=0}
    =-\nabla\cdot(q\phi).
$
{We have}
$q_t(y)=q(T_t^{-1}(y)) \operatorname{det}(J_{T_t}(T_t^{-1}(y)))^{-1}$. {Thus,} 
\[
\left.\partial_t q_t(y)\right|_{t=0}
    = \left.\partial_t q(T_t^{-1}(y))\right|_{t=0} \operatorname{det}(J_{T_0}(T_0^{-1}(y)))^{-1} + q(T_0^{-1}(y)) \left.\partial_t\operatorname{det}(J_{T_t}(T_t^{-1}(y)))^{-1}\right|_{t=0}.
\]
Note that $T_0=T_0^{-1}$ is the identity function, and hence $J_{T_0}=I_d$ and $T_0^{-1}(y)=y$. It follows that
\begin{equation}\label{eq:qtId}
\left.\partial_t q_t(y)\right|_{t=0}
    = \left.\partial_t q(T_t^{-1}(y))\right|_{t=0} + q(y) \left.\partial_t\operatorname{det}(J_{T_t}(T_t^{-1}(y)))^{-1}\right|_{t=0}.
\end{equation}
By the chain rule we have that
\begin{align*}
\left.\partial_tT_t^{-1}(y)\right|_{t=0}
&=
-[J_{T_0}(T_0^{-1}(y))]^{-1}\partial_tT_t\big|_{t=0}(T_0^{-1}(y)) = -\phi(y).
\end{align*}
It follows that
\begin{equation}\label{eq:qtIdpt1}
    \left.\partial_t q(T_t^{-1}(y))\right|_{t=0} = \nabla q(T_0^{-1}(y))^\top \left.\partial_tT_t^{-1}(y)\right|_{t=0} = - \nabla q(y)^\top \phi(y).
\end{equation}
We now turn to computing $\left.\partial_t\operatorname{det}(J_{T_t}(T_t^{-1}(y)))^{-1}\right|_{t=0}$. As we did not assume second spatial derivatives for $T_t$, the quantity $\operatorname{det}(J_{T_t}(T_t^{-1}(y)))^{-1}$ might not be differentiable at every $t$, but the derivative at zero exists and can be computed. We start by noting that $J_{T_0}\equiv I_d$ and by the fundamental theorem of algebra
\[
J_{T_t}(x) = I_d + \int_0^t \partial_s J_{T_s}(x) ds.
\]
By continuity of $J_{\partial_tT_t}$, we can swap the order of integration and write
\[
J_{T_t}(x) = I_d + \int_0^t J_{\partial_s T_s}(x) ds.
\]
It follows that for any $t\neq 0$,
\begin{align*}
    \frac{J_{T_t}(T_t^{-1}(y))-I_d}{t} &= \frac{1}{t} \int_0^t J_{\partial_s T_s}(T_t^{-1}(y)) ds = \int_0^1 J_{\partial_s T_s}\big|_{s=tu}(T_t^{-1}(y)) du,
\end{align*}
where the second line follows by a change of variables. Again by continuity, we can take the limit $t\rightarrow0$ on both sides and inside the integral, to obtain
\[
\partial_t J_{T_t}(T_t^{-1}(y))\big|_{t=0} = \int_0^1 J_{\partial_sT_s|_{s=0}}(T_0^{-1}(y))du = J_\phi(y).
\]
Define $A_t := J_{T_t}(T_t^{-1}(y))$. We are thus interested in computing $\partial_t\operatorname{det}(A_t)^{-1}\big|_{t=0}$. Note that $\operatorname{det}(A_0)=\operatorname{det}(I_d)=1$ and $\operatorname{det}(A_t)$ is continuous, so for a small neighbourhood of $t=0$, we have $\operatorname{det}(A_t)\neq0$.

By the Jacobi formula, we have 
\[
\partial_t\operatorname{det}(A_t)\big|_{t=0}=\operatorname{det}(A_t) \operatorname{tr}(A_t^{-1} \partial_t A_t)
\big|_{t=0} = \operatorname{tr}(\partial_t A_t\big|_{t=0}) = \operatorname{tr}(J_\phi(y))=\operatorname{div}\phi.
\]
Now,
\begin{equation}\label{eq:qtIdpt2}
    \partial_t\operatorname{det}(J_{T_t}(T_t^{-1}(y)))^{-1}\big|_{t=0} = -\frac{\partial_t \operatorname{det}(A_t)\big|_{t=0}}{\operatorname{det}(A_0)^2} = -\operatorname{div}\phi.
\end{equation}
Combining \eqref{eq:qtId}, \eqref{eq:qtIdpt1} and \eqref{eq:qtIdpt2}, we obtain
\[
\partial_tq_t(y)\big|_{t=0} = -\nabla q(y)^\top \phi(y) + q(y) (-\operatorname{div}\phi(y)) = -\nabla \cdot (q\phi)(y),
\]
which proves the claim.

Writing
$ Z_\alpha(t)
    :=
    \int_{\R^d} q_t^\alpha(x) p^{1-\alpha}(x)\,\mathrm dx,
$ Condition \eqref{eq:cond1} allows us to differentiate under the integral sign and we obtain
\begin{align*}
    \left.\frac{\mathrm d}{\mathrm dt}Z_\alpha(t)\right|_{t=0}
    &=\int_{\R^d} \alpha q_0(x)^{\alpha-1} p(x)^{1-\alpha}\partial_t q_t(x)\big|_{t=0}\mathrm dx=-\alpha\int_{\R^d} r(x)^{\alpha-1}\nabla\cdot(q\phi)(x)\,\mathrm dx
\end{align*}
Recall that $r=q/p$ and define
\begin{align*}
    &A(x):= r(x)^{\alpha-1} \nabla\cdot (q\phi)(x);\\
    &B(x):= r(x)^\alpha p(x) \phi^\top \nabla \log r(x).
\end{align*}
Note that $\left.\frac{\mathrm d}{\mathrm dt}Z_\alpha(t)\right|_{t=0}=-\alpha \int_{\R^d} A(x) dx$, and by condition \eqref{eq:cond1}, $A \in L^1(\R^d)$. Assumption \eqref{eq:cond3} implies that $B \in L^1(\R^d)$. Now note that by the chain rule
\[
A+(\alpha-1)B = \nabla\cdot (r^\alpha p \phi),
\]
and hence $\nabla\cdot (r^\alpha p \phi) \in L^1(\R^d)$. Since assumption \eqref{eq:cond2} gives us $r^\alpha p \phi \in L^1(\R^d)$, by integration by parts {(and integration using an exhaustion by compact sets)} it follows that
\begin{equation}\label{eq:ABrel}
\int_{\R^d} A(x)+(\alpha-1)B(x) dx = \int_{\R^d} \nabla\cdot (r^\alpha p \phi)(x) dx = 0
\end{equation}

Moreover, using the chain rule again one sees that
\[
\E_{Q_{\alpha|P}} [\mathcal{T}_P\phi(x)] = \frac{1}{Z_\alpha(0)}\int_{\R^d} r(x)^\alpha p(x) (\nabla \cdot \phi(x) + \left(\nabla \log p(x))^\top \phi(x)\right) dx = \frac{1}{Z_\alpha(0)}\int_{\R^d} A(x)-B(x) dx.
\]

Since $D_\alpha(Q_t\|P)=(\alpha-1)^{-1}\log Z_\alpha(t)$,
this gives, using \eqref{eq:ABrel},
\[
\frac{\mathrm d}{{\mathrm d}t}D_\alpha(Q_t\|P)\Big|_{t=0} = -\frac{\alpha}{(\alpha-1)Z_\alpha(0)} \int_{\R^d}A(x)dx = -\frac{1}{Z_\alpha(0)}\int_{\R^d} (A(x)-B(x) ) dx = {-} \E_{Q_{\alpha|P}} [\mathcal{T}_P\phi(x)],
\]
which proves \eqref{eq:renyi-first-variation}.}
\end{proof}

\begingroup
\def\thetheorem{\ref{prop:renyi-stein-dual-norm}}
\begin{proposition}
    Assume that for every $\phi \in \Hk$, one has $\phi \in \mathcal{C}^1(\R^d;\R^d)$ and {$\sup_{x \in \R^d} \|J_\phi(x)\|_{\text{op}}<\infty$}. Assume moreover that Conditions \eqref{eq:cond1}--\eqref{eq:cond3} hold. Then
    \[
    \sup_{\|\phi\|_{\Hk}\leq 1}
    \left| \frac{\mathrm d}{\mathrm dt}       D_\alpha((I+t\phi)_\#Q\|P) \Big|_{t=0}\right|
    =\operatorname{KSD}_K(P,Q_{\alpha\mid P}).
    \]
\end{proposition}
\begin{proof}
    {This result follows immediately as soon as $\phi\in \Hk$ and $T_t=I+t\phi$ satisfy the assumptions of Proposition~\ref{prop:renyi-first-variation}. It is clear that $\partial_t T_t=\phi$ and $J_{\partial_t T_t}=J_\phi$ exist and are continuous, as well as $T_0=I$ and $\partial_tT_t\big|_{t=0}=\phi$. Conditions \eqref{eq:cond1}--\eqref{eq:cond3} follow by definition. It remains to show that for small enough $t$, the function $T_t$ is a diffeomorphism on $\R^d$. Fix $y \in \R^d$ and to note that $x\mapsto y-t\phi(x)$ is a contraction mapping, since $\sup_{x \in \R^d} \|J_\phi(x)\|_{\text{op}}<\infty$ implies that $\phi$ is Lipschitz. By the Banach fixed point theorem, there is a unique $x \in \R^d$ such that $x=y-t\phi(x)$, hence a unique $x\in \R^d$ such that $T_t(x)=y$, implying that $T_t$ is bijective. Moreover, $J_{T_t}= I_d+tJ_\phi$. Since the operator norm of $J_\phi$ is uniformly bounded, $I_d+tJ_\phi$ is invertible, and by the inverse function theorem,  $J_{T_t^{-1}}=J_{T_t}^{-1}$.}
\end{proof}

\addtocounter{theorem}{-1}
\endgroup

\subsection{Proofs left out in Subsection ~\ref{ssec: gaussian mechanism}}
\label{app: proofs of gaussian mechanism}

\begingroup
\def\thetheorem{\ref{prop:gaussianExactEscordKSD}}
\begin{proposition}[Escort KSD]
    {Under appropriate conditions (see Appendix~\ref{sec:KSDSetup})}, the KSD based on 
the kernel $K = K_{B,A}$ satisfies 
\begin{equation*}
\KSD_{K_{B,A}}(P,Q_{\alpha\vert P})^2
=
c_{B,\Sigma}\,\alpha^2\,(\Sigma^{-1}\delta)^\top A  \Sigma^{-1}\delta,
\end{equation*}
where 
\begin{equation*}
c_{B,\Sigma}
:=
\E[k_B(X,Y)]
=
\{\det(I+2\Sigma B^{-1})\}^{-1/2}.
\end{equation*}  
\end{proposition}
\addtocounter{theorem}{-1}
\endgroup
\begin{proof}
We start by showing that, for a general kernel~$K$, the KSD satisfies 
    \begin{align}
{\KSD}_K(P,Q_{\alpha\vert P})^2
&= \E_{X,Y\sim Q_{\alpha\vert P}}
\left[
\alpha^2(\Sigma^{-1}\delta)^\top
K(X,Y)
(\Sigma^{-1}\delta)
\right],
\label{eq:score-difference-ksd}
\end{align}
where $X$ and $Y$ are independent.
{We first calculate the density of $Q_{\alpha|P}$. Note that
\[
\left(\frac{q(x)}{p(x)}\right)^\alpha p(x) = \frac{1}{(2\pi)^{d/2}\operatorname{det}(\Sigma)^{1/2}} \exp\left(-\frac{1}{2}(x-m_\mathbf{D}-\alpha\delta)^\top\Sigma^{-1} (x-m_{\mathbf{D}}-\alpha\delta)\right) \exp\left(\frac{\alpha(\alpha{-}1)}{2}(\delta^\top\Sigma^{-1}\delta)\right),
\]
hence ${Q_{\alpha\vert P} = \mathcal N(m_\mathbf{D} + {{\alpha}}}\delta, {\Sigma}$).}

Furthermore, the score functions are
$s_P(x)=-\Sigma^{-1}(x-m_\mathbf{D}),$ and $s_{Q_{\alpha\vert P}}(x)=-\Sigma^{-1}(x-m_\mathbf{D}-\alpha\delta),$ 
so that
\begin{equation}
 s_P(x)-s_{Q_{\alpha\vert P}}(x)=-\alpha\Sigma^{-1}\delta
 \label{eq:constant-score-difference}
\end{equation}
for every $x\in\R^d$.   
Overall, we thus have by \eqref{eq: KSD2 as score difference} that
\begin{align*}
{\KSD}_K(P,Q_{\alpha\vert P})^2
&=\notag
\E_{X,Y\sim Q_{\alpha\vert P}}
\left[
(s_P(X)-s_{Q_{\alpha\vert P}}(X))^\top
K(X,Y)
(s_P(Y)-s_{Q_{\alpha\vert P}}(Y))
\right]\\
&= \E_{X,Y\sim Q_{\alpha\vert P}}
\left[
\alpha^2(\Sigma^{-1}\delta)^\top
K(X,Y)
(\Sigma^{-1}\delta)
\right],
\label{eq:score-difference-ksd}
\end{align*}
where $X$ and $Y$ are independent. 

Now let $K = K_{B,A}$. Using  \eqref{eq:score-difference-ksd} and \eqref{eq:constant-score-difference}, and observing that the score difference is constant, gives the second statement, where the expression of $c_{B,\Sigma}$ follows as for $X,Y\stackrel{\mathrm{iid}}\sim Q_{\alpha\vert P}$ we have $X-Y\sim N(0,2\Sigma)$.
\end{proof}

\paragraph{Isotropic kernel}
We start with the isotropic kernel~$K_{B,I_d}(x,y)=k_B(x,y)I_d.$ Then by Proposition~\ref{prop:gaussianExactEscordKSD} with $A=I_d$ we obtain
\begin{equation} \label{eq:scalar-vector-ksd}
    \KSD_{K_{B,I_d}}(P,Q_{\alpha\vert P})^2
=
c_{B,\Sigma}\,\alpha^2\,\delta^\top  \Sigma^{-2}\delta.
\end{equation}
The resulting KSD involves a quadratic weight $\alpha^2\Sigma^{-2}$ while the  R\'{e}nyi divergence involves only $\alpha\Sigma^{-1}$, see Proposition~\ref{prop:gaussian-renyi}. As the next proposition shows, this mismatch yields a looser upper bound on the R\'{e}nyi divergence.

\begin{proposition}
\label{cor:scalar-vector-rdp-comparison}
Let $\lambda_{\min}(\Sigma)$ and $\lambda_{\max}(\Sigma)$ be the extreme eigenvalues of $\Sigma$.  Then
\begin{equation*}
\frac{\lambda_{\min}(\Sigma)}{2\alpha c_{B,\Sigma}}
\KSD_{K_{B,I_d}}(P,Q_{\alpha\vert P})^2
\leq
D_\alpha(Q\|P)
\leq
\frac{\lambda_{\max}(\Sigma)}{2\alpha c_{B,\Sigma}}
\KSD_{K_{B,I_d}}(P,Q_{\alpha\vert P})^2,
\label{eq:scalar-vector-rdp-bounds}
\end{equation*}
where both inequalities become equalities whenever $\Sigma=\sigma^2I_d$ for some $\sigma \in\R_+$. 
\end{proposition}
\begin{proof}
Diagonalize $\Sigma$. Termwise,
\[
\lambda_{\min}(\Sigma)\,\delta^\top\Sigma^{-2}\delta
\leq
\delta^\top\Sigma^{-1}\delta
\leq
\lambda_{\max}(\Sigma)\,\delta^\top\Sigma^{-2}\delta.
\]
Combine this with Proposition~\ref{prop:gaussian-renyi} and
\eqref{eq:scalar-vector-ksd}.
\end{proof}

\paragraph{Anisotropic kernel}
We now consider an anisotropic kernel $K_{B,A}$ by setting $A = \Sigma$ to match the covariance structure of the privacy mechanism, i.e. $K_{B,\Sigma}(x,y):=k_B(x,y)\Sigma.$
\begingroup
\def\thetheorem{\ref{thm:covariance-matched}}
\begin{theorem}
For the kernel~$K = K_{B,\Sigma}$ we obtain,
\begin{equation*}
D_\alpha(Q\|P)
=
\frac{1}{2\alpha c_{B,\Sigma}}
\KSD_{K_{B,\Sigma}}(P,Q_{\alpha\vert P})^2 .
\end{equation*}
  {Thus, {if} $\KSD_{K}(P,Q_{\alpha\vert P})^2\leq  \epsilon$ uniformly  over all neighbourhoods $\mathbf{D}\sim\mathbf{D'}$ {then the Gaussian mechanism satisfies} 
 {$(\alpha,\frac{\epsilon}{2\alpha c_{B,\Sigma}})$-RDP.} }
\end{theorem}
\addtocounter{theorem}{-1}
\endgroup
\begin{proof}
By Proposition~\ref{prop:gaussianExactEscordKSD},
\[
\KSD_{K_{B,\Sigma}}(P,Q_{\alpha\vert P})^2
=\alpha^2
(\Sigma^{-1}\delta)^\top
\E[k_B(X,Y)\Sigma]
(\Sigma^{-1}\delta) = \alpha^2\E[k_B(X,Y)]
\delta^\top \Sigma^{-1}\delta
\]
and thus 
\begin{equation*}
\KSD_{K_{B,\Sigma}}(P,Q_{\alpha\vert P})^2
=
c_{B,\Sigma}\,\alpha^2\,\delta^\top\Sigma^{-1}\delta.
\end{equation*}
A comparison with Proposition~\ref{prop:gaussian-renyi} gives the result.
\end{proof}

\subsection{Proofs left out in Subsection ~\ref{ssec: laplace mechanism}} \label{app: proofs of laplace mechanism}
\begingroup
\def\thetheorem{\ref{prop:laplace-renyi}}
\begin{proposition}[R\'{e}nyi divergence]
For $P$, $Q$ the Laplace output distributions of a neighbouring dataset pair $(\mathbf{D},\mathbf{D'})$,
\begin{equation*}
   {
    D_\alpha(Q\|P)
    =D_\alpha(P\|Q)
    =R_{\alpha}\bigg(\frac{\delta}{b}\bigg).
    }
\end{equation*}
In particular, $ D_1(Q\|P)=\frac{\delta}{b}+e^{-\frac{\delta}{b}}-1$, and $D_\infty(Q\|P)=\frac{\delta}{b}.$
\end{proposition}
\addtocounter{theorem}{-1}
\endgroup
\begin{proof}
By translation invariance, take $m_\mathbf{D}=0$ and define $Z_{\alpha,\delta,b}:= \int_{\mathbb R} q(x)^\alpha p(x)^{1-\alpha}\,\mathrm dx$. {Let $q_{\delta}$ be the density of $Q_{\delta} = \text{Lap}(\delta,b)$ and note that as $m_\mathbf{D} = 0$ we have $Q = Q_{\delta}$. Furthermore, $q_{\delta}(x) = \frac{1}{2b}\exp\left(-\frac{|x-\delta|}{b}\right) = q_{-\delta}(-x)$ while $p(x) = q_0(x) = p(-x)$ which implies that 
$D_\alpha(Q\|P) = D_\alpha(Q_\delta\|P) = D_\alpha(Q_{-\delta}\|P)$. In particular, we may wlog assume that $\delta >0$.}
Splitting the integral over $(-\infty,0)$, $(0,\delta)$, and $(\delta,\infty)$ gives
\begin{equation}
   Z_{\alpha,\delta,b} 
    =
    \frac{
        \alpha e^{(\alpha-1)\frac{\delta}{b}}
        +(\alpha-1)e^{-\alpha\frac{\delta}{b}}
    }{2\alpha-1}.
    \label{eq:laplace-renyi-normalizer}
\end{equation}
Hence the first result follows as 
$D_\alpha(Q\|P)=(\alpha-1)^{-1}\log Z_{\alpha,\delta,b}$.
The change of variables $x\mapsto \delta-x$ interchanges $p$ and $q$, proving
symmetry for this particular location family.  The limits
$\alpha\downarrow1$ and $\alpha\to\infty$ give the remaining results; see also \cite[Proposition~6]{mironov2017renyi}.
\end{proof}
\begin{remark}
If $B\to\infty$ in every direction, then $k_B\to1$ and
$c_{B,\Sigma}\to1$. The theorem then reduces to the constant covariance-matched
kernel
$K_\Sigma(x,y)=\Sigma,$  
for which we have the equality 
\begin{equation*}
    D_\alpha(Q\|P)=\frac{1}{2\alpha}\KSD_{K_\Sigma}(P,Q_{\alpha\vert P})^2 = \frac{\alpha}{2}\KSD_{K_\Sigma}(P,Q)^2.
\end{equation*}
In particular, the expression for the R\'{e}nyi divergence can be already recovered with a constant operator-valued kernel. 
\end{remark}

\begingroup
\def\thetheorem{\ref{prop:laplace-general-kernel-ksd}}
\begin{proposition}[Escort KSD]
Under the {conditions detailed in Appendix~\ref{sec:KSDSetup}}, the KSD based on the scalar kernel~$k$ satisfies 
\begin{align*}
{\KSD}_K(P,Q_{\alpha\vert P})^2
&= \frac{4\alpha^2 }{b^2}
    \mathbb E_{X,Y\sim Q_{\alpha\vert P}}\left[
        k(X,Y)
        \mathbf 1_{\{X \in I_\delta\}}
        \mathbf 1_{\{Y \in I_\delta\}}
     \right],
\end{align*}
where $X$ and $Y$ are independent {and $I_{\delta} := (\min\{m_\mathbf{D},m_\mathbf{D}+\delta\},\max\{m_\mathbf{D},m_\mathbf{D}+\delta\}$}.
\end{proposition}
\addtocounter{theorem}{-1}
\endgroup
\begin{proof} 
The escort $Q_{\alpha\vert P}$ has density
$$\frac{\mathrm dQ_{\alpha\vert P}}{\mathrm dx}
    =
    \frac{q(x)^\alpha p(x)^{1-\alpha}}
    {Z_{\alpha,\delta,b}},
$$ where the normalization constant $Z_{\alpha,\delta,b}$ is given in
\eqref{eq:laplace-renyi-normalizer}.    
Furthermore, the corresponding score functions are defined almost everywhere and given by
$$ s_P(x)=-\frac{1}{b}\sgn(x-m_\mathbf{D})$$ and $${s_{Q_{\alpha\vert P}}(x)
    =-\frac{(1-\alpha)}{b}\sgn(x-m_\mathbf{D}) -\frac{\alpha}{b}\sgn(x-(m_\mathbf{D}+\delta))= (1-\alpha)s_P(x)+\alpha s_{Q}(x),}$$ 
so that 
\begin{equation}
    s_P(x)-s_{Q_{\alpha\vert P}}(x)
    =\alpha(s_P(x)-s_{Q}(x))
    =-\frac{2\alpha}{b}
    \mathbf \mathbf {\sgn(\delta)\,1_{\{x \in I_\delta\}}} 
    \label{eq:laplace-escort-score-difference}
\end{equation}
for almost every $x\in\R$. Overall, we thus obtain by \eqref{eq: KSD2 as score difference} that
\begin{align*}
{\KSD}_K(P,Q_{\alpha\vert P})^2
&=
\E_{X,Y\sim Q_{\alpha\vert P}}
\left[
(s_P(X)-s_{Q_{\alpha\vert P}}(X))^\top
K(X,Y)
(s_P(Y)-s_{Q_{\alpha\vert P}}(Y))
\right] \notag\\
&= \frac{4\alpha^2 }{b^2}
    \mathbb E_{X,Y\sim Q_{\alpha\vert P}}\left[
        k(X,Y)
        \mathbf 1_{\{X \in I_\delta\}}
        \mathbf 1_{\{Y \in I_\delta\}}
     \right],
\end{align*}
where $X$ and $Y$ are independent. 
\end{proof}

\paragraph{Gaussian RBF kernel} 
We start with the {\it normalized Gaussian RBF kernel}  given by
\begin{equation}
    k_h(x,y)
    :=\exp\left(-\frac{(x-y)^2}{2h^2}\right).
    \label{eq:laplace-rbf-kernel}
\end{equation}
Furthermore, for $\rho:=\frac{h}{b}$ define 
\begin{equation}
    J_{\alpha,\rho}(u)
    := \frac{(2\alpha -1)^2}{C_{\alpha}(u)^2} 
    \int_0^u \int_0^u e^{-(2\alpha-1)(t+s)}
    e^{-(t-s)^2/(2\rho^2)}\,\mathrm dt \,\mathrm ds,
    \label{eq:J-rho-integral}
\end{equation}
where ${C_{\alpha}(u) := \alpha + (\alpha -1) e^{-(2\alpha -1)u}}$.

\begin{proposition}
\label{prop:laplace-rbf-ksd}
For $k_h$ the RBF kernel {in \eqref{eq:laplace-rbf-kernel}} we obtain 
\begin{equation*}
    {
    \operatorname{KSD}_{k_h}(P,Q_{\alpha\vert P})^2
    =\frac{\alpha^2 }{b^2} J_{\alpha,\rho}\bigg(\frac{\abs{\delta}}{b}\bigg).
    }
\end{equation*}
{
Whenever $b^2\operatorname{KSD}_{k_h}(P,Q_{\alpha\vert P})^2 \in [0,J_{\alpha,\rho} (\infty))\subsetneqq [0,\infty)$, we thus have
\begin{equation*}
    D_\alpha(Q\|P)
    =
    R_\alpha\left(
        J_{\alpha,\rho}^{-1}\left(
            {\frac{b^2}{\alpha^2}} \operatorname{KSD}_{k_h}(P,Q_{\alpha\vert P})^2
        \right)
    \right).
\end{equation*}}
\end{proposition}
\begin{proof}
{First, note that the RBF kernel fulfils $k(x,y) = k(-x,-y)$. As $$x\in I_{\delta} \iff -x\in I_{-\delta}$$ the expression in Proposition~\ref{prop:laplace-general-kernel-ksd} is invariant under a sign change of $\delta$. In particular, we may assume wlog that $\delta >0$.}

Let $x\in (m_\mathbf{D},m_\mathbf{D}+{\delta})$. Then the escort density is given by 
\begin{align*}
    \frac{\mathrm d Q_{\alpha\vert P}}{\mathrm dx}(x) 
    = \frac{1}{2b} \frac{1}{Z_{\alpha,\delta,b}} e^{-\alpha\frac{\abs{x-m_\mathbf{D}-\delta}}{b}} e^{-(1-\alpha)\frac{\abs{x-m_\mathbf{D}}}{b}}
    = \frac{1}{2b} \frac{1}{Z_{\alpha,\delta,b}} e^{(2\alpha-1)\frac{x-m_\mathbf{D}}{b}} e^{-\frac{\delta}{b}}.
\end{align*}  
Applying this to Proposition~\ref{prop:laplace-general-kernel-ksd}  and changing variables to $u=(x-m_\mathbf{D})/b$ and $v=(y-m_\mathbf{D})/b$ gives
\begin{align*}
   {b^2} \operatorname{KSD}_{k_h}(P,Q_{\alpha\vert P})^2
   &=\frac{\alpha^2}{b^2 \,Z_{\alpha,\delta,b}^2} e^{-2\alpha \frac{\delta}{b}} 
    \int_0^\frac{\delta}{b}\int_0^\frac{\delta}{b}
        e^{(2\alpha -1)(u+v)}
        e^{-(u-v)^2/(2\rho^2)}
    \,\mathrm du\,\mathrm dv.
\end{align*}
Now note that $$Z_{\alpha,\delta,b} = \frac{e^{(\alpha -1)\frac{\delta}{b}}}{2\alpha -1} (\alpha + (\alpha -1) e^{-(2\alpha -1)\frac{\delta}{b}}) = \frac{e^{(\alpha -1)\frac{\delta}{b}}}{2\alpha -1}  C_{\alpha}\left(\frac{\delta}{b}\right).$$  Then substituting $s = u-\frac{\delta}{b}$ and $t = v-\frac{\delta}{b}$ in the above equality yields 
\begin{align*}
   {b^2} \operatorname{KSD}_{k_h}(P,Q_{\alpha\vert P})^2
   &=\frac{\alpha^2 (2\alpha -1)^2}{b^2 \,C_{\alpha}\big(\frac{\delta}{b}\big)^2}  
    \int_0^\frac{\delta}{b}\int_0^\frac{\delta}{b}
        e^{-(2\alpha -1)(t+s)}
        e^{-(t-s)^2/(2\rho^2)}
    \,\mathrm ds\,\mathrm dt.
\end{align*}

To obtain the second statement, we analyze the function $J_{\alpha,\rho}\big(\frac{\abs{\delta}}{b}\big)$. As $\alpha >1$, the double integral  is increasing while $C_{\alpha}$ is decreasing. Therefore, $J_{\alpha,\rho}$ is increasing as $\frac{\abs{\delta}}{b}\to\infty$. Furthermore, the function starts at  $J_{\alpha,\rho}(0)  = 0$ and saturates at $$J_{\alpha,\rho}(\infty)    =\frac{\sqrt{2\pi}\,\rho\,(2\alpha-1)}{\alpha^2} \exp\left(\frac{(2\alpha-1)^2\rho^2}{2} \right) \left[1-\Phi\left((2\alpha-1)\rho\right)\right] < \infty,$$
    where $\Phi$ denotes the standard normal distribution function.
Consequently $J_{\alpha,\rho}$ has an inverse from
$[0,J_{\alpha,\rho}(\infty))$ to $[0,\infty)$.
\end{proof}

The inversion above does not yield a useful global linear KSD bound on Rényi divergence. Indeed, $J_{\alpha,\rho}$ is bounded with $J_{\alpha,\rho}(\infty)<\infty $ whereas $R_{\alpha}\big(\frac{\abs{\delta}}{b}\big)\to\infty$ as
$\frac{\abs{\delta}}{b}\to\infty$. Thus, the KSD based on the RBF kernel loses sensitivity to increasingly large displacements even though the Rényi divergence continues to grow. 
{
Thus, while the RBF kernel captures the changes of the R\'{e}nyi divergence well locally, a different kernel choice is needed to obtain global sensitivity.} 

\paragraph{Constant kernel}
\label{subsec:laplace-constant-kernel}

Consider the rank-one kernel
$
    k_0(x,y):=1.
$ 

\begingroup
\def\thetheorem{\ref{thm:laplace-escort-derivative}}
\begin{theorem}
For $k_0(x,y)\equiv 1$,
\begin{equation*}
    {\operatorname{KSD}_{k_0}(P,Q_{\alpha\vert P})
    =\frac{\alpha}{b}\frac{e^{(\alpha-1)\frac{\abs{\delta}}{b}}-e^{-\alpha\frac{\abs{\delta}}{b}}}{       \alpha e^{(\alpha-1)\frac{\abs{\delta}}{b}}+(\alpha-1)e^{-\alpha\frac{\abs{\delta}}{b}}}.}
\end{equation*}
Set $\kappa_\alpha:=b\operatorname{KSD}_{k_0}(P,Q_{\alpha\vert P})
    \in[0,1).$ Then 
\begin{equation}
  {
    \begin{aligned}
    D_\alpha(Q\|P) = R_{\alpha}\bigg(\frac{\abs{\delta}}{b}\bigg)
    ={}&
    \frac{1}{2\alpha-1}
    \log\frac{1}{1-\kappa_\alpha}
    -
    \frac{\alpha}{(\alpha-1)(2\alpha-1)}
    \log\left(
        1+\frac{\alpha-1}{\alpha}\kappa_\alpha
    \right).
    \end{aligned}
    }
    \tag{\ref{eq:laplace-renyi-from-escort-kappa}}
\end{equation}
\end{theorem}
\addtocounter{theorem}{-1}
\endgroup
\begin{proof}
{
As in Proposition~\ref{prop:laplace-rbf-ksd}, we may assume wlog that $\delta > 0$ as the constant kernel fulfils $k_0(x,y) = k_0(-x,-y)$.}

The escort probability of the interval between the modes is
\begin{equation*}
    {\P_{X\sim Q_{\alpha\vert P}}}(m_\mathbf{D}<X<m_\mathbf{D}+\delta)
    =
    \frac{
        e^{(\alpha-1)\frac{\delta}{b}}-e^{-\alpha\frac{\delta}{b}}
    }{
        2\left(
            \alpha e^{(\alpha-1)\frac{\delta}{b}}
            +(\alpha-1)e^{-\alpha\frac{\delta}{b}}
        \right)
    }.
    \label{eq:laplace-escort-interval-probability}
\end{equation*}
Combining this with Proposition~\ref{prop:laplace-general-kernel-ksd} gives
the first equality. 
By further applying this equality to $\kappa_\alpha:=b\operatorname{KSD}_{k_0}(P,Q_{\alpha\vert P})$ and solving for $\frac{\delta}{b}$ yields
$$\frac{\delta}{b}=\frac{1}{2\alpha-1}\log\left[\frac{\alpha+(\alpha-1)\kappa_\alpha}{\alpha(1-\kappa_\alpha)}\right].$$
 Finally, applying Proposition~\ref{prop:laplace-renyi} yields the second equality.
\end{proof}

\begin{remark}\label{rem: laplace-renyi-path-integral}
{As $k_0(x,y) = k_0(-x,-y)$, we may assume wlog that $\delta >0$, see the proof of Theorem~\ref{thm:laplace-escort-derivative}.}
Differentiating $\frac{1}{b}R_{\alpha}(u)$ and evaluating at the point $u= \frac{{t}}{b}$ yields the right hand side of \eqref{eq:laplace-renyi-from-escort-kappa} and thus for $Q_t = P_{m_\mathbf{D}+t}$, $t\in[0,\delta]$, we obtain
\begin{equation*}
    {
    \frac{\mathrm d}{\mathrm dt}D_\alpha(Q_t\|P) = \frac{\mathrm d}{\mathrm d u} \frac{1}{b} R_{\alpha}(u)\bigg\vert_{u=\frac{{t}}{b}}
    =
    \operatorname{KSD}_{k_0}(P,(Q_t)_{\alpha\vert P}).
    }
\end{equation*}
Consequently, integration from $0$ to ${\delta}$ gives
\begin{equation*}
    {
    D_\alpha(Q\|P) = D_\alpha(Q_{{{\delta}}}\|P) - D_\alpha(Q_0\|P)
    =
    \int_0^{{{\delta}}}
    \operatorname{KSD}_{k_0}(P,(Q_t)_{\alpha\vert P})\,\mathrm dt.
    }
\end{equation*}
In particular, the first identity is a realization of Proposition~\ref{prop:renyi-first-variation} while the second identity shows a global version thereof.
\end{remark}

Although both kernel choices $k_h$ and $k_0$ yield a bounded KSD, the constant kernel~$k_0$ provides an explicit relation between the KSD and the displacement~$\delta$ that can be inverted analytically. This allows the R\'{e}nyi divergence to be recovered from the KSD for all $\delta\geq 0$. For the RBF kernel $k_h$ such a relation exists only locally.

\subsection{Proofs left out in Subsection ~\ref{ssec: exponential mechanism}} \label{app: proofs of exponential mechanism}
We start with a detailed introduction to the exponential mechanism. 

Let $(\mathcal Y,\mathcal F,\nu)$ be a measurable output space equipped with a fixed, data-independent base measure~$\nu$. {Assume that $\mathcal Y\subseteq\mathbb R^d$ and that
$ \nu(\mathrm dy)=h_0(y)\,\mathrm dy$, with $h_0(y)>0$, 
where $h_0$ is smooth and independent of the dataset.} We remark that the need to specify the base measure is essential on non-Euclidean and infinite-dimensional output spaces; see \cite{awan2019exponential}.  
For a dataset $\mathbf{D}\in\mathfrak{D}$, let $u_\mathbf{D}:\mathcal Y\to\mathbb R$ be a measurable utility and let $\beta>0$ be an inverse-temperature parameter.  Define the exponential mechanism by 
\begin{equation}
    P_\mathbf{D}(\mathrm dy)
    =p_\mathbf{D}(y)\,\nu(\mathrm dy),
    \qquad
    p_\mathbf{D}(y)
    :=\frac{\exp\{\beta u_\mathbf{D}(y)\}}{\exp(A(u_0))},
    \label{eq:exp-general-mechanism-definition}
\end{equation}
where $A$ is the (general) log-partition function defined by
\begin{equation}
    A(u_t)
    :=
    \log\int_{\mathcal Y}
        \exp\left\{\beta\bigl(u_\mathbf{D}(y)+t\delta (u)(y)\bigr)\right\}
    \,\nu(\mathrm dy).
    \label{eq:exp-general-A-definition}
\end{equation}
We assume throughout that every partition function appearing below is finite.

We now fix an ordered neighbouring pair $(\mathbf{D},\mathbf{D'})$ and write $
    P:=P_\mathbf{D}, 
    Q:=P_{\mathbf{D'}},$ and $
    \delta (u):=u_{\mathbf{D'}}-u_\mathbf{D}.$ 
Then
\begin{equation*}
    \frac{\mathrm dQ}{\mathrm dP}(y)
    =
    \frac{\exp\{\beta\delta (u)(y)\}}
         {\mathbb E_P[\exp\{\beta\delta (u)(Y)\}]}.
    \label{eq:exp-likelihood-ratio}
\end{equation*}

In the following we focus on the regular natural exponential family that can be obtained by setting $\beta = 1$ and $u_{\mathbf{D}} (y) = u_{\eta_\mathbf{D}} (y) :=\eta_\mathbf{D}^{\mathsf T}T(y)$ where $\eta_\mathbf{D} \in \Eta$ is the natural parameter corresponding to the dataset~$\mathbf{D}$, $\Eta\subset \R^r$ is an open set and $T$ is a sufficient statistic~$
    T: \R^d \rightarrow R^r 
$ that is {assumed to be differentiable}. 
We denote ${P}_{\eta}(\dd y) = p_{\eta}(y) \nu(\dd y) = p_{\eta}(y) h_0(y) \dd y$ for
\begin{equation}
    p_\eta(y) 
    :=\exp\{\eta^{\mathsf T}T(y)-A(\eta)\},
    \label{eq:regular-exponential-family}
\end{equation}  
where $A$ is again the log-partition function defined in \eqref{eq:exp-general-A-definition} and, with a slight abuse of notation, we write  $A(\eta_t) := A((u_{\eta})_t)$.
Then {the Fisher information matrix is $I(\eta_t) = \nabla^2 A(\eta_t)$. We assume that the exponential family is minimal, so that $I(\eta) $ is positive definite for every $\eta \in \Eta$.}   

For the fixed ordered neighbouring pair $(\mathbf{D},\mathbf{D'})$ set $\eta = \eta_\mathbf{D}$, $\eta' = \eta_{\mathbf{D'}}$ and $\delta = \eta_{\mathbf{D'}} - \eta_\mathbf{D}$, and define $\eta_t := \eta + t \delta$. Then $\delta (u)(y) = \delta ^{\mathsf T}T(y)$. Finally, we assume that $\eta_t \in \Eta$ for any $t \in [0, \alpha].$

{We now prove an expanded version of Proposition \ref{prop:exp-exact-renyi}.}
\begingroup
\def\thetheorem{\ref{prop:exp-exact-renyi}}
\begin{proposition}[R\'enyi divergence]
Let $P = P_{\eta}$ be a member of the natural exponential family defined above.
Then for every $\alpha>1$ for which $A(u_\alpha)<\infty$,
\begin{equation}
   {
    D_\alpha(Q\|P)
    =
 \frac{
      A(\eta_\alpha)-\alpha A(\eta_1)+(\alpha-1)A(\eta_0)
  }{\alpha-1} 
    = \frac{\alpha}{2} \delta^\top \bar{I}((\eta_t)_{t=0}^\alpha) \delta
    }
    \tag{\ref{eq:exp-renyi-log-partition2}}
\end{equation}
with $  \bar{I}((\eta_t)_{t=0}^\alpha):= \Big\{  \frac{2}{\alpha} \int_0^1 t  I(\eta_t) \, dt + \frac{2}{\alpha(\alpha-1)} \int_1^\alpha (\alpha -t)  I(  \eta_t )  \, dt \Big\}.$
\end{proposition}
\addtocounter{theorem}{-1}
\endgroup

{The scaling in \eqref{eq:exp-renyi-log-partition2} is motivated by $\int_0^1 t \,dt + \frac{1}{\alpha-1} \int_1^\alpha (\alpha -t) \,dt = \frac{\alpha}{2}.$}

\begin{proof}
{The first equality is a standard result for R\'{e}nyi divergence within the natural exponential family and can, for example, be found in {(15) of \cite{thierrin2022renyi} or in} 
\cite[Appendix~I]{wang2019subsampled}. {For completeness we} include a proof below, together with the second representation in terms of the Fisher information.}
Using \eqref{eq:exp-general-mechanism-definition},
\begin{align*}
    \int \left(\frac{\mathrm dQ}{\mathrm dP}\right)^\alpha
    \,\mathrm dP
    &={}
    \int
    p_{\mathbf{D'}}(y)^\alpha p_\mathbf{D}(y)^{1-\alpha}
    \,\nu(\mathrm dy) ={}
    \exp\left\{
        A(u_\alpha)-\alpha A(u_1)+(\alpha-1)A(u_0)
    \right\}.
\end{align*}
Taking $(\alpha-1)^{-1}$ times the logarithm gives the first equality in 
\eqref{eq:exp-renyi-log-partition}. 

Now let $P = P_{\eta}$ be a member of the natural exponential family. Then 
\begin{align*}
D_\alpha(Q\|P)
   &  =
    \frac{
        A(\eta_\alpha)
        -\alpha A(\eta_1)
        +(\alpha-1)A(\eta_0)
    }{\alpha-1} \nonumber \\
    &= {[t \delta^\top \nabla A(\eta_t)]_0^1 - \int_0^1 \delta^\top\nabla A(\eta_t) \dd t + \frac{1}{\alpha -1 } \bigg([(\alpha -t) \delta^\top\nabla A(\eta_t)]_1^\alpha {+}  \int_1^\alpha   \delta^\top\nabla A(\eta_t) \dd t\bigg)}\\
    &= {\int_0^1 t \delta^\top\nabla^2 A(\eta_t)\delta  \dd t + \frac{1}{\alpha -1 }  \int_1^\alpha (\alpha - t)\delta^\top\nabla^2 A(\eta_t)\delta \dd t}\\
    &= \int_0^1 t \delta^\top I( \eta_t) \delta \, dt + \frac{1}{\alpha-1} \int_1^\alpha (\alpha -t) \delta^\top I( \eta_t) \delta \, dt \nonumber \\
    &= \frac{\alpha}{2} \delta^\top \Big\{  \frac{2}{\alpha} \int_0^1 t  I(\eta_t) \, dt + \frac{2}{\alpha(\alpha-1)} \int_1^\alpha (\alpha -t)  I(  \eta_t )  \, dt \Big\} \delta.
\end{align*}

\end{proof}

\begingroup
\def\thetheorem{\ref{prop:exp-exact-escort-ksd}}
\begin{proposition}(Escort KSD)

Let $P = P_{\eta}$ be a member of the natural exponential family {\eqref{eq:exp-mechanism-definition}}.
Under the {conditions laid out in Appendix~\ref{sec:KSDSetup}} 
\begin{equation*}
   {
    \operatorname{KSD}_K(
        P,Q_{\alpha\vert P}
    )^2
    =
    \alpha^2\,\delta^{\mathsf T}
    G_K(\eta_\alpha)
    \delta.
    }
\end{equation*}
\end{proposition}
\addtocounter{theorem}{-1}
\endgroup
\begin{proof}
First, note that the escort $Q_{\alpha\vert P}$ is given by
\begin{equation*}
    {
    Q_{\alpha\vert P}(\mathrm dy)
    \propto
    \exp\left\{
        \beta\bigl[(1-\alpha)u_\mathbf{D}(y)+\alpha u_{\mathbf{D'}}(y)\bigr]
    \right\}\nu(\mathrm dy)
    }
\end{equation*}
and can thus be viewed as a member of the exponential family with function $$    u_{\alpha,\mathbf{D},\mathbf{D'}}(y):=u_\mathbf{D}(y)+\alpha(u_{\mathbf{D'}}(y)-u_\mathbf{D}(y))$$ and base measure $ \nu(\mathrm dy)=h_0(y)\,\mathrm dy$, i.e. $ Q_{\alpha\vert P} = P_{u_{\alpha,\mathbf{D},\mathbf{D'}}}$. 
Furthermore, the corresponding score functions are given by
\begin{equation*}
    s_P(y)
    =\beta\nabla u_\mathbf{D}(y)+\nabla\log h_0(y) \quad\text{ and }\quad s_{Q_{\alpha\vert P}}(y)=\beta\nabla u_{\alpha,\mathbf{D},\mathbf{D'}}(y)+\nabla\log h_0(y).
    \label{eq:exp-score}
\end{equation*}
Thus, the base-measure contribution cancels in the score function comparison and we obtain
$$
   { s_P(y)-s_{Q_{\alpha\vert P}}(y)
    =-\alpha\,\beta\nabla\delta (u)(y).}
$$ 
Overall, we thus have by \eqref{eq: KSD2 as score difference} that
\begin{align*}
    \operatorname{KSD}_K(P,Q_{\alpha\vert P})^2
    &= \E_{X,Y\sim Q_{\alpha\vert P}} \left[(s_P(X)-s_{Q_{\alpha\vert P}}(X))^\top K(X,Y)(s_P(Y)-s_{Q_{\alpha\vert P}}(Y))\right] \notag\\
    &=\alpha^2\,\beta^2\mathbb E_{X,Y\sim Q_{\alpha\vert P}}
    \left[
        \nabla\delta (u)(X)^{\mathsf T}
        K(X,Y)
        \nabla\delta (u)(Y)
    \right].
\end{align*}
Now let $P = P_{\eta}$. Then $\nabla \delta (u) = J_T(y)^{\mathsf T}\delta$, where $J_T(y)\in\mathbb R^{r\times d}$ is defined as the Jacobian of $T$. 
Thus, as  $\beta = 1$ and $Q_{\alpha\vert P} = P_{\eta_\alpha}$,
\begin{equation*}
   {
    \operatorname{KSD}_K(
        P,Q_{\alpha\vert P}
    )^2
    =
     \alpha^2\,\delta^{\mathsf T}
    G_K(\eta_\alpha)
    \delta.
    }
\end{equation*}
\end{proof}

\begingroup
\def\thetheorem{\ref{thm:exp-escort-ksd-certificate}}
\begin{theorem}
Under Assumptions~ \ref{ass: restrictedpoincare} and \ref{ass: coercivity} we have
    \begin{align}\tag{\ref{eq:diffbound}}
     D_\alpha (Q  || P) \le  \frac{C_\Eta}{2 \alpha  \kappa_\Eta} \KSD_{K} (P , Q_{\alpha | P})^2.
\end{align}
{Thus, {if} $\KSD_{K}(P,Q_{\alpha\vert P})^2\leq  \epsilon$ uniformly  over all neighbourhoods $\mathbf{D}\sim\mathbf{D'}$ 
{then the exponential mechanism satisfies} $(\alpha,\frac{ C_\Eta}{2\alpha \, \kappa_\Eta}\epsilon)$-RDP.} 
\end{theorem}
\addtocounter{theorem}{-1}
\endgroup
\begin{proof}
As $T$ is a sufficient statistic, we have that for any $\eta\in\Eta$ the Fisher information fulfils the identity
$$I(\eta) = \nabla^2 A(\eta) = \text{Cov}_{P_{\eta}}(T(X)),$$
see also \cite[Corollary 3.1]{wainwright2008graphical}. In particular, 
$$\delta^\top I(\eta) \delta  =\delta^\top \text{Cov}_{P_{\eta}}(T(X)) \delta =  \text{Cov}_{P_{\eta}}(\delta^\top T(X)) = \text{Var}_{P_{\eta}}(\delta^\top T(X)).$$
    Thus, by Proposition~\ref{prop:exp-exact-renyi} we have
    \begin{align*}
    D_\alpha(Q\|P)
    &= \frac{\alpha}{2} \delta^\top \Big\{  \frac{2}{\alpha} \int_0^1 t  I(\eta_t) \, dt + \frac{2}{\alpha(\alpha-1)} \int_1^\alpha (\alpha -t)  I(  \eta_t )  \, dt \Big\} \delta\\
    &= \frac{\alpha}{2}  \Big\{  \frac{2}{\alpha} \int_0^1 t  \Var_{P_{\eta_t}}\left(\delta^\top T(X)\right) \, dt + \frac{2}{\alpha(\alpha-1)} \int_1^\alpha (\alpha -t)  \Var_{P_{\eta_t}}\left(\delta^\top T(X)\right)  \, dt \Big\} \\
    &\leq \frac{\alpha\, C_\Eta}{2} \Big\{  \frac{2}{\alpha} \int_0^1 t  \E_{P_{\eta_t}}\left\|J_T(X)^\top \delta\right\|^2 \, dt + \frac{2}{\alpha(\alpha-1)} \int_1^\alpha (\alpha -t)  \E_{P_{\eta_t}}\left\|J_T(X)^\top \delta\right\|^2  \, dt \Big\}\\
    &= \frac{\alpha\, C_\Eta}{2} \bar{J}((\eta_t)_{t=0}^\alpha)\\
    &\leq \frac{\alpha\, C_\Eta }{2 \, \kappa_\Eta}\delta^\top G_K(\eta_\alpha) \delta\\
    &= \frac{C_\Eta}{2 \alpha\,\kappa_\Eta}\KSD_{K} (P , Q_{\alpha | P})^2
\end{align*}
where the two inequalities are a consequence of Assumption \ref{ass: restrictedpoincare} and \ref{ass: coercivity}, while the last equality follows from Propostition~\ref{prop:exp-exact-escort-ksd}.
\end{proof}

\begingroup
\def\thetheorem{\ref{cor:exp-laplace}}
\begin{corollary}[Laplace family]
Let $P_\eta =\operatorname{Lap}(0,-\eta^{-1}) $, $\eta \in \Eta\subset \R_{<0}$, be the centered Laplace family, written in natural form as
$p_{\eta}(x) = \exp(\eta \abs{x} -A(\eta)) = -\frac{\eta}{2} e^{\eta \abs{x}}$ with sufficient statistic~$T(x) = \abs{x}$ and 
{Lebesgue} 
base measure.
Assume that $\Eta$ is chosen such that $\eta_{\min}\leq \eta_t\leq \eta_{\max}<0$ for all $t\in[0,\alpha]$ and let $K(x,y) = k(x,y) = xy$.
    Then $\KSD_{K} (P , Q_{\alpha | P})^2 \leq \epsilon$ uniformly over all $\mathbf{D}\sim\mathbf{D'}$ implies $(\alpha,\frac{\eta_{\min}^2}{2\alpha \eta_{\max}^2}\epsilon )$-RDP.
\end{corollary}
\addtocounter{theorem}{-1}
\endgroup
\begin{proof}
We study $P = P_\eta = \operatorname{Lap}(0,-\eta^{-1})$,  $\eta \in \Eta\subset \R_{<0}$, which belongs to the regular natural exponential family with sufficient statistic $T(x)=|x|$ and Lebesgue base measure~$\nu(\dd x) = \dd x$. Then $p_{\eta}(x) = \exp(\eta \abs{x} -A(\eta)) = -\frac{\eta}{2} e^{\eta \abs{x}}$. We remark that this construction is different to the Laplacian mechanism discussed in Subsection ~\ref{ssec: laplace mechanism}. 

We show that Assumptions~\ref{ass: restrictedpoincare} and \ref{ass: coercivity} are satisfied with $C_\Eta = \frac{{1}}{\eta^2_{\max}} $ and $\kappa_\Eta = \frac{1}{\eta^2_{\min}}$. Consequently, an application of Theorem~\ref{thm:exp-escort-ksd-certificate} yields
    $$ D_\alpha (Q  || P) \le  \frac{\eta_{\min}^2}{2\alpha \eta_{\max}^2} \KSD_{K} (P , Q_{\alpha | P})^2$$
    and the claim follows.
    
As $P_\lambda$ belongs to the Laplace family, {and $T(x)= |x|$, we have for every $\delta$ that $\Var_{P_\eta} (\delta |X|) = \frac{\delta^2}{\eta^2} $ and $\E_{P_\eta} [(J_T(X) \delta)^2] = \delta^2.$ Hence} 
Assumption~\ref{ass: restrictedpoincare} holds uniformly with $C_\Eta=\frac{{1}}{\eta_{\max}^2}.$

For  Assumption~\ref{ass: coercivity}, consider the positive semidefinite kernel~$k(x,y)=xy.$ Since $J_T(x)=\sgn(x)$ almost everywhere, we have 
\begin{align*}
    G_K(\eta_{\alpha}) 
    &=\E_{X,Y\stackrel{\mathrm{iid}}{\sim}P_{\eta_\alpha}}\left[\sgn(X)XY\sgn(Y)\right] =
    \left(\E_{P_{\eta_\alpha}}|X|\right)^2=\frac{1}{{\eta_\alpha}^2}.
\end{align*}
On the other hand, $\E_{P_{\eta_\alpha}}\left[|J_T(X)\delta|^2\right]=\delta^2$ for every $t\in[0,\alpha]$
and thus
\begin{align*}
    \bar{J}((\eta_t)_{t=0}^\alpha) &= \Big\{  \frac{2}{\alpha} \int_0^1 t  \delta^2 \, dt + \frac{2}{\alpha(\alpha-1)} \int_1^\alpha (\alpha -t) \delta^2 \, dt \Big\}= \delta^2.
\end{align*}
Since $G_K(\eta_\alpha)\geq\eta_{\min}^{-2}$, it follows that
$$\bar{J}((\eta_t)_{t=0}^\alpha)\leq \eta_{\min}^2\, \delta^2G_K(\eta_\alpha).$$
Thus, Assumption~\ref{ass: coercivity} holds with $\kappa_\Eta=\frac{1}{\eta_{\min}^2}$.
Overall, Theorem~\ref{thm:exp-escort-ksd-certificate} thus yields
$$ D_\alpha (Q  || P) \le  \frac{\eta_{\min}^2}{2\alpha \eta_{\max}^2} \KSD_{K} (P , Q_{\alpha | P})^2.$$
\end{proof}

\begingroup
\def\thetheorem{\ref{cor:exp-gaussian}}
\begin{corollary}[Gaussian family]
Let $P = P_\eta$ belong to the Gaussian
natural exponential family with sufficient statistic $T(x)=x$, base measure $\nu = \mathcal N(a,\Sigma)$ for  $a\in\R^d$ and $\Sigma\succ 0$. Set
$K(x,y)=\psi(\norm{x-y})I_d$ for some ${\psi:\R\to\R_+}$. Then, for $ \lambda_{\max}(\Sigma) $ the largest eigenvalue of $\Sigma$ and ${\kappa_\Eta = \E_{{Z}\sim\mathcal{N}(0,2\Sigma)}[\psi(\norm{{Z}})] }$, a bound $\KSD_{K} (P , Q_{\alpha | P})^2 \leq \epsilon$ uniformly over all $\mathbf{D}\sim\mathbf{D'}$ implies $(\alpha,\frac{ \lambda_{\max}(\Sigma)}{2\alpha \, \kappa_\Eta}\epsilon )$-RDP.
\end{corollary}
\addtocounter{theorem}{-1}
\endgroup
\begin{proof}
Let $p_{\eta}(x) = \exp(\eta^\top x - A(\eta)) =  \exp(\eta^\top x - a^\top \eta - \frac{1}{2}\eta^\top \Sigma \eta) $ and consider  
the regular natural exponential family $P_\eta =\mathcal{N}(a + \Sigma \eta, \Sigma)$ with sufficient statistic $T(x) = x$ and base measure~$\nu = \mathcal{N}(a,\Sigma)$. In particular, choosing $\eta_\mathbf{D} = \Sigma^{-1} m_\mathbf{D}$ and $a = 0$ recovers the Gaussian mechanism, see Subsection~\ref{ssec: gaussian mechanism}.

We show that Assumptions~\ref{ass: restrictedpoincare} and \ref{ass: coercivity} are satisfied with $C_\Eta = \lambda_{\max}(\Sigma) $ and $\kappa_\Eta = c_k= \E_{X,Y\sim\mathcal{N}(0,2\Sigma)}[k(X,Y)] $ such that an application of Theorem~\ref{thm:exp-escort-ksd-certificate} yields
$$D_\alpha (Q  || P) \le  \frac{\lambda_{\max}(\Sigma)}{2 \alpha c_k} \KSD_{K} (P , Q_{\alpha | P})^2$$
which implies the claim.
Clearly, $J_T(x) \equiv I_d$ is constant. Thus, 
$$\Var_{P_{\eta_t}}\left(\delta^\top T(X)\right) = \delta^\top \Sigma \delta \leq \lambda_{\max}(\Sigma) \norm{\delta}^2 =  \lambda_{\max}(\Sigma)\,\E_{P_{\eta_t}}\left\|J_T(X)^\top \delta\right\|^2$$
and thus Assumption~\ref{ass: restrictedpoincare} is fulfilled with $C_{\Eta} = \lambda_{\max}(\Sigma)$ the largest eigenvalue of $\Sigma$. 

For Assumption \ref{ass: coercivity} , note that $\E_{P_{\eta_t}}\left\|J_T(X)^\top \delta\right\|^2 = \norm{\delta}^2$ again implies $\bar{J}((\eta_t)_{t=0}^\alpha) = \norm{\delta}^2,$ see the proof of Corollary~\ref{cor:exp-laplace}.
Furthermore, we have $G_K (\eta_\alpha) = \E_{X,Y\stackrel{\mathrm{iid}}{\sim}P_{\eta_\alpha}}\left[k(X,Y)\right]I_d$. Thus, taking a scalar kernel $K(x,y) = k(x,y) I_d$ for which $G_K (\eta_\alpha) =  c_k I_d $ is constant with some $c_k >0$, Assumption~\ref{ass: coercivity} is satisfied with $\eta_{\Eta} = c_k$. This kernel assumption includes  the constant kernel~$k(x,y) \equiv 1$ with $c_k = 1$ or a translation invariant kernel~$k(x,y) = \psi(\norm{x-y})$ as $X-Y\sim \mathcal{N}(0,2\Sigma)$ independently of $\eta_\alpha$ such that $c_k = \E_{Z \sim \mathcal{N}(0,2\Sigma)}[\psi(\norm{Z})]$. 

As all assumptions are satisfied, Theorem~\ref{thm:exp-escort-ksd-certificate} yields 
$$D_\alpha (Q  || P) \le  \frac{\lambda_{\max}(\Sigma)}{2 \alpha c_k} \KSD_{K} (P , Q_{\alpha | P})^2.$$
{{Corollary \ref{cor:exp-gaussian} illustrates} that Assumption~\ref{ass: restrictedpoincare} introduces some looseness compared with the mechanism-specific results of Subsection~\ref{ssec: gaussian mechanism}, but allows the certificate to apply to a substantially broader class of mechanisms.}
\end{proof}

\begingroup
\def\thetheorem{\ref{cor:exp-quartic}}
\begin{corollary}[Quartic family]
Let $P = P_\eta$ belong to the natural exponential family with sufficient statistic $T(x) = x$ and base density $h_0(x)\propto \exp\left(-\frac12x^\top\Sigma x
-\frac{\gamma}{4}\sum_{i=1}^d x_i^4\right),$  for some $ \Sigma\succ 0$ and $\gamma>0$. Consider the constant kernel $K(x,y) = I_d$ and let $\lambda_{\min}(\Sigma)$ be the smallest eigenvalue of $\Sigma$. Then $\KSD_{K} (P , Q_{\alpha | P})^2 \leq \epsilon$ uniformly over all $\mathbf{D}\sim\mathbf{D'}$ implies $(\alpha,\frac{ \epsilon}{2\alpha\lambda_{\min}(\Sigma) } )$-RDP.
\end{corollary}
\addtocounter{theorem}{-1}
\endgroup
\begin{proof}
Let $p_\eta(x)= \exp\left(\eta^\top x-A(\eta)\right)$ with $$A(\eta) = \log\int_{\R^d}
    \exp(\eta^\top x-\frac12x^\top \Sigma x-\frac{\gamma}{4}\sum_{i=1}^d x_i^4)dx$$
for some symmetric $ \Sigma\succ 0$ and $\gamma>0$. Set $T(x) = x$ and consider the base measure $ \nu(dx)= h_0(x) \dd x$ with $$h_0(x) = \exp\left(-\frac12 x^\top \Sigma x-\frac{\gamma}{4}\sum_{i=1}^d x_i^4\right).$$ Then for any $\eta \in\Eta$ the corresponding regular natural exponential family is given by $$P_\eta(\dd x) =  \exp\left(\eta^\top x - A(\eta) -\frac12x^\top \Sigma x-\frac{\gamma}{4}\sum_{i=1}^d x_i^4\right)dx =  p_\eta(x) h_0(x) \dd x.$$

We show that the Assumptions~\ref{ass: restrictedpoincare} and \ref{ass: coercivity} are satisfied with $C_\Eta = {(\lambda_{\min}(\Sigma))^{-1}}$, and $\kappa_\Eta = 1$. Consequently, an application of Theorem~\ref{thm:exp-escort-ksd-certificate} yields
    $$ D_\alpha(Q\|P)
    \leq
    \frac{1}{2\alpha\lambda_{\min}(\Sigma)}
    \KSD_K(P,Q_{\alpha|P})^2$$
    and thus the claim.

We first show that Assumption~\ref{ass: restrictedpoincare} is satisfied using the Brascamp–Lieb inequality, see \cite[Section~4.9.2]{bakry2014analysis}. We consider the negative log density $W = -\log(p_\eta h_0)$ of $P_\eta$. First, note that
$$W(x) = -\log(p_\eta(x) h_0(x)) = -\eta^\top x +\frac12x^\top \Sigma x+\frac{\gamma}{4}\sum_{i=1}^d x_i^4 $$
is smooth and differentiating twice gives
$$ \nabla^2 W(x) = - \nabla^2 \log(p_\eta(x) h_0(x)) =\Sigma+3\gamma\diag(x_1^2,\ldots,x_d^2)\succeq\lambda_{\min}(\Sigma) I_d,$$
where $\lambda_{\min}(\Sigma) >0 $ denotes the smallest eigenvalue of $\Sigma$ that is strictly positive as $\Sigma \succ 0$. Thus, the negative log density is convex. By the Brascamp–Lieb inequality, see \cite[Section~4.9.2]{bakry2014analysis}, it follows that 
$$\Var_{P_\eta}(f) \leq \frac{1}{\lambda_{\min}(\Sigma)}
    \E_{P_\eta}\|\nabla f(X)\|^2,$$
for every sufficiently smooth function~$f:\R^d\to \R$. 
In particular, taking $f(x) = \delta^\top T(X)$, Assumption~\ref{ass: restrictedpoincare} is satisfied with $ C_\Eta=\frac{1}{\lambda_{\min}(\Sigma)}.$

For Assumption \ref{ass: coercivity}, consider the constant kernel $K(x,y)=I_d$. 
Then as $J_T(x) = I_d$,
$$ G_K(\eta_\alpha) =\E_{X,Y\stackrel{\mathrm{iid}}{\sim}P_{\eta_\alpha}}\left[ J_T(X)K(X,Y)J_T(Y)^\top\right]=I_d.$$
Moreover,
$\E_{P_{\eta_t}}\left\|J_T(X)^\top\delta\right\|^2=\|\delta\|^2$ for every $t\in[0,\alpha]$. Consequently, $\bar J((\eta_t)_{t=0}^\alpha)    =\|\delta\|^2$, see the proof of Corollary~\ref{cor:exp-laplace}.
Thus, Assumption~\ref{ass: coercivity} holds with $\kappa_\Eta=1.$

Overall, an application of Theorem~\ref{thm:exp-escort-ksd-certificate} yields
   $$ D_\alpha(Q\|P)
    \leq
    \frac{1}{2\alpha\lambda_{\min}(\Sigma)}
    \KSD_K(P,Q_{\alpha|P})^2.$$
\end{proof}

\subsection{Additional information for Section~\ref{ssec: renyi pufferfish privacy}}\label{app: pufferfish}
We start by 
{briefly} recalling the Pufferfish framework, which is given by  a triple $(\SPF,\QPF,\TPF)$. Here, $\SPF$ is a collection of secrets, where a secret can be interpreted as an event concerning $\mathbf{D}$ and is formally defined a measurable subset of $\mathfrak{D}$. Furthermore, $\QPF\subseteq\SPF\times\SPF$ is the collection of protected secret pairs. These pairs are ordered as  R\'{e}nyi divergence is {asymmetric, i.e. $D_{\alpha}(P\| Q) \neq D_{\alpha}(Q\|P)$}. In particular, if  both orientations are intended, both $(s_i,s_j)$ and $(s_j,s_i)$ should be included in $\QPF$ or verified separately. Finally, $\TPF$ {indexes} a class of data-generating distributions~$\mathbf{P}_\theta$. We interpret $\theta\in \TPF$ as admissible adversarial prior knowledge. { Then for any $\theta\in\TPF$,  $\mathbf{P}_{\theta}$ defines the dataset distribution on $\mathfrak{D}$ and we assume that $\mathbf{D}\sim\mathbf{P}_{\theta}$. In particular, the dataset~$\mathbf{D}$ is now random under $\theta\in\TPF$.}

Let $M$ be a randomized mechanism. For a secret $s\in\SPF$ with positive probability under $\theta$, define the secret-conditioned output law
\begin{align*}
P_{\theta,s}
:=
\Law\bigl(M(\mathbf{D})\mid s,\theta\bigr),
\qquad \mathbf{D}\sim\textbf{P}_\theta.
\end{align*}
{Thus, to sample from $P_{\theta,s}$, we first sample a dataset~$\mathbf{D}$ from $\textbf{P}_\theta$ conditional on the secret $s$ and then apply the randomized mechanism~$M$. The following remark collects important properties of R\'{e}nyi Pufferfish privacy.}
\begin{remark}\label{prop:rpp-properties}[Established RPP properties, \cite[Proposition 2.1 and 2.2]{pierquin2024renyi}]
If $M$ is $(\alpha,\varepsilon)$-RPP, then:
\begin{enumerate}
\item every deterministic or randomized post-processing of $M$ is also $(\alpha,\varepsilon)$-RPP;
\item for every $\delta\in(0,1)$, $M$ is
$ 
\left(
\varepsilon+\frac{\log(1/\delta)}{\alpha-1},\delta
\right)$
-Pufferfish private.
\end{enumerate}
Unlike ordinary record-level RDP, RPP does not have a universal mechanism-agnostic adaptive composition theorem. In particular, conditioning on a released transcript updates the adversary's prior, which may move it outside the declared class $\TPF$. \cite{pierquin2024renyi} therefore develops privacy amplification by iteration as an alternative for contractive noisy algorithms. 
\end{remark}

To link this framework with escort distributions, fix a prior $\theta\in\TPF$ and define the {\it Pufferfish attainable escort family}  by 
\begin{align}\label{eq:attainable-family}
\EPF_\alpha(\mathbf{P}_{\theta})
:=  \Big\{
(P_{\theta, s_j})_{\alpha| \mathbf{P}_{\theta, s_i}}: (s_i,s_j)\in\QPF;
\mathbf{P}_\theta(s_j)>0,  
 \mathbf{P}_\theta(s_i)>0  \Big\}.
\end{align}
The Pufferfish attainable escort family captures the comparison specification imposed by the Pufferfish framework as it contains only escorts generated by the secret comparisons and 
priors {{$\mathbf{P}_\theta$}}. It can be far smaller than the set of all probability measures. 

\subsection{Additional information for Subsection~\ref{ssec: task aware kernel design}}\label{app: task aware kernel design}
\begingroup
\def\thetheorem{\ref{thm:task-rbf}}
\addtocounter{theorem}{-1}
\begin{theorem}
    Let $\mathbb{S}$ be a public task subspace with orthonormal basis~$U_\mathbb{S}\in\R^{d\times r}$ and kernel $K_{B,A_\mathbb{S}^d}$. Then
\begin{equation*}
D_\alpha((U_\mathbb{S}^\top)_{\#}Q\|(U_\mathbb{S}^\top)_{\#}P)
=
\frac{1}{2\alpha c_{B,\Sigma}}
\KSD_{K_{B,A_\mathbb{S}^d}}(P,Q_{\alpha\mid P})^2.
\end{equation*}
If furthermore $ \Sigma^{-1}\delta\in \mathbb{S}$, then 
 \begin{equation*}
 D_\alpha(Q\|P)=
 \frac{1}{2\alpha c_{B,\Sigma}}
 \KSD_{K_{B,A_\mathbb{S}^d}}(P,Q_{\alpha\mid P})^2.
 \end{equation*}
\end{theorem}
\endgroup
\begin{proof}
As $P$ and $Q$ are Gaussians, we obtain $$(U_\mathbb{S}^{\top})_{\#}P \sim \mathcal{N}(U_\mathbb{S}^{\top}m_{\theta,s_i},U_\mathbb{S}^{\top}\Sigma U_\mathbb{S}) \quad  \mbox{ and } \quad (U_\mathbb{S}^{\top})_{\#}Q_{\alpha\vert P} \sim \mathcal{N}(U_\mathbb{S}^{\top}(m_{\theta,s_i}+\alpha\delta),U_\mathbb{S}^{\top}\Sigma U_\mathbb{S}).$$  
Note that the adapted covariance~$\Sigma_{\mathbb{S}} := U_\mathbb{S}^{\top}\Sigma U_\mathbb{S}\in\R^{r\times r}$ is positive definite, i.e. $U_\mathbb{S}^{\top}\Sigma U_\mathbb{S}\succ 0$. We then choose~$K = K_{B,A}$ with $$A = A_{\mathbb{S}}^d = (\Sigma U_{\mathbb{S}})\Sigma_{\mathbb{S}}^{-1}(\Sigma U_{\mathbb{S}})^\top\in \mathbb R^{d\times d}$$ to 
{obtain} a positive-semidefinite matrix of the correct dimension. As $A$ fulfils $\Sigma^{-1}A\Sigma^{-1} = U_\mathbb{S}\Sigma_\mathbb{S}^{-1} U_\mathbb{S}^\top$, an application of Propositions~\ref{prop:gaussian-renyi} and \ref{prop:gaussianExactEscordKSD} analogous to Theorem~\ref{thm:covariance-matched} yields the first claim.

Now assume that $\Sigma^{-1}\delta\in\mathbb{S}$. Then as $\text{range}(U_{\mathbb{S}} = \mathbb{S}$, there exist a $v\in\R^r$ such that $\Sigma^{-1}\delta = U_\mathbb{S} v$. Then by Proposition~\ref{prop:gaussianExactEscordKSD},
\begin{align*}
    \KSD_{K_{B,A_\mathbb{S}^d}}(P,Q_{\alpha\mid P})^2 &= c_{B,\Sigma}\,\alpha^2\,(\Sigma^{-1}\delta)^\top A_\mathbb{S}^d  \Sigma^{-1}\delta\\
    &= c_{B,\Sigma}\,\alpha^2\, v^\top U_\mathbb{S}^\top (\Sigma U_{\mathbb{S}})\Sigma_{\mathbb{S}}^{-1} U_{\mathbb{S}}^\top\Sigma U_\mathbb{S} v\\
    &= c_{B,\Sigma}\,\alpha^2\, v^\top \Sigma_{\mathbb{S}}\Sigma_{\mathbb{S}}^{-1} \Sigma_{\mathbb{S}} v\\
    &= c_{B,\Sigma}\,\alpha^2\, v^\top \Sigma_{\mathbb{S}} v\\
    &= c_{B,\Sigma}\,\alpha^2\, \delta^\top \Sigma^{-1} \delta
\end{align*}
and the claim follows from Proposition~\ref{prop:gaussian-renyi}.
\end{proof}

\subsection{Additional information for Subsection~\ref{ssec: task aware exponential mechanism}}\label{app: task aware}
\begingroup
\def\thetheorem{\ref{ass: restrictedpoincare}a}
\begin{assumption} \label{ass: restrictedpoincarePufferfish}
Assume that there exist a constant $0<C_\Eta<\infty$ such that, for every $\eta\in\Eta$, every
$\delta\in\Delta_\eta^{\QPF}$ and every $t\in [0,\alpha]$,
\begin{align*}
\Var_{P_{\eta_t}}\left(\delta^\top T(X)\right)
&\leq C_\Eta \,\E_{P_{\eta_t}}\left\|J_T(X)^\top \delta\right\|^2.
\end{align*} 
\end{assumption}
\addtocounter{theorem}{-1}
\endgroup

\begingroup
\def\thetheorem{\ref{ass: coercivity}a}
\begin{assumption}\label{ass: coercivityPufferfish}
Assume  that there exists a constant $\kappa_\Eta >0$ such that, for every $\eta\in\Eta$ and every $\delta \in \Delta_{\eta}^\QPF$,
    \begin{align*}  
    \bar{J}((\eta_t)_{t=0}^\alpha) \le \frac{1}{\kappa_\Eta} \delta^\top G_K(\eta_\alpha) \delta, 
\end{align*}
where $\bar{J}((\eta_t)_{t=0}^\alpha)$ is defined in Assumption~\ref{ass: coercivity}.
\end{assumption}
\addtocounter{theorem}{-1}
\endgroup

\begingroup
\def\thetheorem{\ref{prop: task aware exponential mechanism certificate}}
\begin{proposition}
Let $\mathbb{S}$ be a public task subspace with projector $U_{\mathbb{S}}$ and define $K_{\mathbb{S}}(x,y):= k(x,y)U_{\mathbb{S}}U_{\mathbb{S}}^\top$ to be a task-aware kernel with scalar kernel~$k$.
Assume that Assumptions~\ref{ass: restrictedpoincarePufferfish} and \ref{ass: coercivityPufferfish} are satisfied with $K=K_{\mathbb{S}}$.
Then 
    \begin{align*}
     D_\alpha(Q\|P) \le  \frac{ C_\Eta}{2\alpha \, \kappa_\Eta} \KSD_{K_{\mathbb{S}}} (P , Q_{\alpha | P})^2.
\end{align*}
{A uniform bound on $\KSD_{K_{\mathbb{S}}}$ for all $\theta\in\TPF$ and $(s_i,s_j)\in\QPF$ implies the corresponding RPP guarantee.}
\end{proposition}
\addtocounter{theorem}{-1}
\endgroup
\begin{proof}
Follows directly from Theorem~\ref{thm:exp-escort-ksd-certificate}  by replacing $P_\eta$ by $P_{\theta,s}$.
\end{proof}

\begingroup
\def\thetheorem{\ref{cor: task aware privacy example}}
\addtocounter{theorem}{-1}
\begin{corollary}
Let $P = P_{\theta,s}$ belong to the natural exponential family with sufficient statistic $T(x) = \frac{1}{2} (a^\top x)^2 \in \R$, $x\in \R^d$ and $a\in\R^d$ such that $\norm{a}=1$, and base density $h_0(x)$ 
as in Corollary~\ref{cor:exp-quartic}. Let $\mathbb{S}= \text{span}(a)$  be the public task subspace. 
Assume that $\Delta_\eta^\QPF \subset \R_+$ for all ${\eta\in \Eta}$ and that ${\eta_{\max} := \sup_{\substack{\eta\in \Eta \\ \delta \in \Delta_\eta^\QPF}} (\eta + \alpha {\delta}) < (a^\top \Sigma^{-1} a)^{-1}}$. Then, $\KSD_{K_{\mathbb{S}}} (P , Q_{\alpha | P})^2 \leq \epsilon$ uniformly over all ${\theta\in\TPF}$ and $(s_i,s_j)\in\QPF$ implies $(\alpha,\frac{ {\epsilon} C_\Eta}{2\alpha \, \kappa_\Eta})$-RPP, 
where $C_\Eta = a^\top (\Sigma - a \eta_{\max} a^\top)^{-1} a $ and $\kappa_{\Eta} = \inf_{\substack{\eta\in\Eta\\ \delta\in\Delta_\eta^\QPF}} \E_{X\sim P_{\eta + \alpha\delta}}[(a^\top X)^2]$.
\end{corollary}
\endgroup
\begin{proof}
Fix a pair of secrets~$(s_0,s_1)\in\QPF$ and recall that $\eta = \eta_{\theta,s_0}$ while $\delta = \eta_{\theta,s_1} -\eta_{\theta,s_0} \in \Delta^\QPF_\eta$ such that $\eta + \delta = \eta_{\theta,s_1}$.
Set $P = P_{\theta,s_0} = P_\eta$ and $Q =  P_{\theta,s_1} = P_{\eta + \delta}$. Then the order-$\alpha$ escort is $Q_{\alpha| P} = P_{\eta + \alpha \delta}.$
We show that Assumptions~\ref{ass: restrictedpoincarePufferfish} and \ref{ass: coercivityPufferfish} are fulfilled with constants $C_\Eta =a^\top \big( \Sigma- \eta_{\max} a a ^\top \big)^{-1}a $ and $\kappa_{\Eta} = \frac{m_\alpha^2}{\bar{m_\alpha}}$, where $m_t = \E_{P_{\eta +t\delta}}[(a^\top X)^2 ]$, $t\in [0,\alpha]$ and 
$\bar{m}_\alpha =   \frac{2}{\alpha} \int_0^1 t  m_t  \, dt + \frac{2}{\alpha(\alpha-1)} \int_1^\alpha (\alpha -t) m_t \, dt.$
Then, an application of Proposition~\ref{prop: task aware exponential mechanism certificate} yields 
 \begin{align} \label{eq: task aware exponential example quatric}
     D_\alpha(Q\|P) \le  \frac{ C_\Eta}{2\alpha \, \kappa_\Eta} \KSD_{K_{\mathbb{S}}} (P , Q_{\alpha | P})^2
\end{align}
and the claim follows. 

We first study the more general case of $\Delta_\eta^\QPF \subset \R$.
As $\eta^{\max}_{\eta,\delta} := \max\{\eta,\eta + \alpha \delta\} < (a^\top \Sigma^{-1} a)^{-1}$ we have that
$$ \Sigma- (\eta + t\delta) a a ^\top \succ \Sigma- \eta^{\max}_{\eta,\delta} a a ^\top =: B_{\eta,\alpha} \succ 0 $$ uniformly for $t\in[0,\alpha]$. 
The negative log-density~$ W_t$ of $P_{\eta + t\delta}$, $t\in [0,\alpha]$, given by
$$ W_t(x) = \frac12 x^\top (\Sigma - (\eta + t\delta) a a^\top) x + \frac{\gamma}{4} \sum_{j=1}^d x_j^4 $$
is smooth and differentiating twice gives
$$\nabla^2 W_t(x) = \Sigma - (\eta + t\delta) a a ^\top + 3\gamma \diag(x_1^2,\ldots,x_d^2) \succeq B_{\eta,\alpha} \succ 0.$$
Thus, the negative log density is convex. By the Brascamp-Lieb inequality, see \cite[Section 4.9.2]{bakry2014analysis}, it follows that 
$$ \Var_{P_{\eta+t\delta}} (f) \le 
\E_{P_{\eta+t\delta}} [ \nabla f(X)^\top B_{\eta,\alpha}^{-1} \nabla f(X)]$$
 for all $f: \R_d \rightarrow \R $ that are in $L^2 (P_\nu).$ In particular, taking $f(x) = \delta T(x) = \frac{\delta}{2} (a^\top x)^2 $ and thus $\nabla f(x) = \delta a^\top x a$, which yields 
 $$ \Var_{P_{\eta+t\delta}} (f) \le  a^\top 
B_{\eta,\alpha}^{-1} a \E_{P_\eta} [\delta^2 (a^\top X a)^2] =  a^\top B_{\eta,\alpha}^{-1} a \E_{P_\eta} [  (J_T(X)\delta)^2].$$
In particular, {Assumption~\ref{ass: restrictedpoincarePufferfish}} is fulfilled with $C_\Eta =  a^\top \big( \Sigma- \eta_{\max} a a ^\top \big)^{-1}a$, where $\eta_{\max} = \sup_{\eta\in\Eta, \delta\in \Delta_\eta^\QPF} \eta^{\max}_{\eta,\delta}$.

Furthermore, note that $J_T(x) \delta = \delta\, a^\top xa \in\mathbb{S}$. Thus, the specified privacy directions all fall into the task-aware subspace. Furthermore, as $\mathbb{S} = \text{span}(a)$, we have $U_\mathbb{S} = a$. Taking the scalar kernel $k_a(x,y) = (a^\top x)(a^\top y)$, we obtain  $K_{\mathbb{S}} = k_a(x,y) U_{\mathbb{S}} U_{\mathbb{S}}^\top = (a^\top x)(a^\top y){a a^\top}
$, see Proposition~\ref{prop: task aware exponential mechanism certificate}. Then as $\norm{a} = 1$,
\begin{align*}
  G_{\mathbb{S}}(\eta + t \delta) &=  \mathbb E_{X,Y\sim P_{\eta +t\delta}}
    \left[
        J_T(X) K_{\mathbb{S}}(X,Y)J_T(Y)^{\mathsf T}
    \right] \\
    &=  E_{X,Y\sim P_{\eta +t\delta}}\left[
       a^\top X a (a^\top X)(a^\top Y) a^\top a (a^\top Y a)^{\mathsf T}
    \right]\\
    &=  E_{X,Y\sim P_{\eta +t\delta}}\left[
         (a^\top X)^2(a^\top Y)^2 
    \right]\\
    &=  E_{X\sim P_{\eta +t\delta}}\left[
         (a^\top X)^2 
    \right]^2 = m_t^2.
\end{align*}


On the other hand, we observe that $\E_{P_{\eta +t\delta}}\left\| J_T(X)^\top \delta\right\|^2 = \delta^2\E_{P_{\eta +t\delta}}[(a^\top X)^2 ]$ and thus we obtain 
\begin{align*}
    \bar{J}((\eta +t\delta)_{t=0}^\alpha) = \delta^2 \Big\{  \frac{2}{\alpha} \int_0^1 t  m_t  \, dt + \frac{2}{\alpha(\alpha-1)} \int_1^\alpha (\alpha -t) m_t \, dt \Big\} = \delta^2 \bar{m}_{\alpha}.
\end{align*} 
Then {Assumption \ref{ass: coercivityPufferfish}} 
is fulfilled with $\kappa_\Eta = \frac{m_\alpha^2}{\bar{m}_{\alpha}}$. 
Under the assumption that $\Delta_\eta^\QPF \subset \R_+$  {we can obtain a simpler bound as follows. As  $\delta \in \Delta_\eta^\QPF >0$,}  we have
$$\frac{\dd}{\dd t}m_t = \frac{\dd}{\dd t} \E_{\eta + t\delta}[(a^\top X)^2] = 2 \frac{\dd}{\dd t} \E_{\eta + t\delta}[T(X)] =  2 \delta \nabla^2 A(\eta + t\delta) = 2 \delta \Var_{\eta + t\delta}(T(X)) \geq 0. $$
Hence, $m_t$ is non-decreasing in $t$ and thus $m_t \leq m_\alpha$ for all $t\in[0,\alpha]$.
Consequently, $$\bar{m}_{\alpha} \leq m_\alpha \Big\{  \frac{2}{\alpha} \int_0^1 t   \, dt + \frac{2}{\alpha(\alpha-1)} \int_1^\alpha (\alpha -t)  \, dt \Big\} = m_\alpha.$$
Thus Assumption~\ref{ass: coercivityPufferfish} is {also} satisfied with $\kappa_\Eta = {m_\alpha}$.  
The claim now follows with Proposition~\ref{prop: task aware exponential mechanism certificate}.
\end{proof}

\section{Numerical illustrations}\label{app:track} 
{To illustrate the KSD certificates developed in Sections~\ref{sec:ksdprivate} and \ref{sec:pufferfish privacy}, we compare R'{e}nyi divergence with escort KSD quantities in several controlled settings.}

{The experiments were carried out on a Dell laptop featuring an Intel Core i7-11850H CPU @ 2.50GHz, 32 GB of RAM, and an NVIDIA GeForce MX450 GPU (2 GB dedicated VRAM).} { The experiments can be reproduced with the code provided under \url{https://anonymous.4open.science/r/KSD_Renyi_DP-495A/}. Where randomness is controlled by a fixed random seed, the corresponding seed is specified in the code.}

\subsection{Gaussian mechanism: kernel choice}
{
We first use the Gaussian mechanism, for which both Rényi divergence and the escort KSD are available in closed form, to illustrate how the kernel choice affects the tightness of the KSD certificate.}

\begin{figure}
        \centering
\includegraphics[width=0.7\linewidth]{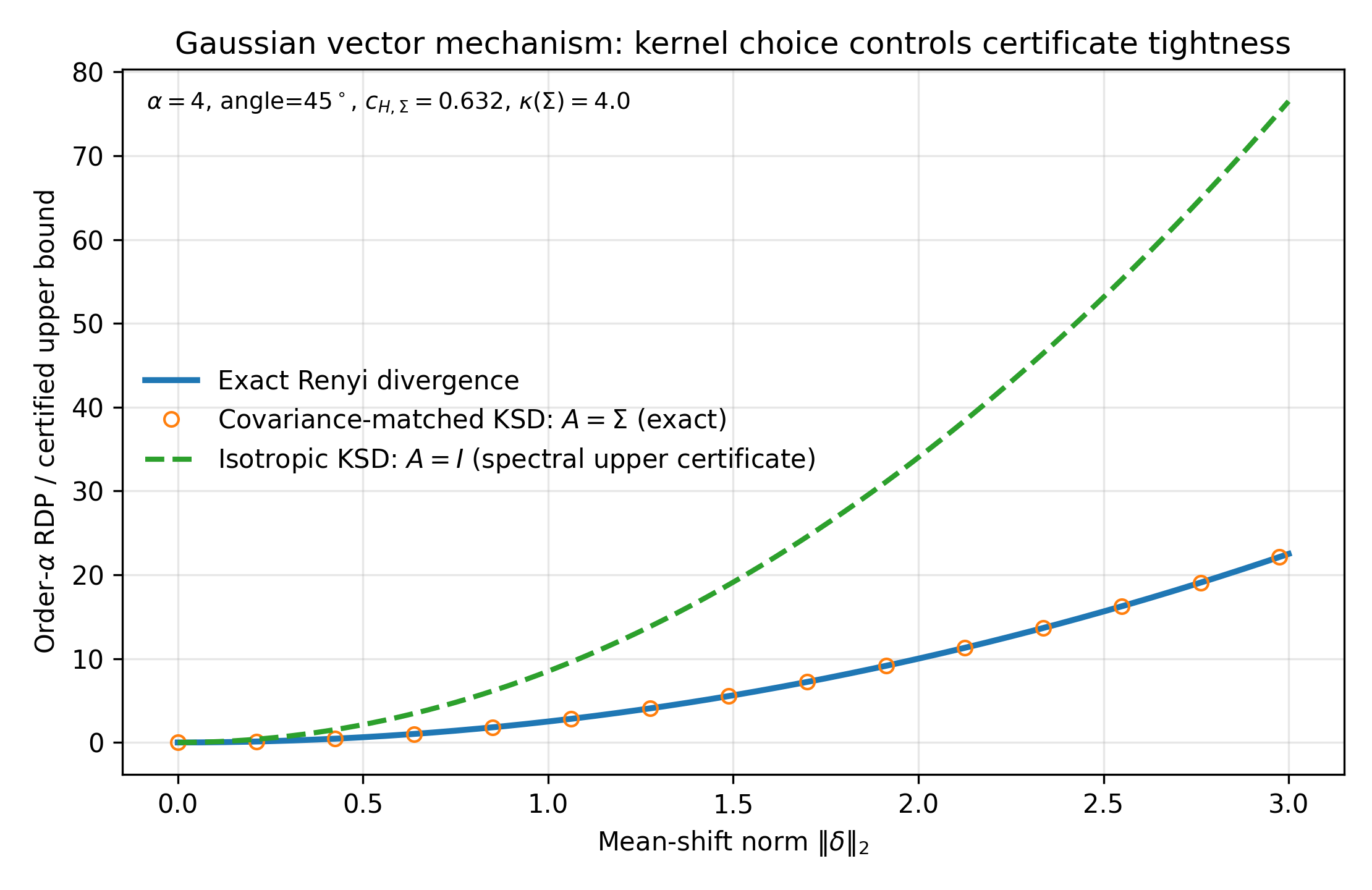}
        \caption{Numerical validation of Theorem~\ref{thm:covariance-matched}: varying the size of the displacement }
        \label{fig:Gaussiancert} 
    \end{figure}

\begin{figure}
        \centering
\includegraphics[width=0.7\linewidth]{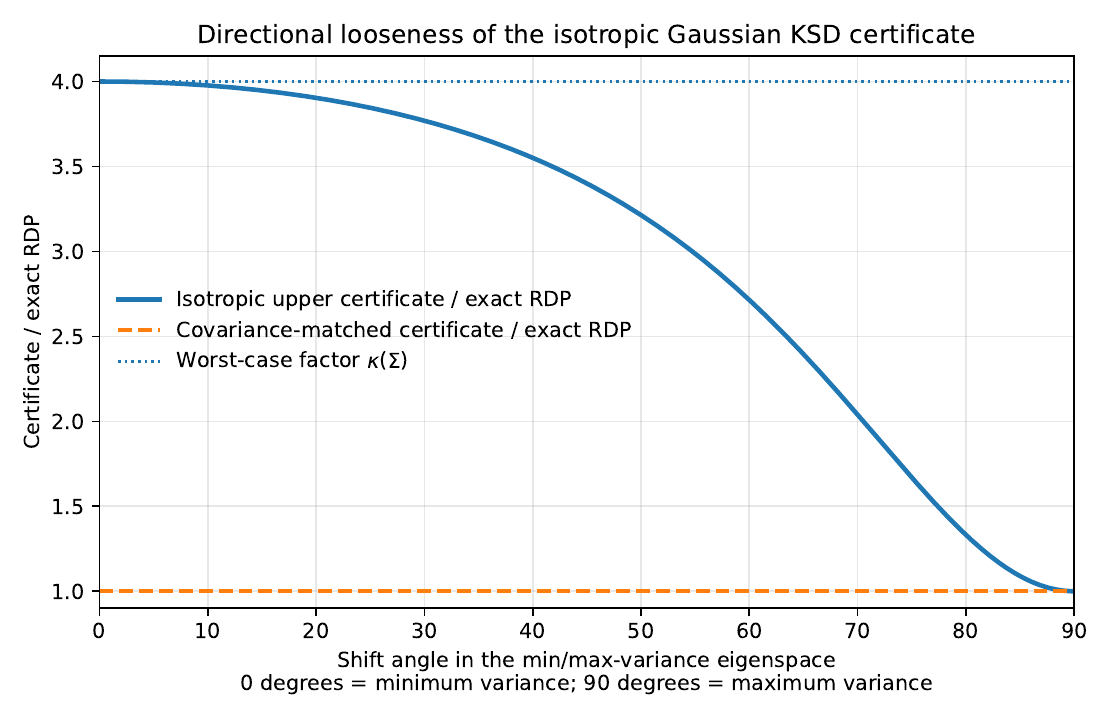}
        \caption{Numerical validation of Theorem~\ref{thm:covariance-matched}: varying the angle}
        \label{fig:Gaussiancert2} 
    \end{figure}

With the notation {in Subsection ~\ref{ssec: gaussian mechanism}}
    we use { 
    \begin{eqnarray*}     
\Sigma =
\begin{pmatrix}
  0.5 & 0 \\
  0 & 2 
\end{pmatrix}
    \end{eqnarray*} 
    with condition number $\kappa (\Sigma) {= \frac{\lambda_{\max}(\Sigma)}{\lambda_{\min}(\Sigma)} = \frac{2}{0.5}} =4$}, and we use $\delta = r (1/\sqrt{2}, 1/\sqrt{2})^\top$, a replacement in direction of 45 degrees,   with $||\delta ||_2 \in [0, 3]$ for the shift between different output means, equally spaced to get 15 values.
     The parameter for the R\'{e}nyi divergence is $\alpha=4$. {For the isotropic operator-valued kernel  in \eqref{eq:gaussiank} we take $B =  4 I_2$, giving $c_{B, \Sigma} = (5/2)^{-1/2}=0.632$. 
     Figure \ref{fig:Gaussiancert} confirms that, with {the covariance-matched kernel~{$K_{B,\Sigma}$}}, the identity in Theorem~\ref{thm:covariance-matched} holds (up to numerical variation);  for large changes $||\delta||_2$, the isotropic kernel~${K_{B,I_d}}$ gives only a loose upper bound. Figure \ref{fig:Gaussiancert2} instead varies the direction of the shift, $\delta (\theta) = (\cos(\theta), \sin(\theta))^\top$, keeping the norm fixed at $|| \delta||_2 = 1$. It shows that shift in a low-noise direction is easier to detect than an equally large shift in a high-noise direction. With the given  $\Sigma$ the smallest-noise-variance direction is $(1,0)$,  corresponding to the smallest  eigenvalue $\lambda_{\min}(\Sigma) = 1/2$, giving 0 degrees; the highest-noise-variance direction is $(0,1)$, corresponding to the largest eigenvalue $\lambda_{\max}(\Sigma) =2$, giving 90 degrees. }
    
    \medskip 

\subsection{Exponential family certificates}\label{app: exponential family certificates}
{
We next consider quartic exponential families to illustrate the KSD certificate beyond settings in which the Rényi divergence is readily available in closed form.}

\begin{figure}
         \centering
         \includegraphics[width=0.7\linewidth]{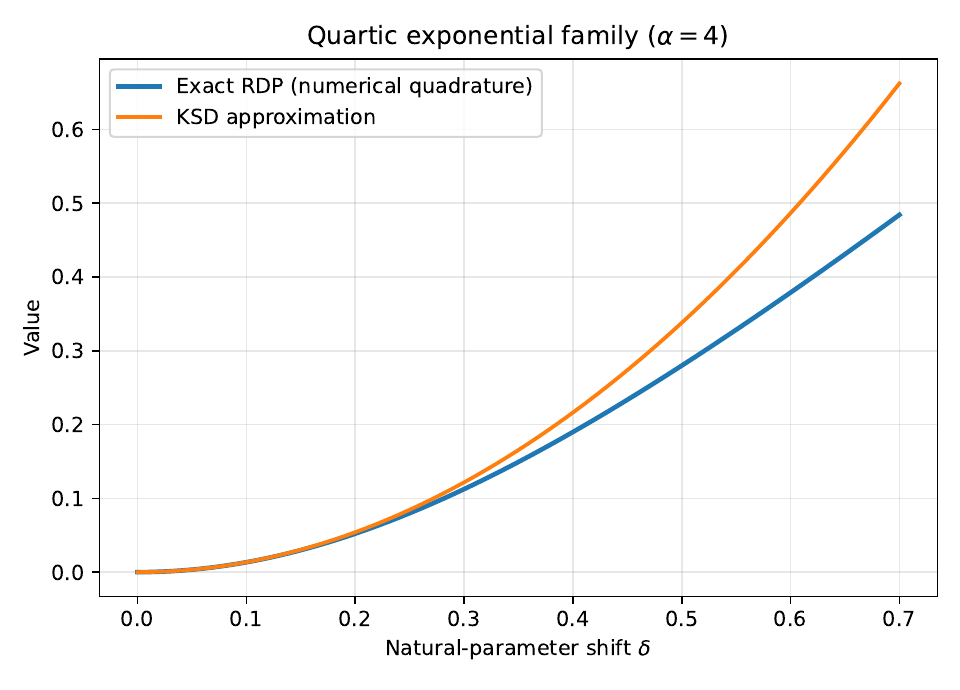}
         \caption{The KSD approximation to RDP for the quartic exponential family. The ``exact'' RDP is an approximation obtained by numerical quadrature.}
         \label{fig:quartic}
     \end{figure}

\paragraph{One-dimensional quartic example}
 A non-Gaussian quartic exponential family $p_\eta(x) \propto \exp(\eta x - x^4/4)$, with constant-kernel escort KSDs; {for its properties see for example \cite{holly2011fourth}}.  Here we set the baseline natural parameter to zero so that $\theta=0$ and we use a constant kernel. 
     {In this example, for  the Fisher information $I(\eta)= I(-\eta)$
     has negative derivative for $\eta>0$ and henc has its unique global maximum at   $\eta=0$. Thus,}  from \eqref{eq:exp-renyi-log-partition} we obtain the bound 
     \begin{align}
          D_\alpha(Q\|P)
    &\le \frac{\alpha}{2} \delta^\top I(0) \delta. \label{eq:crude}
     \end{align}
     As $I(0) = {2  \frac{ \Gamma \left( \frac34 \right)}{\Gamma \left( \frac14 \right)} \approx 0.6759}$ has an explicit formula, it can be well approximated numerically. Moreover, in this one-dimensional case, $I(0)$ equals the variance. Instead of using the bound in Corollary \ref{cor:exp-quartic}, in this instance we can directly use that $\KSD_K(P, Q_{\alpha|P})^2 = \alpha^2 \delta^\top \delta$ to obtain  the sharper bound 
     \begin{align*}
          D_\alpha(Q\|P)
    &\le \frac{I(0)}{2\alpha } \KSD_K(P, Q_{\alpha|P})^2.
    \end{align*} 
      Figure \ref{fig:quartic} shows that this sharper  KSD approximation is reasonable for small parameter shifts.  {{
      This example is another  illustration that in special cases, the bound in Corollary \ref{cor:exp-quartic} can be improved.}} 

\paragraph{Coupled quartic example}
In this example we assess the quality of the KSD certificate for the exponential mechanism derived in Subsection ~\ref{ssec: exponential mechanism}. We focus on a  high-dimensional continuous {quartic} exponential family. In the framework of Corollary~\ref{cor:exp-quartic} assume that $\Sigma\succ \mu I_d \succ 0$ for some $\mu >0$ and set $\gamma = 1$, and consider the family
$$
p_t(x) \propto \exp\Big\{(\eta+t\delta)^\top x
-\frac12 x^\top \Sigma x-\frac14\sum_{i=1}^d x_i^4\Big\}.
$$
Then $\Sigma\succ \mu I_d$ implies $\lambda_{\min}(\Sigma) \geq \mu$ and Corollary \ref{cor:exp-quartic} yields that Assumptions~\ref{ass: restrictedpoincare} and \ref{ass: coercivity} are fulfilled with constant $C_\Eta = \frac{1}{\mu}$ and $\kappa_\Eta = 1$. In particular, the proof of Corollary~\ref{cor:exp-quartic} yields
\begin{align}
    \label{eq:quartic-renyi-ksd-bound}
    D_\alpha(Q\|P)
    \leq
    \frac{1}{2\alpha\mu}
    \KSD_K(P,Q_{\alpha|P})^2 = \frac{\alpha}{2\mu}\norm{\delta}_2^2,
\end{align}
where the last equality follows from Proposition~\ref{prop:exp-exact-escort-ksd} as $G_K(\eta) = I_d$; see the proof of Corollary~\ref{cor:exp-quartic}. 
In particular, evaluating the privacy certificate does not require a normalising constant.

To evaluate the tightness of this certificate, we estimate the R\'{e}nyi divergence independently by Metropolis-adjusted Langevin
sampling at a grid of path parameters and two {\it thermodynamic-integration {(TI)} identities}  
\begin{align}    
D_\alpha(P_1|P_0)
&=\frac{\int_0^\alpha m(t)dt-\alpha\int_0^1m(t)dt}{\alpha-1},
\qquad m(t)=\mathbb E_t[\delta^\top X], \label{eq:ti1}
\\
D_\alpha(P_1|P_0)
&=\int_0^1t,v(t)dt
+\frac1{\alpha-1}\int_1^\alpha(\alpha-t)v(t)dt,
\qquad v(t)=\operatorname{Var}_t(\delta^\top X).
\label{eq:ti2}
\end{align}
{To see this, writing $ A(\theta_t):=\log Z_t, $  differentiation under the integral sign gives $ A'(t) = m(t). $
With \eqref{eq:exp-renyi-log-partition}
 and using 
$ A(b)-A(a)=\int_a^b m(t)\,\mathrm{d}t, $  we obtain 
\eqref{eq:ti1}. {As we integrate over the mean path, we think of this \eqref{eq:ti1} as a mean-path thermodynamic-integration identity.}
Differentiating once more gives $A''(t) = \operatorname{Var}_{P_t} \left( \delta^\top X \right). $ With $v(t) := \operatorname{Var}_{P_t} \left( \delta^\top X \right)$,  Taylor's formula applied to  \eqref{eq:exp-renyi-log-partition} yields 
\eqref{eq:ti2}.}

\begin{figure}
    \centering
\includegraphics[width=0.7\linewidth]{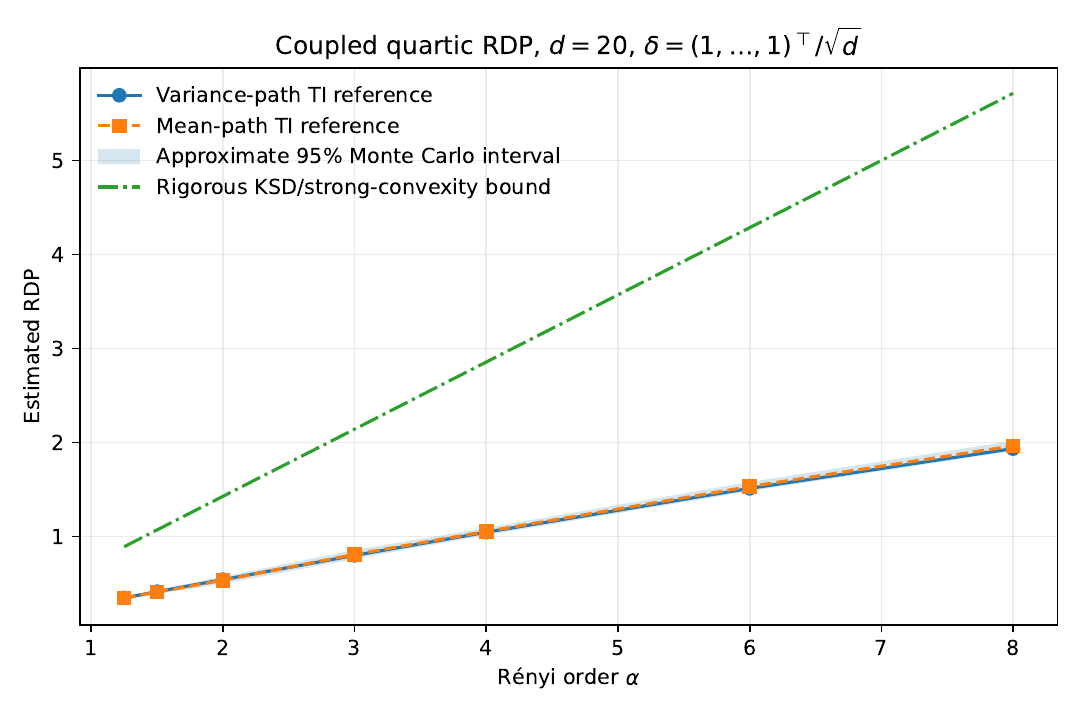}
    \caption{R\'enyi divergence for a fixed neighbouring natural-parameter displacement $\delta= \frac{1}{\sqrt{d}} (1, \ldots, 1)^\top$ in the coupled quartic exponential family, with $d=20$. The solid and dashed curves are numerical RDP estimates obtained from the mathematically equivalent variance-path and mean-path thermodynamic-integration identities \eqref{eq:ti2} and \eqref{eq:ti1}. Their agreement provides an MCMC and quadrature diagnostic. The dash-dotted curve is the rigorous strong-convexity/KSD upper bound $\alpha\|\delta\|_2^2/(2\mu)$. The shaded region is an approximate Monte Carlo uncertainty interval for the mean-path estimate.}
    \label{fig:renyi}
\end{figure}

{We again validate the KSD approximation. }
{In this experiment, we vary $\alpha \in \{1.5, 2, 4, 8\}$. The matrix $\Sigma=  (s+\mu) I_d - s B \in \R^{d \times d}$ is chosen based on the symmetric matrix $B$ with entries $0$ on the diagonal and for $i \ne j$, $B_{ij} = \rho^{|i-j|} $ with $\rho=0.55$, normalised to have spectral radius 1; then $ \Sigma - \mu I_d = s( I_d-B)\succ 0 $. We take  $\eta=0, s=0.4$,  $\mu = 0.7$, and $\delta = \frac{1}{\sqrt{d}} (1, 1, \ldots)^\top$.} 
In Figure \ref{fig:renyi},  $d=20$, and the horizontal axis is the Rényi order $\alpha$.  The plot 
compares three ways of quantifying its order-$\alpha$ Rényi divergence. 
The dashed line labelled {\it{Mean-path TI reference}}  is obtained by estimating $m(t)$ from MCMC samples from $P_t$, interpolating the estimates over a grid of $t$-values, and numerically evaluating \eqref{eq:ti1}.  The solid line with circular markers labelled {\it{Variance-path TI reference}} is obtained by estimating $v(t)$  from the same path samples and numerically evaluating \eqref{eq:ti2}. As \eqref{eq:ti1} and \eqref{eq:ti2} are mathematically equivalent, their numerical agreement  provides a diagnostic for the accuracy of the MCMC sampling, the path discretisation, the interpolation, and the numerical quadrature. 
The dash-dotted line labelled {\it{Rigorous KSD/strong-convexity bound}}  plots the right-hand side of \eqref{eq:quartic-renyi-ksd-bound}. 
It is a rigorous upper bound, and its distance from the two TI curves shows the conservativeness of the uniform strong-convexity argument. The shaded region is an approximate Monte Carlo uncertainty interval around the mean-path thermodynamic-integration estimate. It is constructed by perturbing the estimated path means \(m(t)\) according to their estimated Monte Carlo standard errors, recomputing \eqref{eq:ti1}, and taking empirical quantiles of the resulting values. This band is a numerical diagnostic rather than a rigorous confidence band. In particular, it does not fully account for dependence between different path points, interpolation error, quadrature error, MCMC bias, or imperfect mixing. 
{The exact KSD bound is clearly conservative, but of the same order of magnitude.}

\subsection{Task-aware kernel restriction}
\label{app:densequartic}
{Finally, we return to the quartic exponential-family setting of Subsection~\ref{app: exponential family certificates} to illustrate the task-aware KSD certificate of Corollary~\ref{cor: task aware privacy example}.}

In Example \ref{ex:densequartic} 
 we use $d=40, \alpha= \in \{ 1, 2, 4, 8\}, \gamma=0.1$, $\delta = 0.1094 $, and $\Sigma$ and $a$ are generated such that ${\delta}  a^\top \Sigma^{-1} a =0.05$. 
The values for $a$ and the entries for $\Sigma$ were generated at random. The particular instance for $a$ and $\Sigma$ which is used for Table \ref{tab:dense-quartic-alpha-study} can be found in the file {\tt{dense\_quartic\_instance.npz}} in the Outputs folder of the anonymous github site available at \url{https://anonymous.4open.science/r/KSD_Renyi_DP-495A/}. 

 As $\delta > 0$, the proof of Corollary~\ref{cor: task aware privacy example} gives $\kappa_\Eta = m_\alpha$ and Inequality~\ref{eq: task aware exponential example quatric} yields 
 \begin{align}\label{bound:ksd}
     D_\alpha (Q || P) =  D_\alpha (P_{\eta + \delta} || P_\eta) \le \frac{\alpha \delta^2}{2} C_\Eta m_\alpha. 
 \end{align}
 Furthermore, since $p_{\eta}(-x) = p_{\eta}(x)$ while $x\mapsto a^\top x$ is odd, we obtain $\E_{X\sim P_{\eta + \alpha \delta}}[a^\top X] = 0$. Thus, $m_{\alpha } = \Var_{X\sim P_{\eta + \alpha \delta}}(a^\top X)$ and applying the Brascamp-Lieb inequality as in the proof of Corollary~\ref{cor: task aware privacy example} but with $f(x) = a^\top x$ yields
 $$m_\alpha = \Var_{X\sim P_{\eta + \alpha \delta}}(a^\top X) \leq a^\top \big( \Sigma- \eta_{\max} a a ^\top \big)^{-1}a =  C_\Eta.$$  In particular, we obtain the task-unaware coarser fallback bound 
 \begin{align}\label{bound:fallback}
     D_\alpha (p_b || P_0) \le \frac{\alpha \delta^2}{2} C_\Eta^2 . 
 \end{align}
 While the KSD can be calculated in closed form, the escort moment $m_\alpha$ still has to be estimated numerically; we use MCMC for this task. {More details can be found in the file {\tt{dense\_quartic\_metadata.json}} on the anonymous github site.}

\section{Details of the real-data ACSIncome illustration}
\label{app:acs-real-data}

Here we provide the details for the real-data experiment in Section \ref{sec:acs-real-data}, which uses the ACSIncome data from the Folktables benchmark \citep{ding2021retiring} {available at {\url{https://github.com/socialfoundations/folktables}}}, {which is released under the MIT license}, to give a real-data
illustration of the privacy certificates developed above.
{The experiments were carried out on a Dell laptop featuring an Intel Core i7-11850H CPU @ 2.50GHz, 32 GB of RAM, and an NVIDIA GeForce MX450 GPU (2 GB dedicated VRAM).} {The experiments can be reproduced with the code provided under \url{https://anonymous.4open.science/r/KSD_Renyi_DP-495A/}. The seed is specified in the code.}

The dataset we use,  ACSIncome,   a standard benchmark analysed for example in \cite{bullwinkel2022evaluating}, is derived
from the US Census American Community Survey Public Use Microdata Sample; {we use ${\rm{survey\_year="2018", horizon="1-Year"}}$, and {$\rm{survey="person"}$}.}

Our task is to predict $y={\bf{1}}({\text{an individual's annual income exceeds \$50,000}})$, using demographic and employment variables such as age, education, occupation, marital status, hours worked, sex, and race. 
The ACS records are public and disclosure-protected, so the experiment is a
pseudo-private benchmark.  We nevertheless impose a strict public/private/test
separation.  The 2018 California records form public auxiliary data
$\mathbf{D}_{\mathrm{pub}}$; the 2018 Texas data are the pseudo-private training set.  One private Texas record is the pair $z_i=(x_i, y_i)$; adjacency means replacing one such Texas record.  

\paragraph{Public feature map, model, and task direction}
\label{sec:acs-public-construction-final} 
Let $x$ contain the ten ACSIncome covariates
\texttt{AGEP, COW, SCHL, MAR, OCCP, POBP, RELP, WKHP, SEX, RAC1P}.  Using California only, we construct a feature map $\phi(x)$; we 
standardize \texttt{AGEP} and \texttt{WKHP} using the means and standard deviations from the California data set alone; we one-hot encode the remaining
variables using California-observed categories, and add an intercept.  A
category which appears only in Texas maps to the all-zero block of the relevant
one-hot encoding.   To bound the influence of a single record, we pick a a public constant $B_\phi>0$ and threshold 
  $\phi_B(x)
  =
  \phi(x)\min\left\{1,\frac{B_\phi}{\|\phi(x)\|_2}\right\}.$ 
We abbreviate the feature map $\phi_B$ by $\phi$.
A public logistic regression model $\beta_0$ with a selective ridge penalty is fitted on the public data release  $\mathbf{D}_{\mathrm{pub}}$ with
binary logistic regression negative log-likelihood 
\begin{equation}
  \ell(\beta;(x,y))
=
  \log \left(1+\exp\{\phi(x)^\top\beta\}\right)
  -y\phi(x)^\top\beta 
  \label{eq:ch8-logistic-loss}
\end{equation}
 and ridge matrix
$R_{\mathrm{ridge}}=\operatorname{diag}(0,I_{p-1})$, so that the intercept is
not penalized. The model thus has as overall objective function
$$ \min_\beta \sum_{i=1}^n \ell (\beta; z_i) + \frac{\lambda}{2} \sum_{j=1}^{p-1} \beta_j^2.$$ 
Each Texas record contributes as private record the fixed gradient
\begin{equation*}
  g(z)
:=
  \nabla_\beta\ell(\beta_0;z) =
  \phi(x)\left\{\sigma(\phi(x)^\top\beta_0)-y\right\},
  \label{eq:ch8-fixed-gradient}
\end{equation*}
where $\sigma(t)=(1+e^{-t})^{-1}$. 
The model $\beta_0$, the feature map, and every tuning constant below are fixed
before Texas is accessed.
Instead of the Fisher information matrix we use a public diagonal matrix;  
with
$p_i=\sigma(\phi(x_i)^\top\beta_0)$ for California records, define
\begin{equation*}
  d_j
  =
  \frac{1}{|\mathbf{D}_{\mathrm{pub}}|}
  \sum_{i\in \mathbf{D}_{\mathrm{pub}}}
  p_i(1-p_i)\phi_j(x_i)^2
  +\lambda[R_{\mathrm{ridge}}]_{jj},
  \end{equation*}
  and set $
  P_{\mathrm{pub},jj}
  =
  \min\left\{
  \frac{1}{\max(d_j,d_{\min})},P_{\max}
  \right\}.
$ 
the diagonal matrix 
$P_{\mathrm{pub}}$ is public.   The use of $P_{\max}$ is to cap the diagonal entries to prevent rare on-hot coordinates with small entries from generating very large weights.

{{In our first experiment} we focus on the output-side task restriction of Section~\ref{sec:pufferfish privacy}. In particular, we do not impose an additional Pufferfish restriction on the protected comparisons.} For our public one-dimensional task, let
$$G_{\mathrm{pub}} =
  \{i\in {\mathbf{D}_{\mathrm{pub}}}:{\tt{RAC1P_i}}=2\}. $$
  {The category ${\tt{RAC1P_i}}$ represents the Recoded Detailed Race Code.
It maps an individual's racial identity to an integer value from 1 to 9 based on the U.S. Census Bureau's Public Use Microdata Sample (PUMS) categories; ${\tt{RAC1P_i}=2}$ indicates  
{\it Black or African American alone}. }
We  denote the mean predicted positive-class probability by 
\begin{equation}
  h_{\mathrm{pub}}(\beta)
  =
  \frac{1}{|G_{\mathrm{pub}}|}
  \sum_{i\in G_{\mathrm{pub}}}
  \sigma(\phi(x_i)^\top\beta).
  \label{eq:ch8-public-task}
\end{equation}
The subgroup $G_{\rm pub}$ is used only here to define a public task direction; the resulting certificates should not be interpreted as a race-membership privacy or fairness guarantee.
Set
$
  b_h=\nabla h_{\mathrm{pub}}(\beta_0), 
  a_h=P_{\mathrm{pub}}b_h, $ and $
  u_h=a_h/\|a_h\|_2. $  
{The task subspace is $\mathbb{S} = \operatorname{span}\{u_h\}$ such that $U_{\mathbb{S}} = u_h \in \R^{d\times 1}$. Since the model and \(u_h\) are constructed exclusively from the California data, the task subspace $\mathbb{S}$ is public and fixed independently of the Texas data that we view as private. This is in accordance with the task-aware construction of Subsection~\ref{ssec: task-aware renyi pufferfish privacy}.}
\paragraph{Numerical experiment}
To  illustrate the task-aware mechanism, we use a  Gaussian mechanism at $\alpha_0=10, \epsilon_{\alpha_0} =1.$ This fixed mechanism is then evaluated at $\alpha \in \{ 2, 4, 8, 16\}.$ 
The experiment uses  $20{,}000$ public California and $n=20{,}000$ pseudo-private Texas
records used for private training. {To obtain the 20,000 records for each state, we use stratified subsampling, without replacement, separately within California and Texas.  using the binary income label $y$ as the stratum. We use seed 2026 for California and 2027 for Texas. The privacy guarantee in this experiment treats the already selected 20,000 Texas records as the private dataset. The results thus do not apply to the entire original Texas data, but only to the sample already obtained.}

The dimension is $p=82$; the  public parameters are $B_\phi=8$, 
$\lambda = 10^{-4}$, $d_{\rm min} = 10^{-3}$, and 
 $P_{\rm max} = 25$. We decompose each fixed record
contribution as $g(z)=u_h s_h(z)+r_\perp(z)$, with
$u_h^\top r_\perp(z)=0$, and use California-only thresholds 
$C_\parallel$ and $C_\perp$ for the two components, obtained as 99\% quantiles of 
$| u_h^\top g(z)| $ and $|| (I - u_h u_h^\top) g(z)||_2,$ respectively. For the clipping operator we use the abbreviation $ \operatorname{clip}(t, -c, c) = \max\{ -c; \min (t, c)\}$. The resulting vector
release is
\begin{align*}
Y_\mathbf{D}=& \frac1n\sum_i\Big\{u_h\,\operatorname{clip}(s_h(z_i);C_\parallel)+\operatorname{clip}_2(r_\perp(z_i);C_\perp) \Big\}+\sigma Z,
\end{align*}
where $ Z\sim N(0,I_p).$  
The replace-one sensitivity {of the full mechanism} is bounded by
$\Delta_{\rm full}=2\sqrt{C_\parallel^2+C_\perp^2}/20000$. 
{We now consider its projection on the task subspace $\mathbb{S} = \operatorname{span}(u_h)$. Since $U_{\mathbb{S}} = u_h$, the} the  task-projected privacy {output of Subsection~\ref{ssec: task-aware renyi pufferfish privacy}}  is
$U_{\mathbb{S}}^\top Y_\mathbf{D} =  u_h^\top Y_\mathbf{D}$, whose sensitivity is
$\Delta_h=2C_\parallel/20000$. For the same Gaussian noise,
\begin{equation}
\varepsilon_{\alpha,h}
=\frac{\alpha\Delta_h^2}{2\sigma^2}
=\frac{C_\parallel^2}{C_\parallel^2+C_\perp^2}
\varepsilon_{\alpha,\rm full}
\leq \varepsilon_{\alpha,\rm full}.
   \label{eq:clipping envelope}
\end{equation}
{The quantity $ \varepsilon_{\alpha,\rm full}$ is also called the {\it analytic full clipping envelope}, and $\varepsilon_{\alpha,h}$ is the {\it analytic projected clipping envelope}.}
{The smaller privacy budget results from evaluating privacy for the projected release $u_h^\top Y_\mathbf{D}$, rather than from a stronger guarantee for the full vector-valued release $Y_\mathbf{D}$.}
With  $ \sigma= \Delta_{\rm full} \sqrt{\frac{10}{2}}$, we set $\Sigma=\sigma^2 I_p$, and use the full kernel
$K_{\rm full}=\sigma^2 I_p$ as well as the projector kernel
$$K_h(x,y)= {K_{B,A_{\mathbb{S}}^d}(x,y) = (\Sigma U_\mathbb{S})(U_\mathbb{S}^{\top}\Sigma U_\mathbb{S})^{-1}(\Sigma U_\mathbb{S})^\top = }\sigma^2u_hu_h^\top$$ {with underlying kernel $k_B(x,y) \equiv 1$ as  defined in Subsection ~\ref{ssec: task aware kernel design}}. For {a fixed neighbouring pair $\mathbf{D}\sim \mathbf{D'}$, let $\delta = \E[Y_\mathbf{D'}] - \E[Y_\mathbf{D}]$ denote the corresponding displacement of the means. As $k_B(x,y) \equiv 1$, we have $c_{B,\Sigma} = 1$ and Theorem~\ref{thm:covariance-matched} gives} 
$D_\alpha^{\rm full}
=\frac1{2\alpha}\operatorname{KSD}_{K_{\rm full}}^2 $. Similarly, Theorem~\ref{thm:task-rbf} gives, for the task-projected release, $
D_\alpha^h
=\frac1{2\alpha}\operatorname{KSD}_{K_h}^2.
$
In particular, 
$$ D_\alpha^{\rm full} = D_\alpha^h + \frac{\alpha}{2 \sigma^2} || (I - u_h u_h^\top) \delta ||_2^2.$$
{Thus, for each neighbouring pair, the difference between the full and task-projected privacy losses is exactly the contribution of the mean shift orthogonal to the public task subspace.} 
This shows that the task-aware KSD exactly certifies the task-aware RDP of the projected mechanism.
Figure \ref{fig:taskawareresults} in the main text, {which plots $ D_\alpha^{\rm full} $ and $D_\alpha^h$}, shows that the task-aware RDP certificate uses a smaller privacy budget than the standard release; the dashed lines indicate empirical 95\% quantiles. {The empirical percentiles are not privacy guarantees. Furthermore, we remark that the projected release requires a smaller R\'{e}nyi privacy budget because the guarantee concerns only the publicly specified task direction; it does not imply the same privacy guarantee for the full vector-valued release.}

\medskip 
{{\bf A Pufferfish experiment.}
The previous experiment illustrated the projected mechanism in which a subspace is private. In this example we pick a discrete Pufferfish example. We protect the recorded sex-race configuration of a target participant $i$ from the Texas dataset; the relevant indicators are 
$A_i = {\bf 1}({{\tt{SEX_i}=2}})$ ({\it female}) and $B_i = {\bf 1}({{\tt{RAC1P_i}=2}})$. Suppose that the target's admissible race codes are ${\tt{RAC1P}} \in \{1, 2\}$, where $1$ codes for {\it White alone}. 
Thus there are four possible sensitive-attribute configurations for $(A_i, B_i)$, namely $\mathcal{S} = \{ s_i^{00}, s_i^{01}, s_i^{10}, s_i^{11}\}$ with $s_i^{ab} = \{ A_i = a, B_i =b\}$, where $(a,b) \in {\mathcal{C}} = \{ (0,0), (0,1), (1,0), (1,1)\} $. The prior class assigns arbitrary positive probabilities to the four admissible sex-race completions. We write a record as $z_i=(w_i, a_i, b_i)$, where $w_i$ contains all remaining inputs for $i$ used by the mechanism; this includes the income label. The {\it context} is the set $w= (i, D_{-i}, w_i),$ where $D_{-i}$ is the set of everybody  else's input. Pufferfish privacy then addresses the question how distinguishable different sex-race configurations of this participant are based on the released statistic, when all other attributes and records are already known. 
In our experiment we take 
${\mathcal{Q}} = \{ (s_i^{ab}, s_i^{a'b'}): (a,b) \ne (a', b').\} $ 
For the class $\Theta$ let $w$ denote everyone else's records and the target's nonsensitive attributes. Given $w$ there are four ways for completing the dataset; let $D_w^{ab}$ denote the completion with $s_i^{ab}$. Then let 
$$\pi_{ab} = \mathbb{P}_{w, \pi} (D  = D_w^{ab})$$ denote a probability distribution on the completions; we assume that $\pi_{ab} > 0$ for $(a,b) \in {\mathcal C}$, and $\sum_{(a,) \in {\mathcal C}} \pi_{ab} =1$. We take 
$\Theta $ as the set of all such $(w, \pi)$ pairs. 

Conditioning on a secret $s_i^{ab}$ then removes the remaining dataset uncertainty. Again we use the Gaussian mechanism as before; with  $g(z)=u_h s_h(z)+r_\perp(z)$ let
$${\tilde{g}}(z) = u_h\,\operatorname{clip}(u_h^\top g(z);C_\parallel)+\operatorname{clip}_2(g(z) - u_h u_h^\top g(z);C_\perp) .$$
The resulting vector release is 
$$Y_\mathbf{D}= \frac1n\sum_j {\tilde{g}}(z_j)+\sigma Z,
$$  
where $ Z\sim N(0,I_p).$ The projected mechanism releases $u_h^\top Y_\mathbf{D}. $
For $(a,b) \in {\mathcal C}$ set $v_w^{ab} = {\tilde{g}} (w_i, a, b)$; then, for a protected comparison $c = (a,b), c'=(a', b')$, the mean displacement is $\delta_{w, c, c'} = \frac1n ( v_w^{c'} - v_w^c).$ With $P = P_{\theta, s_i^c}, Q = P_{\theta, s_i^{c'}}$, we have for $K_{\rm full} = \sigma^2 I$ that 
$$\KSD_{K_{\rm full}} (P, Q_{\alpha|P})^2 = \frac{\alpha^2}{\sigma^2 n^2} || v_w^{c'} - v_w^c||^2,$$
while for the projector kernel $K_h$,
$$\KSD_{K_{h}} (P, Q_{\alpha|P})^2 = \frac{\alpha^2}{\sigma^2 n^2} [u_h^\top ] (v_w^{c'} - v_w^c)]^2.$$ and
$D_\alpha^{\rm full}
=\frac1{2\alpha}\operatorname{KSD}_{K_{\rm full}}^2 $ while 
$D_\alpha^h
=\frac1{2\alpha}\operatorname{KSD}_{K_h}^2.
$
We highlight that the projected guarantee applies only to the projected release, and the privacy specification is conditional on the declared context rather than a general model of population-level demographic dependence.

In the experiments we also consider changes in sex alone, and changes in race alone. 
Figure \ref{fig:puffer} illustrates the results. The clipping envelopes are as in \eqref{eq:clipping envelope}. The shorthand {\it arbitrary contexts} means that the envelope holds for any context $w$.
{
The Race-only, Sex-only and Joint comparisons are performed by fixing a context $w$ and varying the specified sensitive attributes. In Race-only comparisons, the target's sex is held fixed and race is varied, in Sex-only, the target's race is held fixed and sex is varied; in Joint, sex, race or both may vary. For each comparison type, we report the maximum privacy loss over the contexts evaluated in the experiment. These maxima are finite-class RPP parameters only when the evaluated context list is explicitly declared as the complete context class; otherwise they are diagnostics. }
More details can be found on the anonymous github site. 

In Figure \ref{fig:puffer}, for the full vector release, the largest divergences arising from race-only;  sex-only changes are comparable over the evaluated context class. In the projected release, race-only comparisons nearly attain the worst-case divergence for joint sex–race protection. Expanding protection from race-only to all joint comparisons adds little to the required worst-case parameter. 
This figure illustrates that restricting the released output can change the relative importance of different protected comparisons.

\begin{figure}
    \centering
    \includegraphics[width=0.8\linewidth]{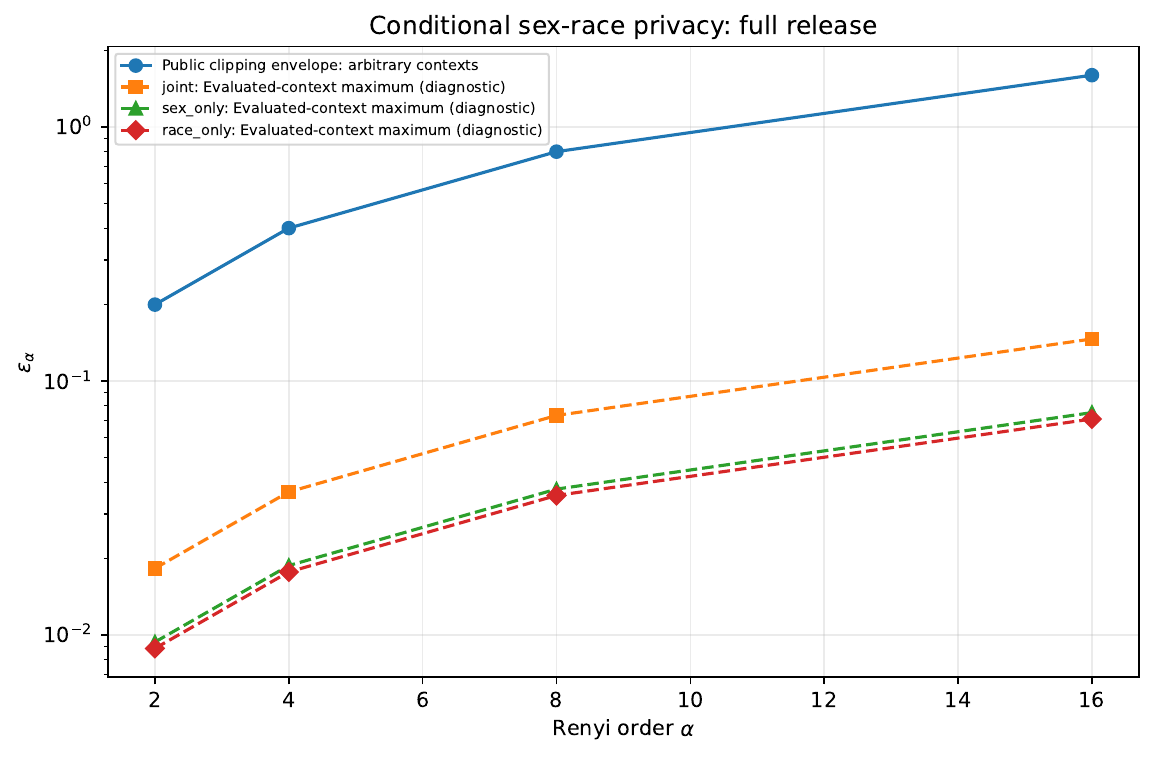}
    \includegraphics[width=0.8\linewidth]{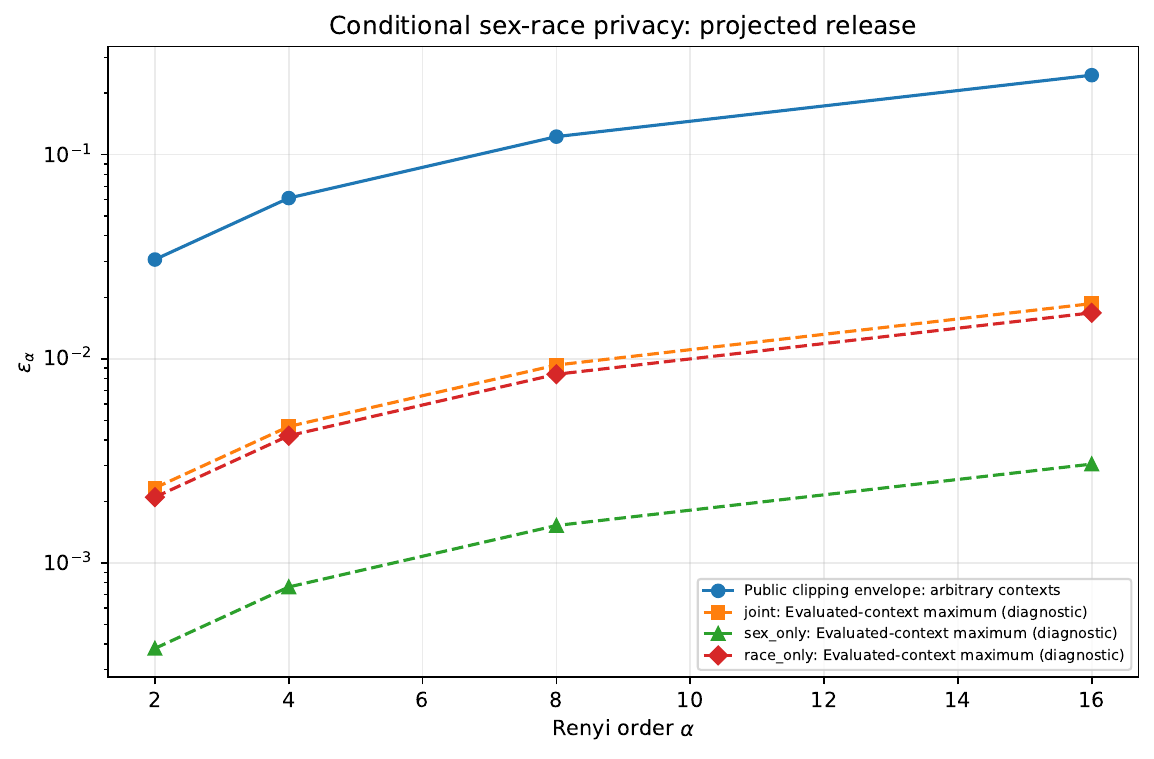}
    \caption{Privacy bounds; top: full release; bottom: projected release. Solid curves show clipping-based upper bounds that are  valid for all admissible fixed contexts. Dashed curves show exact maxima of \(\mathrm{KSD}^2/(2\alpha)=D_\alpha\) over the evaluated contexts and specified secret comparisons. }
    \label{fig:puffer}
\end{figure}}

\end{document}